**Imperial College London**

IMPERIAL COLLEGE OF SCIENCE,
TECHNOLOGY AND MEDICINE

DEPARTMENT OF COMPUTING

# Deep Learning on Real-World Graphs

Emanuele Rossi

**Supervisor**: Prof. Michael Bronstein
**Co-supervisor**: Prof. Stefanos Zafeiriou

Submitted in partial fulfilment of the requirements for the degree of
*Doctor of Philosophy in Computing*

"If we knew what we were doing, it would not be called research, would it?" – Albert Einstein

**Statement of Originality**

I hereby affirm that the work presented in this thesis is my own. Any material or ideas not originating from my efforts are duly cited and referenced in accordance with academic standards.

**Copyright Statement**

The copyright of this thesis rests with the author. Unless otherwise indicated, its contents are licensed under a Creative Commons Attribution-Non Commercial 4.0 International Licence (CC BY-NC). Under this licence, you may copy and redistribute the material in any medium or format. You may also create and distribute modified versions of the work. This is on the condition that: you credit the author and do not use it, or any derivative works, for a commercial purpose. When reusing or sharing this work, ensure you make the licence terms clear to others by naming the licence and linking to the licence text. Where a work has been adapted, you should indicate that the work has been changed and describe those changes. Please seek permission from the copyright holder for uses of this work that are not included in this licence or permitted under UK Copyright Law.

# Abstract

Graph Neural Networks (GNNs) have emerged as the primary tool for learning on graph-structured data. Yet, many GNNs are primarily suited for toy academic benchmarks, often limiting their efficacy in real-world scenarios. This thesis proposes novel approaches to enable the use of GNNs in real-world applications, including but not limited to social networks and recommender systems.

Our contributions are as follows:

- **Scalability**: We introduce Scalable Inception Graph Neural Networks (SIGN) (Chapter 3, [202]). By decoupling graph propagation from model learning, SIGN obtains comparable accuracy to previous models, but with up to 30x faster inference, enabling efficient GNNs on web-scale graphs.

- **Temporality**: We propose Temporal Graph Networks (TGN) (Chapter 4, [200]), a comprehensive framework for learning on evolving graphs. Representing a temporal graph as a series of events, TGN dynamically updates node representations through event-driven messages prior to their aggregation across the graph. TGN outperforms previous methods on both temporal node-classification and link-prediction tasks.

- **Edge Directionality**: Our research (Chapter 5, [201]) shows the importance of leveraging edge directionality, especially in heterophilic tasks. We introduce Directed Graph Neural Networks (Dir-GNN), a general framework to extend any spatial GNN to directed graphs. Incorporating edge directionality with Dir-GNN leads to large accuracy gains in heterophilic scenarios, while leaving performance unchanged on homophilic graphs.

- **Missing Data**: Addressing data incompleteness, Feature Propagation (FP) (Chapter 6, [203]) imputes missing node attributes by propagating existing ones through the graph. Paired with a downstream GNN, FP mitigates accuracy degradation to about 4% even when 99% of the features are absent, vastly outperforming previous methods.

- **Structural Inference**: We propose NuGget (Chapter 7, [205]), a method to reconstruct unknown graphs in a game-theoretic setup. NuGget recovers the network structure of a game from players' equilibrium actions, outperforming previous models without relying on explicit utility functions.

# Acknowledgements

First and foremost, I would like to express my gratitude to Michael: not only for creating the opportunity to pursue a PhD under your supervision while we were both at Twitter, but also for the support and trust you demonstrated throughout the years. Your approach, granting me the freedom to navigate my own path and learn through experience, all while ensuring your timely presence when necessary, was invaluable. I happily remember the whiteboard brainstorming sessions in the first year of my PhD, as well as the long conversations about research over lunch.

I hold a special place in my heart for my former team at Fabula, and subsequently at Twitter—Fabrizio, Davide, Federico, and Ben. The countless days we spent in the office, sharing insightful discussions, long lunches and intense table tennis matches, were both fun and a source of continuous learning. Each one of you has been a special collaborator and friend. I owe particular gratitude to Fabrizio, my companion on this PhD journey: your constant presence and collaborative spirit, despite our research diverging, was of great support. I'm grateful our paths reconnected professionally on my latest paper, marking a fruitful reunion after three years.

To Nils, my manager at Twitter: thank you for having been an emphatic mentor as much as a manager. You showed me the path to thriving in professional relationships and aligning my research with Twitter's strategic initiatives.

To my other collaborators from all over the world—Bertrand, Carlo, Daniel, Farimah, Henry, James, Maria, Matthias, Max, Mirco, Nils, Pietro, Riccardo, Sergey, Shenyang, Stefan, Suvash, Weihua, Yan, Xiaowen, I express sincere thanks; without your diverse perspectives and contributions, my research efforts would have remained incomplete.

To my family—my mother Stefania, my father Andrea, my brother Francesco, my grandparents Franca and Ilario and 'dada' Franca: thank you for being my safe harbour and for understanding my absences over these demanding years. Your unconditional love and belief in me have been my foundation. The warm welcome I receive whenever I return home annihilates the time spent apart; it's as if I never left.

Special thanks to Filippo, Andrea, Enxhell, Eugenio, Erick, Federico and Enrico: your friendship meant a lot to me during this time. Whether it was for a laugh or a listening ear, you were there. Thank you for the memories and the moments of connection that have brightened my path. Thank you also to all the other people I have crossed paths with during these years: however brief our interactions might have been, they left a lasting impact on my journey in ways I could have never anticipated.

Last but certainly not least, to my girlfriend Maria Giulia: you are an incredible companion through life's journey, and my daily reminder that research is futile without a life filled with love. You are uniquely able to help me navigate through tough times, while helping me find deeper meaning in both my personal and professional endeavors. You have shown me that professional aspirations and personal dreams aren't mutually exclusive but can beautifully enhance one another. Since meeting you, I have not only evolved as a researcher, but more so, I have embarked on a journey of profound personal transformation. Thank you!

None of this could have been achieved without your support, and I am beyond grateful to have the opportunity to share this accomplishment with each one of you.

# Contents

# 1 Introduction

> "Structure is essential in building anything that thrives."
> – Henry Cloud

## 1.1 Artificial Intelligence and Machine Learning

Artificial Intelligence (AI) began as an ambitious endeavor to replicate human intelligence in machines. This quest dates back over seventy years when Alan Turing introduced the imitation game to examine a machine's capability to think [239]. The last two decades witnessed an unprecedented acceleration in advancements, starting from recognizing objects in images [104, 141, 220] and solving fundamental biological challenges such as protein folding [126], to the incredible capabilities of language models such as ChatGPT [186] today. This was made possible by the development of Machine Learning (ML) [28], and in particular Deep Learning (DL) [144] models. In ML, the key idea is to make systems learn from large quantities of data, rather than explicitly programming the desired behaviour. DL takes this further by stacking multiple layers one after the other (hence the 'deep') to learn hierarchical and increasingly more complex representations from the data. The key ingredient to make ML [1] successful are vast quantities of data [2], often described as the new oil of the digital era. Data provides ML models with an exclusive window into the real world; after all, ML models can only learn phenomena reflected in the data they're exposed to. This creates the urge to develop models able to learn from data representing a wide variety of real-world systems [180] which cannot be accurately represented by simple sequences (such as text) or grids (such as images).

## 1.2 The Intrinsic Structure of the World

Graphs are flexible data structures that capture entities and their relationships. Their presence is universal - from social and web networks to brain functional networks. Graphs are well suited to represent interactions even at a microscopic scale, whether it's particles in a physical system, chemical bindings in a molecule, or biological interactions between proteins. What makes graphs particularly suitable for such complex systems is their ability to provide valuable insights into the system's organization and functioning. For instance, the community structure in social networks can be uncovered by analyzing the graph structure, revealing the social groups within the network [74]. In biological networks, the structure of the graph can help identify the function of interacting proteins [226]. Given the prevalence of graph structure in real-world phenomena, it is becoming increasingly important to enable machine learning models to learn from graph-structured data.

---

[1] Hereafter, we will use the terms 'Machine Learning' and 'Deep Learning' interchangeably, as commonly done in the literature.
[2] This, of course, needs to be coupled with the right computational resources to process the data.

## 1.3 Machine Learning on Graphs

Learning on graph data is possible using Graph Neural Networks (GNNs) [102], a class of Machine Learning models that typically operate by a message passing mechanism [18]. GNNs aggregate information from the neighbours of a node and create node embeddings that are then used for node-wise classification, graph-wise classification, or link-prediction tasks. GNNs have already led to a sequence of successes in many fields [19, 52, 68, 89, 171, 189, 197], in particular, social sciences [173, 269] and biology [77, 246, 298]. While a plethora of methods have been proposed that work well on toy or 'academic' datasets, these methods often fail to address many of the challenges posed by real-world graphs, such as *scalability*, *temporality*, *missing data*, *directed edges* and *structural inference*. These challenges often preclude the use of GNNs in real-world applications such as social networks or recommender systems.

## 1.4 Scope and Research Questions

This thesis is an effort to overcome the most common challenges precluding the use of GNNs in real-world applications, and it is guided by the following research questions:

SCALABILITY   *Can GNNs be utilized for learning on extremely large graphs?*
Most of the research in the GNN field has focused on small-scale datasets (CORA [212] with only ∼ 5K nodes still being among the most widely used), and relatively little effort has been devoted to scaling these methods to web-scale graphs such as the Facebook or Twitter social networks which do not fit into GPU memory. Scaling is a major challenge precluding the wide application of graph deep learning methods in industrial settings.
We introduce SIGN in Chapter 3. SIGN is a GNN model able to scale to graphs with billions of edges thanks to its decoupling of the graph propagation and learning stages. SIGN leads to a speedup of up to two orders of magnitude in inference compared to previous scalable methods, while obtaining a comparable predictive accuracy.

TEMPORALITY   *Can GNNs be utilized for learning on graphs that change over time?*
The majority of methods for deep learning on graphs assume that the underlying graph is static. However, most real-life systems of interactions such as social networks or biological interactomes are *dynamic*, and often, it is the dynamic structure that contains crucial insights about the system [263]. We introduce TGN in Chapter 4. TGN is a general framework for learning on temporal graphs, and it is the first to combine a memory component with a GNN component. We present applications of TGN for both temporal node classification and link prediction in Chapter 4, where it shows significant advantages over earlier state-of-the-art methods both in terms of efficiency and predictive accuracy.

DIRECTIONALITY   *Can GNNs leverage directed edges?*
Many real-world graphs, such as transaction and interaction networks, have *directed* edges. However, the majority of GNN models discard this information altogether by simply making the graph undirected.
We introduce Dir-GNN in Chapter 5. Dir-GNN is a general framework for extending GNNs to handle edge-directionality effectively. Incorporating edge directionality with Dir-GNN leads to significant gains in predictive accuracy when dealing with heterophilic graphs, i.e. graphs where neighbors tend to have different labels, while leaving performance mostly unchanged on homophilic graphs. Notably, Dir-GNN obtains state-of-the-art results on directed heterophilic node classification tasks, outperforming complex methods specifically designed for heterophilic scenarios.

MISSING DATA   *Can GNNs be utilized for learning on graphs that have partially missing node features?*
GNN models typically assume a fully observed feature matrix, where rows represent nodes and columns represent feature channels. However, in real-world scenarios, each feature is often only observed for a subset of the nodes. For example, demographic information can be available for only a small subset of social network users who gave explicit consent.
We introduce Feature Propagation (FP) in Chapter 6. FP is a pre-processing method which reconstructs missing features through iterative graph propagation. When paired with a downstream GNN, FP can withstand extremely high rates of missing features, leading to only around 4% relative accuracy drop on node classification tasks when 99% of the features are missing. All this while also being extremely scalable and simple to implement, making it particularly suitable for industrial applications.

STRUCTURAL INFERENCE   *Can we recover the graph from players' actions in a network game?*
Most of the GNN literature assumes the graph to be known beforehand. However, in practical settings, the underlying interaction network often remains hidden or uncertain due to privacy reasons or the dynamic nature of interactions. This is often the case in network games, where players are connected in a network (graph), but we only observe their actions. Each player tries to maximise their utility function, which in the case of network games depends both on their own actions and the actions of their neighbours.
We introduce NuGget in Chapter 6. NuGget is a transformer-like model which learns a mapping from the equilibrium actions to the network structure of the game. NuGget outperforms earlier state-of-the-art methods, while also being more general as it does not require explicit knowledge of the utility function.

Additionally to the main contributions listed above, the thesis contains the necessary background in Chapter 2, '*Fast forward to today*' discussion paragraphs in the conclusion section at the end of each chapter, as well as interesting future directions in Chapter 8.

## 1.5 OWN PUBLICATIONS

This thesis is based on the following publications:

- **Emanuele Rossi**\*, Fabrizio Frasca\*, Ben Chamberlain, Davide Eynard, Michael Bronstein, and Federico Monti, "*SIGN: Scalable Inception Graph Neural Networks*", ICML Workshop on Graph Representation Learning 2020 [202].

- **Emanuele Rossi**, Ben Chamberlain, Fabrizio Frasca, Davide Eynard, Federico Monti, and Michael Bronstein, "*Temporal Graph Networks for Deep Learning on Dynamic Graphs*", ICML Workshop on Graph Representation Learning 2020 [200].

- **Emanuele Rossi**, Henry Kenlay, Maria I. Gorinova, Ben Chamberlain, Xiaowen Dong, and Michael M. Bronstein, "*On the Unreasonable Effectiveness of Feature propagation in Learning on Graphs with Missing Node Features*", Learning on Graphs Conference 2022 [203].

- **Emanuele Rossi**, Federico Monti, Yan Leng, Michael M. Bronstein, and Xiaowen Dong, "*Learning to Infer Structures of Network Games*", Proceedings of the 39th International Conference on Machine Learning, ICML 2022 [205].

- **Emanuele Rossi**, Bertrand Charpentier, Francesco Di Giovanni, Fabrizio Frasca, Stephan Günnemann, and Michael M. Bronstein, "*Edge Directionality Improves Learning on Heterophilic Graphs*", arXiv 2023 [201].

The author has additionally contributed to the following publications:

- **Emanuele Rossi**, Federico Monti, Michael Bronstein, Pietro Liò, "*ncRNA Classification with Graph Convolutional Networks*", KDD 2019 Workshop on Deep Learning on Graphs [204].

- Ben Chamberlain, **Emanuele Rossi**, Dan Shiebler, Suvash Sedhain, and Michael Bronstein, "*Tuning Word2vec for Large Scale Recommendation Systems*", RecSys - 14th ACM Conference on Recommender Systems 2020 [39].

- Ben Chamberlain, James Rowbottom, Maria I. Gorinova, Stefan D. Webb, **Emanuele Rossi**, and Michael M. Bronstein, "*Graph Neural Diffusion*", Proceedings of the 38th International Conference on Machine Learning, ICML 2021 [40].

- Mirco Mutti, Riccardo De Santi, **Emanuele Rossi**, Juan Felipe Calderon, Michael M. Bronstein, and Marcello Restelli, "*Provably Efficient Causal Model-Based Reinforcement Learning for Systematic Generalization*", Proceedings of the AAAI Conference on Artificial Intelligence 2022 [177].

- Ben Chamberlain, Sergey Shirobokov, **Emanuele Rossi**, Fabrizio Frasca, Thomas Markovich, Nils Hammerla, Michael M. Bronstein, and Max Hansmire, "*Graph Neural Networks for Link Prediction with Subgraph Sketching*", International Conference on Learning Representations (ICLR) 2022 [41].

- Shenyang Huang, Farimah Poursafaei, Jacob Danovitch, Matthias Fey, Weihua Hu, **Emanuele Rossi**, Jure Leskovec, Michael Bronstein, Guillaume Rabusseau, Reihaneh Rabbany, "*Temporal Graph Benchmark for Machine Learning on Temporal Graphs*", Thirty-sixth Conference on Neural Information Processing Systems Datasets and Benchmarks Track (NeurIPS 2023) [115].

# 2 Background

> "We cannot live only for ourselves. A thousand fibers
> connect us with our fellow men." – Herman Melville

This chapter will describe the essential mathematical tools required to understand graphs. Equipped with this foundation, we will introduce the node classification task, Graph Neural Networks, and conclude with the concept of graph homophily.

## 2.1 Graphs

Graphs are powerful mathematical objects used to describe entities and their relationships. A graph is composed of vertices (or nodes) representing the entities, and edges connecting these vertices, indicating the relationships or interactions between them. Given an undirected graph $G = (V, E)$ with vertex and edge sets $V$ and $E$ respectively, its adjacency matrix $\mathbf{A}$ is defined as $a_{ij} = 1$ if $(i, j) \in E$ and zero otherwise. Let $n = |V|$ and $m = |E|$ be the number of nodes and edges respectively, and $\mathcal{N}(i) = \{j : (i, j) \in E\}$ be the set of neighbors of $i$. We let $\hat{\mathbf{A}} = \mathbf{A} + \mathbf{I}$ be the adjacency matrix with added self-loops, $\hat{\mathbf{D}} = \text{diag}(d_i)$ the degree matrix of $\hat{\mathbf{A}}$ and $\tilde{\mathbf{A}} = \hat{\mathbf{D}}^{-\frac{1}{2}} \hat{\mathbf{A}} \hat{\mathbf{D}}^{-\frac{1}{2}}$ be the symmetrically normalized adjacency matrix. The graph laplacian matrix is defined as $\mathbf{L} = \hat{\mathbf{D}} - \hat{\mathbf{A}}$, with its normalized version being $\mathbf{\Delta} = \hat{\mathbf{D}}^{-\frac{1}{2}} \mathbf{L} \hat{\mathbf{D}}^{-\frac{1}{2}} = \mathbf{I} - \tilde{\mathbf{A}}$. We denote by $\mathbf{X} \in \mathbb{R}^{n \times d}$ the matrix of $d$-dimensional node features and by $\mathbf{x}_i$ its $i$-th row (transposed).

## 2.2 Node Classification

Node classification is a fundamental task in graph-based learning where the primary goal is to predict the label or class of nodes in a graph. Formally, the goal is to infer a function $f : V \to \mathbb{Y}$ that assigns each node to a label in the label space $\mathbb{Y}$. Typically, a portion of the nodes in $V$ have known labels, and the task is to predict the labels for the remaining unlabeled nodes.

While label propagation [295] and random walks [98, 194] have been historically used to tackle node classification tasks, these approaches are unable to leverage the rich node features that graphs often present, limiting themselves to the topological information.

On the other hand, Graph Neural Networks prove especially beneficial for node classification tasks, as they can leverage both the node's features and its relational context within the graph, providing an informative representation to predict the node's label. In practice, GNNs have been extensively employed for various node classification tasks such as categorizing scientific articles in citation networks [136], discerning protein types in biological networks [299], and even in social networks to predict user roles or behaviors [102]. We introduce Graph Neural Networks next.

## 2.3 Graph Neural Networks

Broadly, GNNs are learnable functions which take as input a graph, along with its node features, and output embeddings for the nodes in the graph:

$$\mathbf{H} = \text{GNN}_\Theta(G, \mathbf{X}) \tag{2.1}$$

where $\mathbf{H} \in \mathbb{R}^{n \times h}$ is a matrix of node embeddings and $\Theta$ are the learnable parameters of the model. Depending on the task at hand, a prediction can then be made from the embeddings. In the case of node classification, a simple linear decoder followed by a softmax activation can be used to make node-wise predictions $\mathbf{y}_i = \text{softmax}(\mathbf{H}\boldsymbol{\Omega})$, where $\boldsymbol{\Omega} \in \mathbb{R}^{n \times c}$ is a learnable matrix of parameters.

As common in deep learning, a GNN is obtained by stacking a series of (graph convolutional) layers. Each layer takes as input the graph, as well as the node representations produced by the previous layer:

$$\mathbf{H}^{(k)} = \text{GCL}_{\Theta^{(k)}}(G, \mathbf{H}^{(k-1)}) \tag{2.2}$$

where $\mathbf{H}^{(0)} = \mathbf{X}$.

The output embeddings of a $K$-layer model is then $\mathbf{H} = \mathbf{H}^{(K)}$.

A wide variety of graph convolutional layers have been proposed in the literature [8, 62, 102, 136, 171, 183, 210, 219, 244, 258]. We now describe two popular layers that on top of being extremely successful also represent two different approaches: the convolutional and the message-passing layers.

CONVOLUTIONAL GNNs   The GCN model [136] defines a single graph convolutional layer as

$$\mathbf{H}^{(k)} = \sigma(\tilde{\mathbf{A}}\mathbf{H}^{(k-1)}\mathbf{W}^{(k)}) \tag{2.3}$$

where $\mathbf{W}^{(k)}$ is a matrix of learnable weights and $\sigma$ is a non-linear activation function. This amounts to aggregating node features from direct neighbours in the graphs at each layer, with weights depending on the degree of the two nodes (since $\tilde{\mathbf{A}}_{ij} = \frac{1}{\sqrt{d_i d_j}}$). After the aggregation, a linear transformation is applied to the resulting features, before feeding the output into a non-linearity. GCN-like models are highly efficient as they only consist of a sparse-to-dense matrix multiplication ($\tilde{\mathbf{A}}\mathbf{H}^{(k-1)}$) plus a dense multiplication by $\mathbf{W}^{(k)}$. However, their propagation scheme is fixed and cannot be adapted depending on the features of the source and destination nodes.

MESSAGE PASSING GNNs   A more general perspective sees GNNs as operating through a message-passing mechanism. A Message-Passing Neural Network (MPNN [89]) is a parametric model which *iteratively* applies aggregation maps $\text{AGG}^{(k)}$ and combination maps $\text{COM}^{(k)}$ to compute embeddings $\mathbf{h}_i^{(k)}$ for node $i$ based on messages $\mathbf{m}_i^{(k)}$ containing information on its neighbors. Namely, an MPNN layer is given by

$$\begin{aligned}\mathbf{m}_i^{(k)} &= \text{AGG}^{(k)}\left(\{\!\!\{(\mathbf{h}_j^{(k-1)}, \mathbf{h}_i^{(k-1)}) : (i,j) \in E\}\!\!\}\right) \\ \mathbf{h}_i^{(k)} &= \text{COM}^{(k)}\left(\mathbf{h}_i^{(k-1)}, \mathbf{m}_i^{(k)}\right)\end{aligned} \tag{2.4}$$

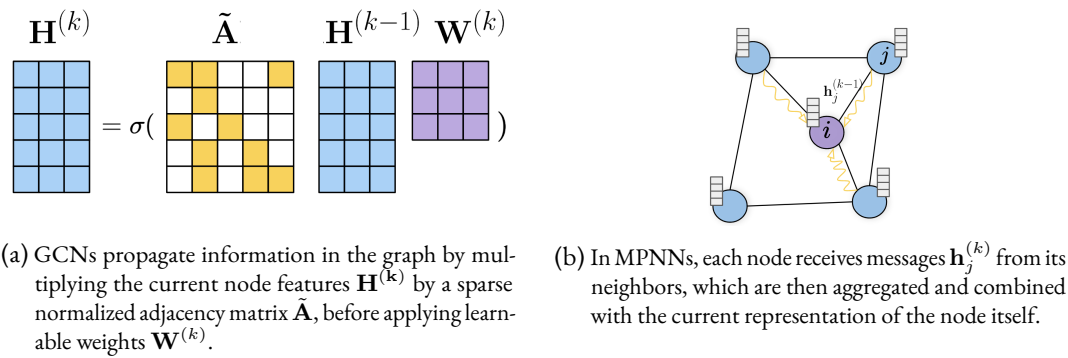

(a) GCNs propagate information in the graph by multiplying the current node features $\mathbf{H}^{(k)}$ by a sparse normalized adjacency matrix $\tilde{\mathbf{A}}$, before applying learnable weights $\mathbf{W}^{(k)}$.

(b) In MPNNs, each node receives messages $\mathbf{h}_j^{(k)}$ from its neighbors, which are then aggregated and combined with the current representation of the node itself.

Figure 2.1: Two popular GNN layers: GCN (left) and MPNN (right).

where $\mathbf{h}_i^{(0)} = \mathbf{x}_i$ and $\{\!\{\cdot\}\!\}$ is a multi-set. The aggregation maps $\text{AGG}^{(k)}$, and the combination maps $\text{COM}^{(k)}$ are *learnable* (usually a small neural network) and their different implementations result in specific architectures.

Independent of the specific implementation, all MPNNs only send messages along the edges of the graph, which makes them particularly suitable for tasks where edges do encode a notion of similarity – as it is the case when adjacent nodes in the graph often share the same label. Conversely, MPNNs tend to struggle in the scenario where they need to separate a node embedding from that of its neighbours [184], often a challenging problem that has gained attention in the community and which we discuss in detail next.

## 2.4 Homophily and Heterophily

Homophily is a concept born in sociology to indicate the tendency of individuals to bond with other individuals who are similar to them, leading to the common proverb "birds of a feather flock together". More recently, homophily has started to be studied in the context of networks, leading to evidence that social networks present a high degree of homophily [169], i.e., neighbouring nodes tend to share the same attributes.

Recently, it has been shown that most GNNs (implicitly) assume the input graph to be homophilic, leading to poor performance on heterophilic graph tasks [2]. While a reasonable assumption in some settings, homophily is not present in many important applications such as gender classification on social networks or fraudster detection on e-commerce networks.

**Homophily metrics.** Several metrics have been proposed to measure homophily on undirected graphs. *Node homophily* [154, 192] is defined as

$$h = \frac{1}{|V|} \sum_{i \in V} \frac{\sum_{j \in \mathcal{N}(i)} I[y_i = y_j]}{d_i} \quad (2.5)$$

where $I[y_i = y_j]$ is the indicator function with value 1 if $y_i = y_j$ or zero otherwise. Intuitively, node homophily measures the fraction of neighbors with the same label as the node itself, averaged across all nodes. However, since heterophily is a complex phenomenon which is hard to capture with only a single scalar, a better representation of a graph's homophily is the $C \times C$ *class compati-* *bility matrix* **H** [84, 294], capturing the fraction of edges from nodes with label $k$ to nodes with label $l$:

$$h_{kl} = \frac{|(i,j) \in E : y_i = k \wedge y_j = l|}{|(i,j) \in E : y_i = k|}.$$

*Homophilic* datasets are expected to have most of the mass of their compatibility matrices concentrated in the diagonal, as most of the edges are between nodes of the same class. Conversely, *heterophilic* datasets will have most of the mass away from the diagonal of the compatibility matrix.

# 3 Scalable Graph Neural Networks

> "The larger the island of knowledge, the longer the
> shoreline of wonder." - Ralph W. Sockman

## 3.1 Introduction

Until recently, most of the research in the GNN field has focused on small-scale datasets, and relatively little effort has been devoted to scaling these methods to web-scale graphs such as the Facebook or Twitter social networks. Scaling is a major challenge precluding the broad application of graph deep learning methods in industrial settings. Compared to other neural networks where the training loss can be decomposed into individual samples and computed independently, graph convolutional networks diffuse information between nodes along the edges of the graph, making the loss computation interdependent for different nodes. Furthermore, in typical graphs the number of nodes grows exponentially with the increase of the filter receptive field, incurring significant computational and memory complexity. So far, various *graph sampling* approaches [43, 45, 49, 102, 116, 269, 280, 300] have been proposed as a way to alleviate the cost of training graph neural networks by selecting a small number of neighbors to reduce the computational and memory complexity. However, such methods require sampling of the neighbourhood graph each time a new prediction for a node is required, making online inference prohibitively expensive.

CONTRIBUTION    In this chapter, we take a different approach to scalable deep learning on graphs. We propose *SIGN*, a simple scalable Graph Neural Network architecture decoupling the graph propagation from the learning stage. *SIGN* combines graph convolutional filters of different types and sizes that are amenable to efficient precomputation, allowing extremely fast training and inference with complexity independent of the graph structure. Our architecture is able to scale to web-scale graphs without resorting to any sample technique and retaining sufficient expressiveness for effective learning: while being faster in training and, especially, inference (even one order of magnitude speedup), by employing *SIGN* with only one graph convolutional layer we are able to achieve results on par with state-of-the-art on several large-scale graph learning datasets. In particular, *SIGN* obtains state-of-the-art results on `ogbn-papers100M`, the largest public graph learning benchmark, with over 110 million nodes and 1.5 billion edges.

## 3.2 Background and Related Work

### 3.2.1 Graph sampling

For Web-scale graphs such as Facebook or Twitter that typically have $n = 10^8 \sim 10^9$ nodes and $m = 10^{10} \sim 10^{11}$ edges, the diffusion matrix cannot be stored in memory for training, making the straightforward application of graph neural networks impossible. Graph sampling has been shown to be a successful technique to scale GNNs to large graphs by approximating local connectivity with subsampled versions, which are amenable to computation.

3 Scalable Graph Neural NetworksTable 3.1: Theoretical time complexity where $L_c$, $L_{ff}$ is the number of graph convolution and MLP layers respectively, $r$ is the number of diffusion operators, $N$ the number of nodes (in training or inference, respectively), $m$ the number of edges, and $d$ the feature dimensionality (assumed fixed for all layers). For *GraphSAGE*, $k$ is the number of sampled neighbors per node. Forward pass complexity corresponds to an entire epoch where all nodes are seen.

|  | Preprocessing | Forward Pass |
| --- | --- | --- |
| GraphSAGE | $\mathcal{O}(k^{L_c} n)$ | $\mathcal{O}(k^{L_c} n d^2)$ |
| ClusterGCN | $\mathcal{O}(m)$ | $\mathcal{O}(L_c m d + L_{ff} n d^2)$ |
| GraphSAINT | $\mathcal{O}(kn)$ | $\mathcal{O}(L_c m d + L_{ff} n d^2)$ |
| SIGN-$r$ | $\mathcal{O}(rmd)$ | $\mathcal{O}(r L_{ff} n d^2)$ |

NODE-WISE SAMPLING   These strategies perform graph convolutions on *partial* node neighbourhoods to reduce computational and memory complexity and are coupled with minibatch training, where each training step is performed only on a batch of nodes rather than on the whole graph. A training batch is assembled by first choosing $b$ 'optimization' nodes and partially expanding their corresponding neighbourhoods. In a single training step, the loss is computed and optimized only for optimization nodes. Node-wise sampling coupled with minibatch training was first introduced in *GraphSAGE* [102] to address the challenges of scaling GNNs. PinSAGE [269] extended *GraphSAGE* by exploiting a neighbour selection method using scores from approximations of Personalized PageRank [103] via random walks. *VR-GCN* [43] uses control variates to reduce the variance of stochastic training and increase the speed of convergence with a small number of neighbors.

LAYER-WISE SAMPLING   A characteristic of many graphs, in particular 'small-world' social networks, is the exponential growth of the neighbourhood size with a number of hops $L$. Chen et al. [45] and Huang et al. [116] avoid over-expansion of neighbourhoods to overcome the redundancy of node-wise sampling. Nodes in each layer only have directed edges towards nodes of the next layer, thus bounding the maximum amount of computation to $\mathcal{O}(b^2)$ per layer. Moreover, sharing common neighbors prevents feature replication across the batch, drastically reducing the memory complexity during training.

GRAPH-WISE SAMPLING   In Chiang et al. [49] and Zeng et al. [280], feature sharing is further advanced: each batch consists of a connected subgraph and at each training iteration the GNN model is optimized over all nodes in the subgraph. In *ClusterGCN* [49], non-overlapping clusters are computed as a pre-processing step and then sampled during training as input minibatches. *GraphSAINT* [280] adopts a similar approach, while also correcting for the bias and variance of the minibatch estimators when sampling subgraphs for training. It also explores different schemes to sample the subgraphs such as a random walk-based sampler, which is able to co-sample nodes having high influence on each other and guarantees each edge has a non-negligible probability of being sampled.

## 3.3 Scalable Inception Graph Neural Networks

In this work, we propose *SIGN*, an alternative method to scale Graph Neural Networks to very large graphs. The key building block of our architecture is a set of linear diffusion operators represented as $n \times n$ matrices $\mathbf{A}_1, \ldots, \mathbf{A}_r$, whose application to the node-wise features can be pre-computed. For node-wise classification tasks, our architecture has the form ( Fig. 3.1):

$$\begin{aligned} \mathbf{Z} &= \sigma([\mathbf{X}\boldsymbol{\Theta}_0, \mathbf{A}_1\mathbf{X}\boldsymbol{\Theta}_1, \ldots, \mathbf{A}_r\mathbf{X}\boldsymbol{\Theta}_r]) \\ \mathbf{Y} &= \xi(\mathbf{Z}\boldsymbol{\Omega}), \end{aligned} \tag{3.1}$$

where $\boldsymbol{\Theta}_0, \ldots, \boldsymbol{\Theta}_r$ and $\boldsymbol{\Omega}$ are learnable matrices respectively of dimensions $d \times d'$ and $d'(r+1) \times c$ for $c$ classes, and $\sigma, \xi$ are non-linearities, the second one computing class probabilities, e.g. via softmax or sigmoid function, depending on the task at hand. We denote a model with $r$ operators by *SIGN-r*.

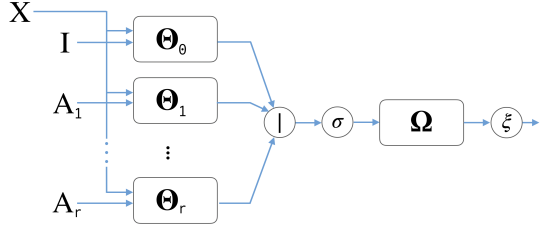

Figure 3.1: The *SIGN* architecture for $r$ generic graph filtering operators. $\boldsymbol{\Theta}_k$ represents the $k$-th dense layer transforming node-wise features downstream the application of operator $k$, | is the concatenation operation and $\boldsymbol{\Omega}$ refers to the dense layer used to compute final predictions.

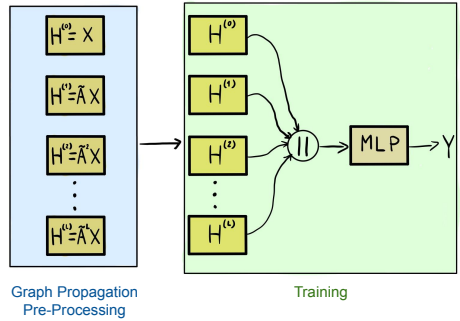

Figure 3.2: *SIGN* introduces a decoupling of the graph propagation and training. In this example, operator $\mathbf{A}_i$ is chosen to be the corresponding power of the symmetrically normalized adjacency matrix $\tilde{\mathbf{A}}^i$, while the learnable weights are condensed into a single MLP applied after concatenation of the precomputed features.

DECOUPLING GRAPH PROPAGATION FROM TRAINING  A key observation is that matrix products $\mathbf{A}_1\mathbf{X}, \ldots, \mathbf{A}_r\mathbf{X}$, in Sec. 3.3 *do not depend* on the learnable model parameters and can be easily precomputed and then re-used during training ( Fig. 3.2). For large graphs, distributed computing infrastructures such as Apache Spark can speed up computation. This effectively reduces the computational complexity of the overall model to that of a Multi-Layer Perceptron (MLP), i.e. $\mathcal{O}(rL_{\text{ff}}nd^2)$, where $d$ is the number of features, the number of nodes in the training/testing graph and $L_{\text{ff}}$ is the overall number of feed-forward layers in the model.

COMPUTATIONAL COMPLEXITY  Tab. 3.1 compares the complexity of our *SIGN* model to the other scalable architectures *GraphSAGE*, *ClusterGCN*, and *GraphSAINT*. While all models scale linearly w.r.t. the number of nodes $N$, we show experimentally that our model is significantly faster than all others due to the fact the forward and backward pass complexity of our model does not depend on the graph structure. Morever, unlike the aforementioned scalable methods, *SIGN* is not based on sampling nodes or subgraphs, operations potentially introducing bias into the optimization procedure.

INCEPTION-LIKE MODULE   Within the *SIGN* framework, it is possible to choose one specific operator $\mathbf{B}$ and to define $\mathbf{A}_k = \mathbf{B}^k$ for $k = 1, \ldots, r$. In this setting, the proposed model is analogous to the popular *Inception module* [229] for classic CNN architectures: it consists of convolutional filters of different sizes determined by the parameter $r$, where $r = 0$ corresponds to $1 \times 1$ convolutions in the inception module (amounting to linear transformations of the features in each node without diffusion across nodes). Owing to this analogy, we refer to our model as the *Scalable Inception Graph Network* (*SIGN*). Additionally, we find connections to multiscale approaches in complex networks [143].

CHOICE OF THE OPERATORS   Generally speaking, the choice of diffusion operators jointly depends on the task, graph structure, and features. In complex networks such as social graphs, operators induced by triangles or cliques might help distinguish edges representing weak or strong ties [95]. In graphs with noisy connectivity, it was shown that diffusion operators based on Personalized PageRank (PPR) or Heat Kernel can boost performance [138]. In our experiments, we choose three specific types of operators: simple (normalized) adjacency, Personalized PageRank-based adjacency, and triangle-based adjacency matrices, as well as their powers. We denote by *SIGN(p,s,t)* with $r = p + s + t$ the configuration using up to the $p$-th, $s$-th, and $t$-th power of simple GCN-normalized, PPR-based, and triangle-based adjacency matrices, respectively. Lastly, when working on directed graphs, *SIGN* can be equipped with powers of the (properly normalized) directed adjacency matrix $\mathbf{A}_d$ and its transpose $\mathbf{A}_d^\top$, in addition to the standard operators built on top of its undirected counterpart $\mathbf{A}_u = \frac{1}{2}(\mathbf{A}_d + \mathbf{A}_d^\top)$.

SIGN AND S-GCN   *SIGN* and *S-GCN* are the only graph neural models which are inherently 'shallow': contrary to standard GNN architectures, graph convolutional layers are not sequentially stacked, but either collapsed into a single linear filtering operation (S-GCN) or applied in parallel to obtain multi-scale node representations capturing diverse connectivity patterns depending on the chosen operators (*SIGN*). This is the crucial feature that allows these models to naturally scale their training and inference to graphs of any size and family given that all graph operations can be conveniently pre-computed. While we notice that *S-GCN* can be considered as a specific configuration of *SIGN*-1, we remark the fact that the more general *SIGN* architecture easily allows to incorporate more expressivity via parallel application of several, possibly, domain-specific, operators. We experimentally show this in the following section where we demonstrate that the *S-GCN* paradigm is indeed too limiting and that a more expressive *SIGN* model is not only able to perform on par with 'deeper' sampling-based models, but also to achieve state-of-the-art results on the largest publicly available graph learning benchmark.

## 3.4 EXPERIMENTS

DATASETS   We evaluated the proposed method on node-wise classification tasks, both in transductive (where test nodes, but not their labels, are available during training) and inductive (where test nodes are not observed during training) settings. Inductive experiments are performed using four datasets: Reddit [102], Flickr, Yelp [280], and PPI [299]. To date, these are the largest graph learning inductive node classification benchmarks available in the public domain. Related tasks are multiclass node-wise classification for `Reddit` and `Flickr` and multilabel classification for `Yelp` and `PPI`. Transductive experiments were performed on the new `ogbn-products` and `ogbn-papers100M` datasets [112]. The former represents an Amazon product co-purchasing network [27] where the task is to predict the category of a product in a multi-class classification setup. The latter represents a directed citation network of $\sim 111$ million academic papers, where the task is to leverage information from the entire citation network to infer the labels (subject areas) of a smaller subset of ArXiv papers. Overall, this dataset is orders of magnitude larger than any existing node classification dataset and is, therefore, the most important testbed for the scalability of *SIGN* and related methods.

Furthermore, we also test the scalability of our method on `Wikipedia` links [256], a large-scale network of links between articles in the English version of Wikipedia.

Statistics for all the datasets are reported in Tab. 3.2.

SETUP    We tested several *SIGN*$(p, s, t)$ configurations, with $p$ the maximum power of the GCN-normalized adjacency matrix, $s$ that of a random-walk normalized PPR diffusion operator [138], and $t$ that of a row-normalized triangle-induced adjacency matrix [174], with weights proportional to edge occurrences in closed triads. PPR-based operators are computed from a symmetrically normalized adjacency transition matrix in an approximated form, with a restart probability of $\alpha = 0.01$ for inductive datasets and $\alpha = 0.05$ in the transductive case. To allow for larger model capacity in the inception modules and in computing final model predictions, we replace the single-layer projections performed by $\boldsymbol{\Theta}_i$ and $\boldsymbol{\Omega}$ modules with multiple feedforward layers. Model parameters are found by minimizing the cross-entropy loss via minibatch gradient descent with the Adam optimizer [131]. Early stopping is applied with a patience of 15. In order to limit overfitting, we apply the standard regularization techniques of weight decay and dropout [227]. Additionally, batch-normalization [119] was used in every layer to stabilize training and increase convergence speed. Architectural and optimization hyperparameters were estimated using Bayesian optimization with a tree Parzen estimator surrogate function [24] over all inductive datasets. As for the the transductive setting, we employ standard exhaustive search on a predefined hyperparameter grid on `ogbn-products`, while on `ogbn-papers100M` we only test a basic configuration that can be found in Supplementary Materials, along with further details on the hyperparameter search spaces. Given that this dataset represents a directed network, we experimented with operators built via asymmetric normalization of the original directed adjacency matrix and its transpose, as well as their powers. The `Wikipedia` dataset, due to the lack of node attributes and labels, is only used to assess scalability: to this end, we randomly generate 100-dimensional node feature vectors and scalar targets and consider the whole network for both training and inference. No hyperparameter tuning is required in this case.

BASELINES    On the inductive datasets, we compare our method with *GCN* [136], *FastGCN* [45], *Stochastic-GCN* [43], *AS-GCN* [116], *GraphSAGE* [102], *ClusterGCN* [49], and *GraphSAINT* [280], which constitute the current state-of-the-art. On `ogbn-products` we compare against scalable sampling-free baselines, i.e. a feed-forward network trained over node features only (*MLP*) and on their concatenation with structural *Node2Vec* embeddings [98], and the sampling-based approaches *ClusterGCN* [49] and *GraphSAINT* [280]. As for `ogbn-papers100M`, *SIGN* is compared with sampling-free baselines: an *MLP* trained on node features and an *S-GCN* model. Sampling-based methods have not been scaled yet to this benchmark. All results for OGB datasets are directly taken from the latest version of the arXiv paper [112] (*v4*, at the time of writing). Lastly, being

Table 3.2: Summary of single (s) and multi-label (m) dataset statistics. Wikipedia is used, with random features, for timing purposes only.

| Dataset | | $m$ | Avg. Deg. | $d$ | Classes | Train / Val / Test |
|---|---|---|---|---|---|---|
| Wikipedia | 12,150,976 | 378,142,420 | 62 | 100 | 2(s) | 100% / — / 100% |
| ogbn-papers100M | 111,059,956 | 1,615,685,872 | 30 | 128 | 172(s) | 78% / 8% / 14% |
| ogbn-products | 2,449,029 | 61,859,140 | 51 | 100 | 47(s) | 10% / 2% / 88% |
| Reddit | 232,965 | 11,606,919 | 50 | 602 | 41(s) | 66% / 10% / 24% |
| Yelp | 716,847 | 6,977,410 | 10 | 300 | 100(m) | 75% / 10% / 15% |
| Flickr | 89,250 | 899,756 | 10 | 500 | 7(s) | 50% / 25% / 25% |
| PPI | 14,755 | 225,270 | 15 | 50 | 121(m) | 66% / 12% / 22% |

Table 3.3: Mean and standard deviation of preprocessing, training (one epoch) and inference times, in seconds, on `ogbn-products` and `Wikipedia` datasets, computed over 10 runs. SIGN-$r$ denotes architecture with $r$ precomputed operators. Preprocessing and training times for ClusterGCN on `Wikipedia` are not reported due to the clustering algorithm failing to complete.

| Model | ogbn-products | | | Wikipedia | | |
|---|---|---|---|---|---|---|
| | Preprocessing | Training | Inference | Preprocessing | Training | Inference |
| ClusterGCN | 36.93 ± 0.52 | 13.34 ± 0.16 | 93.00 ± 0.68 | — | — | 183.76 ± 3.01 |
| GraphSAINT | 52.06 ± 0.54 | 2.89 ± 0.05 | 94.76 ± 0.81 | 123.60 ± 1.60 | 135.73 ± 0.06 | 209.86 ± 4.73 |
| SIGN-2 (Ours) | 88.21 ± 1.33 | 1.04 ± 0.10 | 2.86 ± 0.10 | 192.88 ± 0.12 | 62.37 ± 0.17 | 13.40 ± 0.15 |
| SIGN-4 (Ours) | 160.16 ± 1.20 | 1.54 ± 0.04 | 3.79 ± 0.08 | 326.21 ± 1.14 | 93.84 ± 0.08 | 18.15 ± 0.05 |
| SIGN-6 (Ours) | 226.48 ± 1.43 | 2.05 ± 0.00 | 4.84 ± 0.08 | 459.24 ± 0.14 | 125.24 ± 0.03 | 22.94 ± 0.02 |
| SIGN-8 (Ours) | 297.92 ± 2.92 | 2.53 ± 0.04 | 5.88 ± 0.09 | 598.67 ± 0.82 | 154.73 ± 0.12 | 27.69 ± 0.11 |

*S-GCN* an important baseline for our model, we additionally report its performance on all the other datasets as well. In this case, we choose power $L$ of its (only) operator $\mathbf{A}^L$ as the value $p$ of the best corresponding *SIGN(p,s,t)* configuration and we tune its hyperparameters in the same space searched for *SIGN*.

IMPLEMENTATION  *SIGN* is implemented using Pytorch [190]. All experiments, including timings, were run on an AWS p2.8xlarge instance, with 8 NVIDIA K80 GPUs, 32 vCPUs, a processor Intel(R) Xeon(R) CPU E5-2686 v4 @ 2.30GHz and 488GiB of RAM.

### 3.4.1 RESULTS

INDUCTIVE   Tab. 3.4 presents the results on the inductive dataset. In line with [280], we report the micro-averaged F1 score means and standard deviations computed over 10 runs. For each dataset we report the best performing *SIGN* configuration, specifying the maximum power for each of the three employed operators. *SIGN* outperforms other methods on `Reddit` and `Flickr`, and performs competitively to state-of-the-art on `PPI`. Our performance on `Yelp` is worse than in the other datasets; we hypothesize that a more tailored choice of operators is required to better suit the characteristics of this dataset. Interestingly, *SIGN* significantly outperforms *S-GCN* in all datasets, suggesting that the additional expressivity introduced by the different operators in our model is required for effective learning.

Table 3.4: Micro-averaged F1 scores. For SIGN, we show the best performing configurations.

| Model | Reddit | Flickr | PPI | Yelp |
| --- | --- | --- | --- | --- |
| GCN [136] | 0.933±0.000 | 0.492±0.003 | 0.515±0.006 | 0.378±0.001 |
| FastGCN [45] | 0.924±0.001 | 0.504±0.001 | 0.513±0.032 | 0.265±0.053 |
| Stochastic-GCN [43] | 0.964±0.001 | 0.482±0.003 | 0.963±0.010 | 0.640±0.002 |
| AS-GCN [116] | 0.958±0.001 | 0.504±0.002 | 0.687±0.012 | — |
| GraphSAGE [102] | 0.953±0.001 | 0.501±0.013 | 0.637±0.006 | **0.634±0.006** |
| ClusterGCN [49] | 0.954±0.001 | 0.481±0.005 | 0.875±0.004 | 0.609±0.005 |
| GraphSAINT [280] | 0.966±0.001 | 0.511±0.001 | **0.981±0.004** | **0.653±0.003** |
| S-GCN [258] | 0.949±0.000 | 0.502±0.001 | 0.892±0.015 | 0.358±0.006 |
| SIGN (Ours) | **0.968±0.000** | **0.514±0.001** | 0.970±0.003 | 0.631±0.003 |
| $(p, s, t)$ | (4, 2, 0) | (4, 0, 1) | (2, 0, 1) | (2, 0, 1) |

TRANSDUCTIVE    *SIGN* obtains state-of-the-art results on the `ogbn-papers100M` dataset ( Tab. 3.7), outperforming other sampling-free methods by at least 1.8%. This shows that *SIGN* can scale to massive graphs while retaining ample expressivity. Sampling-based methods have not been scaled yet to this benchmark. In `ogbn-papers100M` only $\sim 1.35\%$ of nodes are labeled; at *each* training and inference iteration sampling-based methods would need to perform computation on subgraphs where the majority of nodes are unlabeled and thus do not contribute to the computation of loss and evaluation metrics. On the contrary, *SIGN* only processes the required labeled nodes given that the graph has already been employed during the one-time pre-computation phase, thus avoiding this redundant computation and memory usage.

`ogbn-products` results are reported in Tab. 3.5. *SIGN* outperforms all other sampling-free methods by at least 2.7%. However, contrary to the inductive benchmarks, sampling methods outperform *SIGN* and appear to generally be more suitable to this dataset. We hypothesise that, on this particular task, sampling may implicitly act as a regularizer, making these methods generalize better to the held-out test set, which in this dataset is sampled from a different distribution w.r.t. training and validation nodes [112]. This phenomenon, as well as its connection to the DropEdge method [199] and the bottleneck problem [6], would be the object of further investigation.

RUNTIME    While performing on par or better than state-of-the-art methods on most benchmarks in terms of accuracy, our method has the advantage of being significantly faster than other methods for large graphs. We perform comprehensive timing comparisons on `ogbn-products` and `Wikipedia` datasets and report average training, inference, and preprocessing times in Tab. 3.3. For these experiments, we run the implementations of *ClusterGCN* and *GraphSAINT* provided in the OGB code repository[1].

We use these datasets rather than `ogbn-papers100M` so we can compare to *ClusterGCN* and *GraphSAINT*. For completeness we report, however, that on `ogbn-papers100M` the best performing *SIGN(3,3,3)* model completes one evaluation pass on the validation set in $1.99 \pm 0.05$ seconds and on the test set in $3.34 \pm 0.04$ seconds (statistics are estimated over 10 runs and include the time required by device data transfers and by the computation of evalution metric).

Our model is faster than *ClusterGCN* and of comparable speed w.r.t. *GraphSAINT* in training[2], while being by far the fastest approach in inference: all *SIGN* architectures are always at least one order of magnitude faster than other methods, with the largest one (8 operators) requiring no more than 30 seconds to perform inference on over 12M nodes. *SIGN*'s preprocessing is slightly longer than other methods, but we notice that most of the calculations can be cast as sparse matrix multiplications and easily parallelized with frameworks for distributed computing. We envision engineering faster and even more scalable *SIGN* preprocessing implementations in future developments of this work. Finally, in order to also study the convergence behaviour of our proposed model, in Fig. 3.3 we plot the validation performance on `ogbn-products` from the start of the training as a function of run time for *ClusterGCN*, *GraphSAINT* and several *SIGN* configurations. We observe that *SIGN* does not only converge to a better validation accuracy than other methods but also exhibits much faster convergence than *ClusterGCN* and comparable speed to *GraphSAINT*.

---

[1] https://github.com/snap-stanford/ogb/tree/master/examples/nodeproppred/products
[2] Traning time is measured as forward-backward time to complete one epoch.

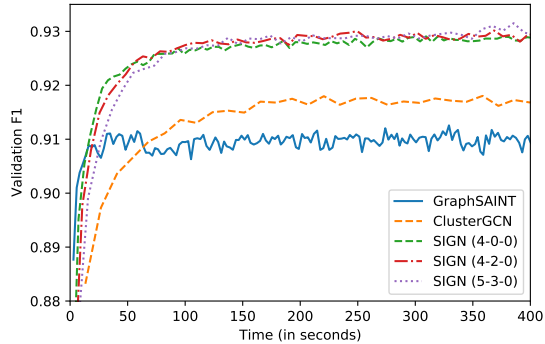

Figure 3.3: Convergence of different methods on `ogbn-products`.

Table 3.5: Performance on `ogbn-products`. SIGN($p$,$s$,$t$) refers to a configuration using $p$, $s$, and $t$ powers of simple, PPR-based, and triangle-based adjacency matrices.

|  | Training | Validation | Test |
| --- | --- | --- | --- |
| MLP | 84.03±0.93 | 75.54±0.14 | 61.06±0.08 |
| Node2Vec | 93.39±0.10 | 90.32±0.06 | 72.49±0.10 |
| S-GCN($L$=5) | 92.54±0.09 | 91.38±0.07 | 74.87±0.25 |
| ClusterGCN | 93.75±0.13 | 92.12±0.09 | 78.97±0.33 |
| GraphSAINT | 92.71±0.14 | 91.62±0.08 | **79.08±0.24** |
| SIGN(3,0,0) | 96.21±0.31 | 92.99±0.05 | 76.52±0.14 |
| SIGN(3,0,1) | 96.46±0.29 | 92.93±0.04 | 75.73±0.20 |
| SIGN(3,3,0) | 96.87±0.23 | 93.02±0.04 | 77.13±0.10 |
| SIGN(5,0,0) | 95.99±0.69 | 92.98±0.18 | 76.83±0.39 |
| SIGN(5,3,0) | **96.92±0.46** | **93.10±0.08** | 77.60±0.13 |

ABLATION STUDY How do different operator combinations affect *SIGN* performance? Results obtained with different choices of operators and their powers are reported in Tab. 3.5, Tab. 3.7 and Tab. 3.6 for, respectively, the transductive `ogbn-products` and `ogbn-papers100M` and inductive datasets. We notice that the best performance is obtained on each benchmark by a specific combination of operators, remarking the fact that each dataset features particular topological and content characteristics requiring suitable filters. Interestingly, we also observe that while PPR operators do not bring significant improvements in the inductive setting (being even harmful in certain cases), they are beneficial on the transductive `ogbn-products`. This finding is in accordance with [138], where the effectiveness of PPR diffusion operators in transductive settings has been extensively studied. Finally, we notice promising results attained in `Flickr` and `PPI` inductive settings by pairing standard adjacency matrices with a triangle-induced one. Studying the effect of operators induced by more complex network motifs is left for future research.

Table 3.6: Impact of various operator combinations on inductive datasets. Best results are in bold.

|  | Reddit | Flickr | PPI | Yelp |
|---|---|---|---|---|
| SIGN(2,0,0) | 0.966±0.003 | 0.503±0.003 | 0.965±0.002 | 0.623±0.005 |
| SIGN(2,0,1) | 0.966±0.000 | 0.510±0.001 | **0.970±0.003** | **0.631±0.003** |
| SIGN(2,2,0) | 0.967±0.000 | 0.495±0.002 | 0.964±0.003 | 0.617±0.005 |
| SIGN(4,0,0) | 0.967±0.000 | 0.508±0.001 | 0.959±0.002 | 0.623±0.004 |
| SIGN(4,0,1) | 0.967±0.000 | **0.514±0.001** | 0.965±0.003 | 0.623±0.004 |
| SIGN(4,2,0) | **0.968±0.000** | 0.500±0.001 | 0.930±0.010 | 0.618±0.004 |
| SIGN(4,2,1) | 0.967±0.000 | 0.508±0.002 | 0.969±0.001 | 0.620±0.004 |

Table 3.7: Results on `ogbn-papers100M`, the largest public graph dataset with over 110 million nodes. SIGN($p,d,f$) refers to a configuration using $p$, $d$, and $f$ powers of simple undirected, directed, and directed-transposed adjacency matrices.

|  | Training | Validation | Test |
|---|---|---|---|
| MLP | 54.84±0.43 | 49.60±0.29 | 47.24±0.31 |
| Node2Vec | — | 55.60±0.23 | 58.07±0.28 |
| S-GCN($L$=3) | 67.54±0.43 | 66.48±0.20 | 63.29±0.19 |
| SIGN(3,0,0) | 70.18±0.37 | 67.57±0.14 | 64.28±0.14 |
| SIGN(3,1,1) | 72.24±0.32 | 67.76±0.09 | 64.39±0.18 |
| SIGN(3,3,3) | **73.94±0.72** | **68.6±0.04** | **65.11±0.14** |

## 3.5 Discussion

DEPTH VS. WIDTH FOR GRAPH NEURAL NETWORKS   Our results have shown that it is possible to obtain competitive – and often state-of-the-art – results with one single graph convolutional layer and, hence, a shallow architecture. An important question is, therefore, when one should apply deep architectures to graphs, where by 'depth' we refer to the number of stacked graph convolutional layers. Deep Graph Neural Networks are notoriously hard to train due to vanishing gradients and feature smoothing [137, 149, 259], and, although recent works have shown that these issues can be addressed to some extent [93, 147, 199, 265, 288], yet extensive experiments conducted in [199] showed that depth often does not bring any significant gain in performance w.r.t. to shallow baselines. A promising direction for future investigation is, rather than 'going deep', to 'go wide', in the sense of exploring more expressive local operators. We believe this to be especially crucial in all those settings where scalability is paramount, such as in industrial large-scale systems.

EXTENSIONS   In our experiments, triangle-based operators showed promising results. Possible extensions can employ operators that account for higher-order structures such as simplicial com-plexes [12], paths [73], or motifs [174] that can be tailored to the specific problem. Furthermore, temporal information can be integrated e.g. in the form of temporal motifs [187].

LIMITATIONS   While our method relies on linear graph aggregation operations of the form $\mathbf{BX}$ for efficient precomputation, it is possible to make the diffusion operator dependent on the node features (and edge features, if available). In particular, graph attention [244] and similar mechanisms [171] use $\mathbf{B}_\theta(\mathbf{X})$, where $\theta$ are learnable parameters. The limitation is that such operators preclude efficient precomputation, which is key to the efficiency of our approach. Attention can be implemented in our scheme by training on a small subset of the graph to first determine the attention parameters, then fixing them to precompute the diffusion operator that is used during training and inference.

## 3.6 CONCLUSION

In this paper we presented *SIGN*, a sampling-free Graph Neural Network model that is able to easily scale to gigantic graphs while retaining enough expressive power to attain competitive results in all large-scale graph learning benchmarks. *SIGN* attains state-of-the-art results on many of them, including the massive `ogbn-papers100M`, currently the largest publicly available node-classification benchmark with $\sim 111\mathrm{M}$ nodes and $\sim 1.6\mathrm{B}$ edges. Our experiments have further demonstrated that the ability of our model to flexibly incorporate diverse, possibly domain-specific, operators is crucial to overcome the expressivity limitations of other sampling-free scalable models such as *S-GCN*, which *SIGN* has constantly outperformed over all datasets. Overall, our architecture achieves an optimal trade-off between simplicity and expressiveness; as it has shown to attain competitive results with fast training and inference, it represents the most suitable architecture for scalable applications to web-scale graphs.

FAST FORWARD TO TODAY   Since the publishing of our original paper [202], several extensions of SIGN have been proposed. For instance, FSGNN [168] incorporates additional normalization techniques, while both SAGN [228] and GAMLP [286] employ self-attention mechanisms on the node feature vectors. Additionally, NARS [275] and SeHGNN [267] extend SIGN to accommodate heterogeneous graphs, while Cini et al. [54] propose an extension for spatio-temporal graphs.

One of SIGN's most compelling contributions is the insight that decoupling graph propagation from model training can yield significant advantages in terms of scalability and simplicity, with negligible impact on predictive performance. This characteristic makes SIGN an excellent fit for production environments, as evidenced by its current deployment within AirBnB's trust and safety predictive services [223]. On large-scale node-classification benchmarks such as those provided by OGB [112] (e.g., ogbn-products, ogbn-mag, ogbn-papers100m), methods following the SIGN paradigm continue to achieve leading results. However, it's worth noting that SIGN-based methods have underperformed in the OGB large-scale heterogeneous node-classification dataset ogbn-mag240m [111]. Here, message-passing models like UniMP [216], in combination with neighbour sampling techniques [102], have taken the lead. Regrettably, as NARS and SeHGNN have not yet been evaluated on this specific challenge, their potential accuracy remains an open question. Consequently, the question of whether sampling methods like Graph-SAGE or preprocessing methods like SIGN offer the best scalability in Graph Neural Networks is still open for debate and likely dependent on the specific application.

An alternative scaling approach involving knowledge distillation has also been explored [285]. Specifically, a GNN model is trained and subsequently distilled into a simpler Multi-Layer Perceptron (MLP) for use during inference. However, this approach has inherent limitations. For example, the distilled MLP can only access node features; it cannot differentiate between nodes with identical features but different local graph structures, leading to potentially inaccurate predictions. Additionally, the necessity of having the graph during training renders this method slower and more complex than SIGN.

Another avenue of research has focused on efficiently reusing historical node embeddings. These embeddings, obtained from previous batches, are leveraged to prune computational sub-trees within the graph [44, 55]. Among these, PyGAS [71] stands out as the most comprehensive, offering a user-friendly framework built atop PyTorch-Geometric [72].

Finally, while general GNN encoders such as SIGN shine in node classification, link prediction tasks have been recently shown to require specialized models which are capable of capturing structural features around specific edges [148, 271, 284]. A prominent class of model addressing this are subgraph methods [283, 284], which construct a subgraph around a target edge, using then a GNN to predict its presence. However, these subgraph methods carry an extremely high computational cost, as they require a GNN forward pass for each possible edge in the graph, preventing their use for medium to large graphs. Motivated by these findings and our hands-on experience with SIGN, we recently developed a scalable model for link prediction [41]. Similar to SIGN's philosophy, this approach separates graph propagation from model training. Its novelty lies in leveraging hash functions to precompute structural features around an edge. Our method, impressively, mirrors the performance metrics of subgraph methods while being up to 1000x faster during inference, enabling accurate link prediction on large graphs for the first time.

# Appendices

## 3.A Datasets

### 3.A.1 Inductive datasets

`Reddit` [102] and `Flickr` [280] are multiclass classification problems, `Yelp` [280] and `PPI` [299] are multilabel classification instances. In `Reddit`, the task is to predict communities of online posts based on user comments. In `Flickr` the task is image categorization based on the description and common properties of online images. In `Yelp` the objective is to predict business attributes based on customer reviews; the task of `PPI` consists in predicting protein functions from the interactions of human tissue proteins. Further details on the generation of the `Yelp` and `Flickr` datasets can be found in [280].

### 3.A.2 Transductive dataset

`ogbn-products` [112] represents an Amazon product co-purchasing network [27] where the task is to predict the category of a product in a multi-class classification setup. Dataset splitting is not random, sales ranking (popularity) is instead used to split nodes into training/validation/test. Top 10% products in the ranking are assigned to the training set, next top 2% to validation and the remaining 88% of products are for testing.

`ogbn-papers100M` [112] represents a directed citation network of $\sim$ 111 million academic papers, where the task is to leverage information from the entire citation network to infer the labels (subject areas) of a smaller subset of ArXiv papers. The splitting strategy is time-based. Specifically, the training nodes (with labels) are all ArXiv papers published until 2017, validation nodes are ArXiv papers published in 2018 and test nodes are ArXiv papers published since 2019.

### 3.A.3 Wikipedia

`Wikipedia` links is a large-scale directed network of links between articles in the English version of Wikipedia. For the sake of our timing experiments the network has been turned into undirected. Node features have been randomly generated with a dimensionality of 100 as in `ogbn-products`.

## 3.B Model Selection and Hyperparameter Tuning

Tuning involved the following architectural and optimization hyperparameters: weight decay, dropout rate, batch size, learning rate, number of feedforward layers and units both in inception and classification modules. For each inductive experiment we chose the set of hyperparameters matching the best average validation loss calculated over 5 runs. For the the transductive setting we kept, instead, the set of hyperparameters with minimum validation loss over a single run. The hyperparameter search space for the inductive setting and grid for the transductive one are described in Tab. 3.B.1. The estimated hyperparameters for each best *SIGN* configuration are reported in Tab. 3.B.2 for inductive datasets and Tab. 3.B.3 for the transductive ones.

Table 3.B.1: Hyperparameter search space/grid. Ranges in the form [*low*, *high*] and sampling distributions. *Inception Layers* and *Classification Layers* are the number of feedforward layers in the representation part of the model (replacing Θ) and the classification part of the model (replacing Ω) respectively. The only exception is represented by *Yelp*, for which the Ω module was kept shallow (no hidden layers) to allow for lighter training and the left bounds on the dropout, learning rate and batch size intervals were lowered to, respectively, 0.0, 0.00001 and 60.

|  | Transductive | Inductive |  |
| --- | --- | --- | --- |
| Hyperparameter | Values | Space | Distribution |
| *Learning Rate* | 0.0001, 0.001 | [0.0001, 0.0025] | Uniform |
| *Batch Size* | 4096, 8192, 16384 | [128, 2048] | Quantized Uniform |
| *Dropout* | 0.5 | [0.2, 0.8] | Uniform |
| *Weight Decay* | 0.0, 0.00001 | [0, 0.0001] | Uniform |
| *Inception Layers* | 1 | 1, 2 | — |
| *Inception Units* | 256, 512 | [128, 512] | Quantized Uniform |
| *Classification Layers* | 1 | 1, 2 | — |
| *Classification Units* | 256, 512 | [512, 1024] | Quantized Uniform |
| *Activation* | PReLU | ReLU, PReLU | — |

Table 3.B.2: Hyperparameters chosen for the best configuration of SIGN on inductive datasets.

| Hyperparameter | Reddit | Flickr | PPI | Yelp |
| --- | --- | --- | --- | --- |
| Learning Rate | 0.000123 | 0.00172 | 0.00144 | 0.00005 |
| Dropout | 0.707 | 0.761 | 0.309 | 0.05 |
| Weight Decay | 9.18e-05 | 9.42e-05 | 3.26e-06 | 4.45e-07 |
| Batch Size | 830 | 330 | 210 | 90 |
| Inception Layers | 1 | 2 | 2 | 2 |
| Inception Units | 460 | 465 | 315 | 320 |
| Classification Layers | 1 | 1 | 2 | 0 |
| Classification Units | 675 | 925 | 870 | — |
| Activation | ReLU | PReLU | ReLU | ReLU |

Table 3.B.3: Hyperparameters chosen for the best configuration of SIGN on `ogbn-product` and those used on the `ogbn-papers100M` dataset.

| Hyperparameter | ogbn-products | ogbn-papers100M |
| --- | --- | --- |
| Learning Rate | 0.0001 | 0.001 |
| Dropout | 0.5 | 0.1 |
| Weight Decay | 0.0001 | 0.0 |
| Batch Size | 4096 | 256 |
| Inception Layers | 1 | 1 |
| Inception Units | 512 | 256 |
| Classification Layers | 1 | 3 |
| Classification Units | 512 | 256 |
| Activation | PReLU | ReLU |

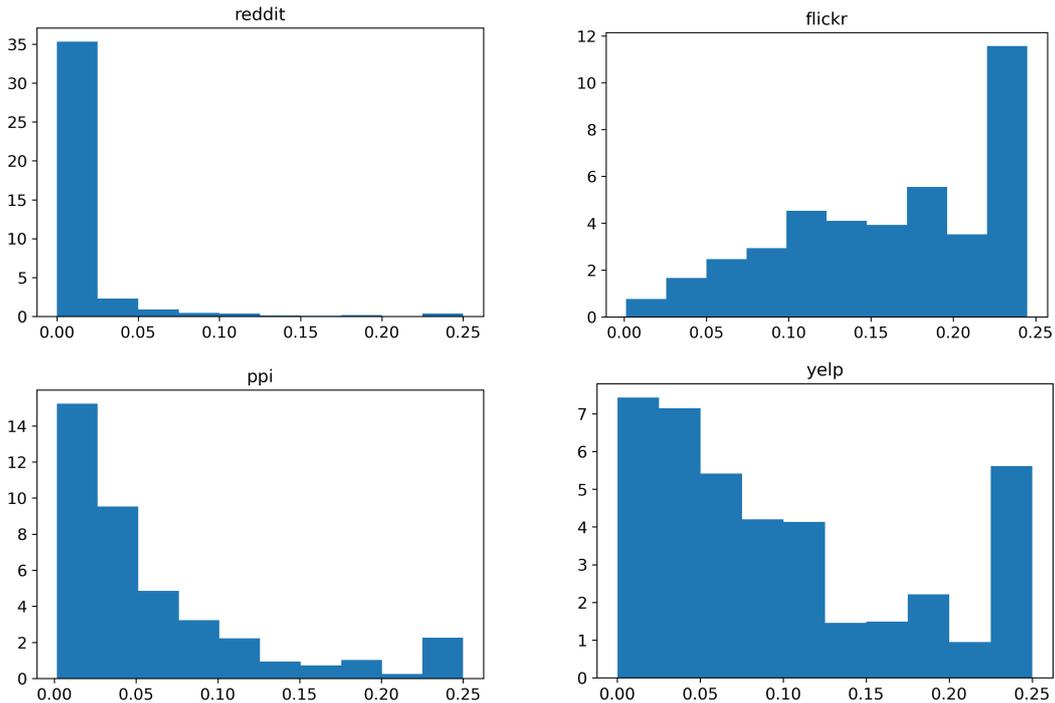

Figure 3.C.1: Normalized frequency distributions for row-wise variations on the diffusion weights of triangle operators over inductive datasets. Variations are measured as the standard deviation on the weight value over original neighborhoods from the test graph.

## 3.C  Triangle-based Operators

The triangle operator encodes the concept of *homophily* with a stronger acceptation with respect to the adjacency matrix: two nodes are connected by an edge only if they are both part of the same closed triad, i.e. if they are connected together and are both connected to the same node. Edge weights are proportional to the amount of triangles an edge belongs to, and they are normalised row-wise so to represent, for each node in a neighbourhood, its relative importance with respect to all the other neighbors.

This brings us to two considerations: first of all, the triangle operator is not carrying information related to nodes which were not already in the neighborhood. Secondly, it emphasizes the connections with those neighbors which are more related to our source node in virtue of the relationship described above. We can thus envision this operator being more useful in those graphs where this kind of relationship can be more discriminative within a neigborhood.

To verify this, in Fig. 3.C.1 we plot the normalized frequency distribution of intra-neighborhood standard deviation for the weights of triangle operators. It is interesting to notice the significantly different trends characterizing `Flickr` and `Reddit`, two datasets where triangle operators have experimentally brought, respectively, relative large and small performance improvement. `Flickr` tends to exhibit larger weight variations than other datasets, while, on the contrary, `Reddit` is the dataset where the smallest intra-neighborhood variation is observed. This suggests how, in `Flickr`, the triangle operator is able to restrict feature aggregation to a subset of the original neighbors –those co-occurring in the larger number of triangles– while in `Reddit` it mostly boils down to uniform averaging, making this operator not much more expressive than a simple adjacency matrix.

For replicability we report that, in the computation of triangle operators for PPI, we retained the self-loops already present in the original dataset. Investigations on how the presence of these edges affects the expressiveness of the triangle operator are left for future work.

# 4 Temporal Graph Neural Networks

"Time flies over us, but leaves its shadow behind." -
Nathaniel Hawthorne

## 4.1 Introduction

The majority of methods for deep learning on graphs assume the underlying graph to be *static*. However, most real-life systems of interactions such as social networks or biological interactomes are *dynamic* in nature, as they have evolving features and connectivity over time. While it is often possible to apply static graph deep learning models [153] to dynamic graphs by ignoring the temporal evolution, this has been shown to be sub-optimal; often, it is the dynamic structure that contains crucial insights about the system [263]. Learning on dynamic graphs is relatively recent, and most works are limited to the setting of discrete-time dynamic graphs represented as a sequence of snapshots of the graph [67, 153, 188, 208, 274, 277]. Such approaches are unsuitable for interesting real-world settings such as social networks, where dynamic graphs are continuous (i.e. edges can appear at any time) and evolving (i.e. new nodes join the graph continuously). Only recently, several approaches have been proposed that support the continuous-time scenario [16, 142, 158, 181, 238, 263].

Contributions. In this chapter, we first propose the generic inductive framework of Temporal Graph Networks (TGNs) operating on continuous-time dynamic graphs represented as a sequence of events, and show that many previous methods are specific instances of TGNs. Second, we propose a novel training strategy that allows the model to learn from the sequentiality of the data while maintaining highly efficient parallel processing. Third, we perform a detailed ablation study of different components of our framework and analyze the tradeoff between speed and accuracy. Finally, we show state-of-the-art performance on multiple tasks and datasets in both transductive and inductive settings, while being much faster than previous methods.

## 4.2 Background

Dynamic Graphs. There exist two main models for dynamic graphs. *Discrete-time dynamic graphs* (DTDG) are sequences of static graph snapshots taken at intervals in time. *Continuous-time dynamic graphs* (CTDG) are more general and can be represented as timed lists of events, which may include edge addition or deletion, node addition or deletion and node or edge feature transformations.

Our temporal (multi-)graph is modelled as a sequence of time-stamped *events* $\mathcal{G} = \{x(t_1), x(t_2), \ldots\}$, representing addition or change of a node or interaction between a pair of nodes at times $0 \leqslant t_1 \leqslant t_2 \leqslant \ldots$. An event $x(t)$ can be of two types: 1) A **node-wise event** is represented by $\mathbf{v}_i(t)$, where $i$ denotes the index of the node and $\mathbf{v}$ is the vector attribute associated with the event. If the index $i$ has not been seen before, the event creates node $i$ (with the given features).

*4 Temporal Graph Neural Networks*

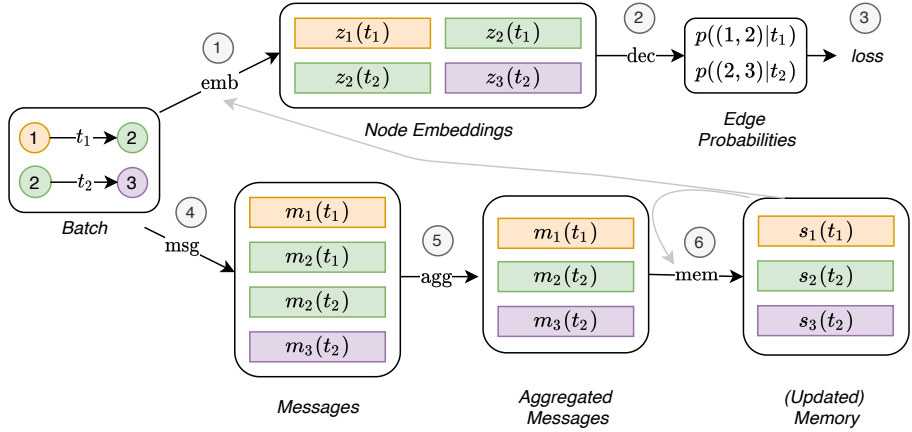

Figure 1: Computations performed by TGN on a batch of time-stamped interactions. *Top:* embeddings are produced by the embedding module using the temporal graph and the node's memory (1). The embeddings are then used to predict the batch interactions and compute the loss (2, 3). *Bottom:* these same interactions are used to update the memory (4, 5, 6). This is a simplified flow of operations, which would prevent the training of all the modules at the bottom as they would not receive a gradient. Sec. 4.3.2 explains how to change the flow of operations to solve this problem and Fig. 2 shows the complete diagram.

Otherwise, it updates the features. 2) An **interaction event** between nodes $i$ and $j$ is represented by a (directed) *temporal edge* $\mathbf{e}_{ij}(t)$ (there might be more than one edge between a pair of nodes, so technically this is a multigraph). We denote by $\mathcal{V}(T) = \{i : \exists \mathbf{v}_i(t) \in \mathcal{G}, t \in T\}$ and $\mathcal{E}(T) = \{(i,j) : \exists \mathbf{e}_{ij}(t) \in \mathcal{G}, t \in T\}$ the temporal set of vertices and edges, respectively, and by $\mathcal{N}_i(T) = \{j : (i,j) \in \mathcal{E}(T)\}$ the neighborhood of node $i$ in time interval $T$. $\mathcal{N}_i^k(T)$ denotes the $k$-hop neighborhood. A *snapshot* of the temporal graph $\mathcal{G}$ at time $t$ is the (multi-)graph $\mathcal{G}(t) = (\mathcal{V}[0,t], \mathcal{E}[0,t])$ with $n(t)$ nodes.

## 4.3 TEMPORAL GRAPH NETWORKS

Following the terminology in [129], a neural model for dynamic graphs can be regarded as an encoder-decoder pair, where an encoder is a function that maps from a dynamic graph to node embeddings, and a decoder takes as input one or more node embeddings and makes a task-specific prediction e.g. node classification or edge prediction. The key contribution of this paper is a novel Temporal Graph Network (TGN) encoder applied on a continuous-time dynamic graph represented as a sequence of time-stamped events and producing, for each time $t$, the embedding of the graph nodes $\mathbf{Z}(t) = (\mathbf{z}_1(t), \ldots, \mathbf{z}_{n(t)}(t))$.

### 4.3.1 CORE MODULES

MEMORY. The memory (state) of the model at time $t$ consists of a vector $\mathbf{s}_i(t)$ for each node $i$ the model has seen so far. The memory of a node is updated after an event (e.g. interaction with another node or node-wise change), and its purpose is to represent the node's history in a compressed format. Thanks to this specific module, TGNs have the capability to memorize long term dependencies for each node in the graph. When a new node is encountered, its memory is initialized as the zero vector, and it is then updated for each event involving the node, even after the model has finished training. While a global (graph-wise) memory can also be added to the model to track the evolution of the entire network, we leave this as future work.

MESSAGE FUNCTION. For each event involving node $i$, a message is computed to update $i$'s memory. In the case of an interaction event $\mathbf{e}_{ij}(t)$ between source node $i$ and target node $j$ at time $t$, two messages can be computed:

$$\mathbf{m}_i(t) = \mathrm{msg}_\mathrm{s}\big(\mathbf{s}_i(t^-), \mathbf{s}_j(t^-), \Delta t, \mathbf{e}_{ij}(t)\big), \quad \mathbf{m}_j(t) = \mathrm{msg}_\mathrm{d}\big(\mathbf{s}_j(t^-), \mathbf{s}_i(t^-), \Delta t, \mathbf{e}_{ij}(t)\big) \tag{4.1}$$

Similarly, in case of a node-wise event $\mathbf{v}_i(t)$, a single message can be computed for the node involved in the event:

$$\mathbf{m}_i(t) = \mathrm{msg}_\mathrm{n}\big(\mathbf{s}_i(t^-), t, \mathbf{v}_i(t)\big). \tag{4.2}$$

Here, $\mathbf{s}_i(t^-)$ is the memory of node $i$ just before time $t$ (i.e., from the time of the previous interaction involving $i$) and $\mathrm{msg}_\mathrm{s}$, $\mathrm{msg}_\mathrm{d}$ and $\mathrm{msg}_\mathrm{n}$ are learnable message functions, e.g. MLPs. In all experiments, we choose the message function as *identity* (id), which is simply the concatenation of the inputs, for the sake of simplicity. Deletion events are also supported by the framework and are presented in Sec. 4.A. A more complex message function that involves additional aggregation from the neighbours of nodes $i$ and $j$ is also possible and is left for future study.

MESSAGE AGGREGATOR. Resorting to batch processing for efficiency reasons may lead to multiple events involving the same node $i$ in the same batch. As each event generates a message in our formulation, we use a mechanism to aggregate messages $\mathbf{m}_i(t_1), \ldots, \mathbf{m}_i(t_b)$ for $t_1, \ldots, t_b \leqslant t$,

$$\bar{\mathbf{m}}_i(t) = \mathrm{agg}(\mathbf{m}_i(t_1), \ldots, \mathbf{m}_i(t_b)). \tag{4.3}$$

Here, agg is an aggregation function. While multiple choices can be considered for implementing this module (e.g. RNNs or attention w.r.t. the node memory), for the sake of simplicity we considered two efficient non-learnable solutions in our experiments: *most recent message* (keep only the most recent message for a given node) and *mean message* (average all messages for a given node). We leave learnable aggregation as a future research direction.

MEMORY UPDATER. As previously mentioned, the memory of a node is updated upon each event involving the node itself:

$$\mathbf{s}_i(t) = \mathrm{mem}\big(\bar{\mathbf{m}}_i(t), \mathbf{s}_i(t^-)\big). \tag{4.4}$$

For interaction events involving two nodes $i$ and $j$, the memories of both nodes are updated after the event has happened. For node-wise events, only the memory of the related node is updated. Here, mem is a learnable memory update function, e.g. a recurrent neural network such as LSTM [107] or GRU [51].

EMBEDDING. The embedding module is used to generate the temporal embedding $\mathbf{z}_i(t)$ of node $i$ at any time $t$. The main goal of the embedding module is to avoid the so-called memory staleness problem [129]. Since the memory of a node $i$ is updated only when the node is involved in an event, it might happen that, in the absence of events for a long time (e.g. a social network user who stops using the platform for some time before becoming active again), $i$'s memory becomes stale. While multiple implementations of the embedding module are possible, we use the form:

$$\mathbf{z}_i(t) = \mathrm{emb}(i, t) = \sum_{j \in \mathcal{N}_i^k([0,t])} h(\mathbf{s}_i(t), \mathbf{s}_j(t), \mathbf{e}_{ij}, \mathbf{v}_i(t), \mathbf{v}_j(t)),$$

where $h$ is a learnable function. This includes many different formulations as particular cases:

*Identity* (id): $\text{emb}(i, t) = \mathbf{s}_i(t)$, which uses the memory directly as the node embedding.

*Time projection* (time): $\text{emb}(i, t) = (1 + \Delta t\, \mathbf{w}) \circ \mathbf{s}_i(t)$, where $\mathbf{w}$ are learnable parameters, $\Delta t$ is the time since the last interaction, and $\circ$ denotes element-wise vector product. This version of the embedding method was used in Jodie [142].

*Temporal Graph Attention* (attn): A series of $L$ graph attention layers compute $i$'s embedding by aggregating information from its $L$-hop temporal neighborhood.

The input to the $l$-th layer is $i$'s representation $\mathbf{h}_i^{(l-1)}(t)$, the current timestamp $t$, $i$'s neighbourhood representation $\{\mathbf{h}_1^{(l-1)}(t), \dots, \mathbf{h}_N^{(l-1)}(t)\}$ together with timestamps $t_1, \dots, t_N$ and features $\mathbf{e}_{i1}(t_1), \dots, \mathbf{e}_{iN}(t_N)$ for each of the considered interactions which form an edge in $i$'s temporal neighbourhood:

$$\mathbf{h}_i^{(l)}(t) = \text{MLP}^{(l)}(\mathbf{h}_i^{(l-1)}(t) \,\|\, \tilde{\mathbf{h}}_i^{(l)}(t)), \tag{4.5}$$

$$\tilde{\mathbf{h}}_i^{(l)}(t) = \text{MultiHeadAttention}^{(l)}(\mathbf{q}^{(l)}(t), \mathbf{K}^{(l)}(t), \mathbf{V}^{(l)}(t)), \tag{4.6}$$

$$\mathbf{q}^{(l)}(t) = \mathbf{h}_i^{(l-1)}(t) \,\|\, \boldsymbol{\phi}(0), \tag{4.7}$$

$$\mathbf{K}^{(l)}(t) = \mathbf{V}^{(l)}(t) = \mathbf{C}^{(l)}(t), \tag{4.8}$$

$$\mathbf{C}^{(l)}(t) = [\mathbf{h}_1^{(l-1)}(t) \,\|\, \mathbf{e}_{i1}(t_1) \,\|\, \boldsymbol{\phi}(t - t_1),\, \dots,\, \mathbf{h}_N^{(l-1)}(t) \,\|\, \mathbf{e}_{iN}(t_N) \,\|\, \boldsymbol{\phi}(t - t_N)]. \tag{4.9}$$

Here, $\boldsymbol{\phi}(\cdot)$ represents a generic time encoding [263], $\|$ is the concatenation operator and $\mathbf{z}_i(t) = \text{emb}(i, t) = \mathbf{h}_i^{(L)}(t)$. Each layer amounts to performing multi-head-attention [243] where the query ($\mathbf{q}^{(l)}(t)$) is a reference node (i.e. the target node or one of its $L-1$-hop neighbors), and the keys $\mathbf{K}^{(l)}(t)$ and values $\mathbf{V}^{(l)}(t)$ are its neighbors. Finally, an MLP is used to combine the reference node representation with the aggregated information. Differently from the original formulation of this layer (firstly proposed in TGAT [263]) where no node-wise temporal features were used, in our case the input representation of each node $\mathbf{h}_j^{(0)}(t) = \mathbf{s}_j(t) + \mathbf{v}_j(t)$ and as such it allows the model to exploit both the current memory $\mathbf{s}_j(t)$ and the temporal node features $\mathbf{v}_j(t)$.

*Temporal Graph Sum* (sum): A simpler and faster aggregation over the graph:

$$\mathbf{h}_i^{(l)}(t) = \mathbf{W}_2^{(l)}(\mathbf{h}_i^{(l-1)}(t) \,\|\, \tilde{\mathbf{h}}_i^{(l)}(t)), \tag{4.10}$$

$$\tilde{\mathbf{h}}_i^{(l)}(t) = \text{ReLu}(\sum_{j \in \mathcal{N}_i([0,t])} \mathbf{W}_1^{(l)}(\mathbf{h}_j^{(l-1)}(t) \,\|\, \mathbf{e}_{ij} \,\|\, \boldsymbol{\phi}(t - t_j))). \tag{4.11}$$

Here as well, $\boldsymbol{\phi}(\cdot)$ is a time encoding and $\mathbf{z}_i(t) = \text{emb}(i, t) = \mathbf{h}_i^{(L)}(t)$. In the experiment, both for the *Temporal Graph Attention* and for the *Temporal Graph Sum* modules we use the time encoding presented in Time2Vec [128] and used in TGAT [263].

The graph embedding modules mitigate the staleness problem by aggregating information from a node's neighbors memory. When a node has been inactive for a while, it is likely that some of its neighbours have been recently active, and by aggregating their memories, TGN can compute an up-to-date embedding for the node. The temporal graph attention is additionally able to select which neighbors are more important based on both features and timing information.

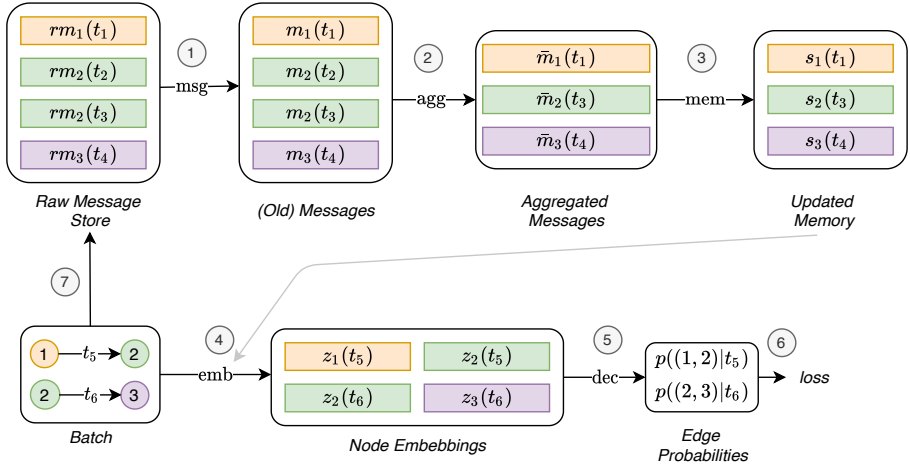

Figure 2: Flow of operations of TGN used to train the memory-related modules. *Raw Message Store* stores the necessary raw information to compute messages, i.e. the input to the message functions, which we call raw messages, for interactions which have been processed by the model in the past. This allows the model to delay the memory update brought by an interaction to later batches. At first, the memory is updated using messages computed from raw messages stored in previous batches (1, 2, 3). The embeddings can then be computed using the just updated memory (grey link) (4). By doing this, the computation of the memory-related modules directly influences the loss (5, 6), and they receive a gradient. Finally, the raw messages for this batch interactions are stored in the raw message store (6) to be used in future batches.

### 4.3.2 Training

TGN can be trained for a variety of tasks such as edge prediction (self-supervised) or node classification (semi-supervised). We use link prediction as an example: provided a list of time-ordered interactions, the goal is to predict future interactions from those observed in the past. Fig. 1 shows the computations performed by TGN on a batch of training data.

The complexity in our training strategy relates to the memory-related modules (*Message function*, *Message aggregator*, and *Memory updater*) because they do not directly influence the loss and therefore do not receive a gradient. To solve this problem, the memory must be updated before predicting the batch interactions. However, updating the memory with an interaction $\mathbf{e}_{ij}(t)$ before using the model to predict that same interaction, causes information leakage. To avoid the issue, when processing a batch, we update the memory with messages coming from previous batches (which are stored in the *Raw Message Store*), and then predict the interactions. Fig. 2 shows the training flow for the memory-related modules. Pseudocode for the training procedure is presented in Algorithm 1.

More formally, at any time $t$, the Raw Message Store contains (at most) one raw message $rm_i$ for each node $i$[1], generated from the last interaction involving $i$ before time $t$. When the model processes the next interactions involving $i$, its memory is updated using $rm_i$ (arrows 1, 2, 3 in Fig. 2), then the updated memory is used to compute the node's embedding and the batch loss (arrows 4, 5, 6). Finally, the raw messages for the new interaction are stored in the raw message store (arrows 7). It is also worth noticing that all predictions in a given batch have access to the same state of the memory. While from the perspective of the first interaction in the batch the memory is up-to-date (since it contains information about all previous interactions in the graph), from the perspective of the last interaction in the batch the same memory is out-of-date, since it lacks information about previous interactions in the same batch. This disincentives the use of a big batch size (in the extreme case where the batch size is a big as the dataset, all predictions would be made using the initial zero memory). We found a batch size of 200 to be a good trade-off between speed and update granularity.

---

[1]The Raw Message Store does not contain a message for $i$ only if $i$ has never been involved in an event in the past.

## 4.4 Related Work

Early models for learning on dynamic graphs focused on DTDGs. Such approaches either aggregate graph snapshots and then apply static methods [4, 5, 106, 118, 153, 214], assemble snapshots into tensors and factorize [67, 161, 276], or encode each snapshot to produce a series of embeddings. In the latter case, the embeddings are either aggregated by taking a weighted sum [268, 293], fit to time series models [99, 117, 175, 218], used as components in RNNs [46, 163, 179, 188, 208, 213, 274], or learned by imposing a smoothness constraint over time [70, 94, 100, 130, 193, 221, 268, 292, 297]. Another line of work encodes DTDGs by first performing random walks on an initial snapshot and then modifying the walk behaviour for subsequent snapshots [61, 66, 162, 261, 277]. Spatio-temporal graphs (considered by [151, 282] for traffic forecasting) are specific cases of dynamic graphs where the topology of the graph is fixed.

CTDGs have been addressed only recently. Several approaches use random walk models [16, 181, 182] incorporating continuous time through constraints on transition probabilities. Sequence-based approaches for CTDGs [142, 158, 237, 238] use RNNs to update representations of the source and destination node each time a new edge appears. Other recent works have focused on dynamic knowledge graphs [60, 79, 90, 262]. Many architectures for continuous-time dynamic graphs are based on a node-wise memory updated by an RNN when new interactions appear. Yet, they lack a GNN-like aggregation from a node's neighbors when computing its embedding, which makes them susceptible to the staleness problem (i.e. a node embedding becoming out of date) while at the same time also limiting their expressive power.

Most recent CTDGs learning models can be interpreted as specific cases of our framework (see Tab. 1). For example, Jodie [142] uses the time projection embedding module $\text{emb}(i, t) = (1 + \Delta t \mathbf{w}) \circ \mathbf{s}_i(t)$. TGAT [263] is a specific case of TGN when the memory and its related modules are missing, and graph attention is used as the Embedding module. DyRep [238] computes messages using graph attention on the destination node neighbors. Finally, we note that TGN generalizes the Graph Networks (GN) model [18] for static graphs (with the exception of the global block that we omitted from our model for the sake of simplicity), and thus the majority of existing graph message passing-type architectures.

## 4.5 Experiments

**Datasets.** We use three datasets in our experiments: Wikipedia, Reddit [142], and Twitter, which are described in detail in Sec. 4.C. Our experimental setup closely follows [263] and focuses on the tasks of future edge ('link') prediction and dynamic node classification. In future edge prediction, the goal is to predict the probability of an edge occurring between two nodes at a given time. Our encoder is combined with a simple MLP decoder mapping from the concatenation of two node embeddings to the probability of the edge. We study both the transductive and inductive settings. In the transductive task, we predict future links of the nodes observed during training, whereas in the inductive tasks we predict future links of nodes never observed before. For node classification, the transductive setting is used. For all tasks and datasets we perform the same 70%-15%-15% chronological split as in Xu et al. [263]. All the results were averaged over 10 runs. Hyperparameters and additional details can be found in Sec. 4.D.

Table 1: Previous models for deep learning on continuous-time dynamic graphs are specific case of our TGN framework. Shown are multiple variants of TGN used in our ablation studies. *method* ($l$,$n$) refers to graph convolution using $l$ layers and $n$ neighbors. †uses t-batches. * uses uniform sampling of neighbors, while the default is sampling the most recent neighbors. ‡message aggregation not explained in the paper. ∥ uses a summary of the destination node neighbourhood (obtained through graph attention) as additional input to the message function.

|            | Mem. | Mem. Updater | Embedding     | Mess. Agg. | Mess. Func. |
|------------|------|--------------|---------------|------------|-------------|
| Jodie      | node | RNN          | time          | —†         | id          |
| TGAT       | —    | —            | attn (2l, 20n)* | —        | —           |
| DyRep      | node | RNN          | id            | —‡         | attn∥       |
| TGN-attn   | node | GRU          | attn (1l, 10n) | last      | id          |
| TGN-2l     | node | GRU          | attn (2l, 10n) | last      | id          |
| TGN-no-mem | —    | —            | attn (1l, 10n) | —         | —           |
| TGN-id     | node | GRU          | id            | last       | id          |
| TGN-sum    | node | GRU          | sum (1l, 10n)  | last      | id          |
| TGN-mean   | node | GRU          | attn (1l, 10n) | mean      | id          |

BASELINES. Our strong baselines are state-of-the-art approaches for continuous time dynamic graphs (CTDNE [182], Jodie [142], DyRep [238] and TGAT [263]) as well as state-of-the-art models for static graphs (GAE [135], VGAE [135], DeepWalk [194], Node2Vec [98], GAT [244] and GraphSAGE [101]).

### 4.5.1 Performance

Tab. 2 presents the results on future edge prediction. Our model clearly outperforms the baselines by a large margin in both transductive and inductive settings on all datasets. The gap is particularly large on the Twitter dataset, where we outperform the second-best method (DyRep) by over 4% and 10% in the transductive and inductive case, respectively. Tab. 3 shows the results on dynamic node classification, where again our model obtains state-of-the-art results, with a large improvement over all other methods.

Due to the efficient parallel processing and the need for only one graph attention layer (see Sec. 4.5.2 for the ablation study on the number of layers), our model is up to $30\times$ faster than TGAT per epoch ( Fig. 3a), while requiring a similar number of epochs to converge.

### 4.5.2 Choice of Modules

We perform a detailed ablation study comparing different instances of our TGN framework, focusing on the speed vs accuracy tradeoff resulting from the choice of modules and their combination. The variants we experiment with are reported in Tab. 1 and their results are depicted in Fig. 3a.

MEMORY. We compare a model that does not make use of memory (TGN-no-mem), with a model which uses memory (TGN-attn) but is otherwise identical. While TGN-att is about $3\times$ slower, it has nearly 4% higher precision than TGN-no-mem, confirming the importance of memory for learning on dynamic graphs, due to its ability to store long-term information about a node which is otherwise hard to capture. This finding is confirmed in Fig. 3b where we compare different models when increasing the number of sampled neighbors: the models with memory consistently outperform the models without memory. Moreover, using the memory in conjunction

Table 2: Average Precision (%) for future edge prediction task in transductive and inductive settings. †Does not support inductive. *Static graph method.

|  | Wikipedia | | Reddit | | Twitter | |
| --- | --- | --- | --- | --- | --- | --- |
|  | Transductive | Inductive | Transductive | Inductive | Transductive | Inductive |
| GAE* | 91.44 ± 0.1 | † | 93.23 ± 0.3 | † | — | † |
| VAGE* | 91.34 ± 0.3 | † | 92.92 ± 0.2 | † | — | † |
| DeepWalk* | 90.71 ± 0.6 | † | 83.10 ± 0.5 | † | — | † |
| Node2Vec* | 91.48 ± 0.3 | † | 84.58 ± 0.5 | † | — | † |
| GAT* | 94.73 ± 0.2 | 91.27 ± 0.4 | 97.33 ± 0.2 | 95.37 ± 0.3 | 67.57 ± 0.4 | 62.32 ± 0.5 |
| GraphSAGE* | 93.56 ± 0.3 | 91.09 ± 0.3 | 97.65 ± 0.2 | 96.27 ± 0.2 | 65.79 ± 0.6 | 60.13 ± 0.6 |
| CTDNE | 92.17 ± 0.5 | † | 91.41 ± 0.3 | † | — | † |
| Jodie | 94.62 ± 0.5 | 93.11 ± 0.4 | 97.11 ± 0.3 | 94.36 ± 1.1 | 85.20 ± 2.4 | 79.83 ± 2.5 |
| TGAT | 95.34 ± 0.1 | 93.99 ± 0.3 | 98.12 ± 0.2 | 96.62 ± 0.3 | 70.02 ± 0.6 | 66.35 ± 0.8 |
| DyRep | 94.59 ± 0.2 | 92.05 ± 0.3 | 97.98 ± 0.1 | 95.68 ± 0.2 | 83.52 ± 3.0 | 78.38 ± 4.0 |
| TGN-attn (Ours) | **98.46** ± 0.1 | **97.81** ± 0.1 | **98.70** ± 0.1 | **97.55** ± 0.1 | **94.52** ± 0.5 | **91.37** ± 1.1 |

Table 3: ROC AUC % for the dynamic node classification. *Static graph method.

|  | Wikipedia | Reddit |
| --- | --- | --- |
| GAE* | 74.85 ± 0.6 | 58.39 ± 0.5 |
| VAGE* | 73.67 ± 0.8 | 57.98 ± 0.6 |
| GAT* | 82.34 ± 0.8 | 64.52 ± 0.5 |
| GraphSAGE* | 82.42 ± 0.7 | 61.24 ± 0.6 |
| CTDNE | 75.89 ± 0.5 | 59.43 ± 0.6 |
| JODIE | 84.84 ± 1.2 | 61.83 ± 2.7 |
| TGAT | 83.69 ± 0.7 | 65.56 ± 0.7 |
| DyRep | 84.59 ± 2.2 | 62.91 ± 2.4 |
| TGN-attn (Ours) | **87.81** ± 0.3 | **67.06** ± 0.9 |

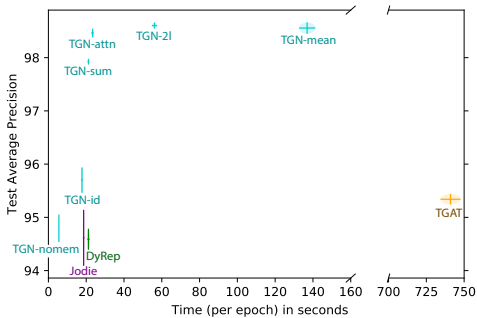

(a) Tradeoff between accuracy (test average precision in %) and speed (time per epoch in sec) of different models.

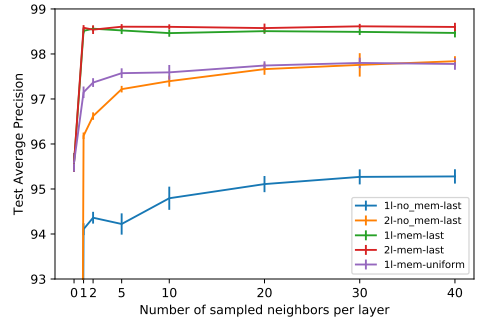

(b) Performance of methods with a varying number of layers and with or without memory when sampling an increasing number of neighbors. *Last* and *uniform* refer to neighbor sampling strategy.

Figure 3: Ablation studies on the Wikipedia dataset for the transductive setting of the future edge prediction task. Means and standard deviations were computed over 10 runs.

with sampling the most recent neighbors reduces the number of neighbors needed to achieve the best performance when used.

EMBEDDING MODULE.    Fig. 3a compare models with different embedding modules (TGN-id, TGN-attn, TGN-sum). The first interesting insight is that the ability to exploit the graph is crucial for performance: we note that all graph-based projections (TGN-attn, TGN-sum) outperform the graph-less TGN-id model by a large margin, with TGN-attn being the top performer at the expense of being only slightly slower than the simpler TGN-sum. This indicates that the ability to obtain more recent information through the graph, and to select which neighbors are the most important are critical factors for the performance of the model.

MESSAGE AGGREGATOR.    We compare two further models, one using the last message aggregator (TGN-attn) and another a mean aggregator (TGN-mean) but otherwise the same. While TGN-mean performs slightly better, it is more than $3\times$ slower.

NUMBER OF LAYERS.    While in TGAT having 2 layers is of fundamental importance for obtaining good performances (TGAT vs TGAT-1l has over 10% difference in average precision), in TGN the presence of the memory makes it enough to use just one layer to obtain very high performance (TGN-attn vs TGN-2l). This is because when accessing the memory of the 1-hop neighbors, we are indirectly accessing information from hops further away. Moreover, being able to use only one layer of graph attention speeds up the model dramatically.

## 4.6 CONCLUSION

We introduce TGN, a generic framework for learning on continuous-time dynamic graphs. We obtain state-of-the-art results on several tasks and datasets while being faster than previous methods. Detailed ablation studies show the importance of the memory and its related modules to store long-term information, as well as the importance of the graph-based embedding module to generate up-to-date node embeddings. We envision interesting applications of TGN in the fields of social sciences, recommender systems, and biological interaction networks, opening up a future research direction of exploring more advanced settings of our model and understanding the most appropriate domain-specific choices.

FAST-FORWARD TO TODAY    Since our original publication [200], TGNs have experienced significant advancements and found use in a myriad of applications. Notably, there's the enhancement of online inference efficiency achieved by decoupling it from graph propagation [249]. Additionally, advancements have been made in data augmentation [252], predicting higher-order temporal patterns [156], and scaling to billions of nodes through optimized data structures and algorithms designed for temporal neighbourhood sampling [291]. Delving deeper, Souza et al. [224] explored the expressivity boundaries of TGNs and combined them with temporal random walks [124, 250] to push past these limitations.

In terms of practical applications, TGNs and their subsequent extensions have been used in areas like fake news detection [222], molecule synthesis [245], and in predicting information pathways [125].

In a recent work, Poursafaei et al. [196] highlighted that the prevailing evaluation framework in temporal graph learning had its inadequacies, primarily due to the small size of datasets leading to performance saturation. To address this, they introduced more comprehensive datasets and a rigorous evaluation framework, showcasing that TGNs could maintain their state-of-the-art performance even three years after their inception. This revelation and its subsequent challenges sparked our collaboration, culminating in the development of TGB[115][2]—a benchmark geared for learning on temporal graphs, introducing large and diverse datasets under a rigorous, unified evaluation protocol.

---

[2] https://tgb-website.pages.dev/

# Appendices

## 4.A Deletion Events

The TGN frameworks also support edge and node deletion events. In the case of an **edge deletion** event $(i, j, t', t)$ where an edge between nodes $i$ and $j$ which was created at time $t'$ is deleted at time $t$, two messages can be computed for the source and target nodes that respectively started and received the interaction:

$$\mathbf{m}_i(t) = \mathrm{msg}_{s'}\big(\mathbf{s}_i(t^-), \mathbf{s}_j(t^-), \Delta t, \mathbf{e}_{ij}(t)\big), \quad \mathbf{m}_j(t) = \mathrm{msg}_{d'}\big(\mathbf{s}_j(t^-), \mathbf{s}_i(t^-), \Delta t, \mathbf{e}_{ij}(t)\big) \tag{4.12}$$

In case of a **node deletion** event, we simply remove the node (and its incoming and outgoing edges) from the temporal graph so that when computing other nodes embedding this node is not used during the temporal graph attention. Additionally, it would be possible to compute a message from the node's feature and memory and use it to update the memories of all its neighbors.

## 4.B TGN Training

When parallelizing the training of TGN, it is important to maintain the temporal dependencies between interactions. If the events on each node were independent, we could treat them as a sequence and train an RNN on each node independently using Back Propagation Through Time (BPTT). However, the graph structure introduces dependencies between the events (the update of a node depends on the current memory of other nodes), which prevents us from processing nodes in parallel. Previous methods either process the interactions one at a time, or use the t-batch [142] training algorithm, which however does not satisfy temporal consistency when aggregating from the graph as in the case of TGN (since the update does not only depend on the memory of the other node involved the interaction, but also on the neighbors of the two nodes).

These issues motivate our training algorithm, which processes all interactions in batches following chronological order. It stores the last message for each node in a message store, to process it before predicting the next interaction for the node. This allows the memory-related modules to receive a gradient. Algorithm 1 presents the pseudocode for TGN training, while Fig. 4.B.1 shows a schematic diagram of TGN.

## 4.C Datasets

Reddit and Wikipedia are bipartite interaction graphs. In the Reddit dataset, users and sub-reddits are nodes, and an interaction occurs when a user writes a post to the sub-reddit. In the Wikipedia dataset, users and pages are nodes, and an interaction represents a user editing a page. In both aforementioned datasets, the interactions are represented by text features (of a post or page edit,

---

[1] For the sake of clarity, we use the same message function for both sources and destination.
[2] We denote with $\mathrm{emb}_{\hat{s}}$ an embedding layer that operates on the updated version of the memory $\hat{\mathbf{s}}$.

**Algorithm 1** Training TGN

1: $\mathbf{s} \leftarrow \mathbf{0}$ ▷ Initialize memory to zeros
2: $\mathbf{m\_raw} \leftarrow \{\}$ ▷ Initialize raw messages
3: **for** each batch $(\mathbf{i}, \mathbf{j}, \mathbf{e}, \mathbf{t})$ in training data **do**
4: $\quad \mathbf{n} \leftarrow$ sample negatives
5: $\quad \mathbf{m} \leftarrow \text{msg}(\mathbf{m\_raw})$ ▷ Compute messages from raw features[1]
6: $\quad \bar{\mathbf{m}} \leftarrow \text{agg}(\mathbf{m})$ ▷ Aggregate messages for the same nodes
7: $\quad \hat{\mathbf{s}} \leftarrow \text{mem}(\bar{\mathbf{m}}, \mathbf{s})$ ▷ Get updated memory
8: $\quad \mathbf{z_i}, \mathbf{z_j}, \mathbf{z_n} \leftarrow \text{emb}_{\hat{\mathbf{s}}}(\mathbf{i}, \mathbf{t}), \text{emb}_{\hat{\mathbf{s}}}(\mathbf{j}, \mathbf{t}), \text{emb}_{\hat{\mathbf{s}}}(\mathbf{n}, \mathbf{t})$ ▷ Compute node embeddings[2]
9: $\quad \mathbf{p_{pos}}, \mathbf{p_{neg}} \leftarrow \text{dec}(\mathbf{z_i}, \mathbf{z_j}), \text{dec}(\mathbf{z_i}, \mathbf{z_n})$ ▷ Compute interactions probs
10: $\quad l = \text{BCE}(\mathbf{p_{pos}}, \mathbf{p_{neg}})$ ▷ Compute BCE loss
11: $\quad \mathbf{m\_raw_i}, \mathbf{m\_raw_j} \leftarrow (\hat{\mathbf{s}}_\mathbf{i}, \hat{\mathbf{s}}_\mathbf{j}, \mathbf{t}, \mathbf{e}), (\hat{\mathbf{s}}_\mathbf{j}, \hat{\mathbf{s}}_\mathbf{i}, \mathbf{t}, \mathbf{e})$ ▷ Compute raw messages
12: $\quad \mathbf{m\_raw} \leftarrow \text{store\_raw\_messages}(\mathbf{m\_raw}, \mathbf{m\_raw_i}, \mathbf{m\_raw_j})$ ▷ Store raw messages
13: $\quad \mathbf{s_i}, \mathbf{s_j} \leftarrow \hat{\mathbf{s}}_\mathbf{i}, \hat{\mathbf{s}}_\mathbf{j}$ ▷ Store updated memory for sources and destinations
14: **end for**

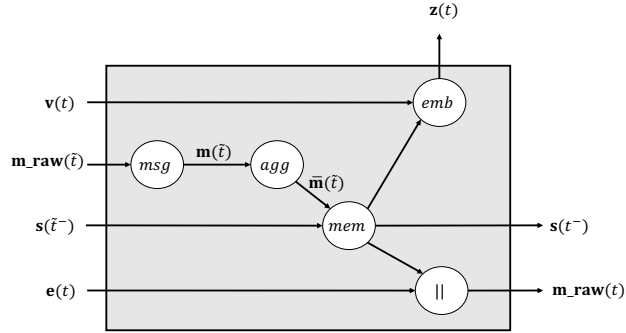

Figure 4.B.1: Schematic diagram of TGN. $\mathbf{m\_raw}(t)$ is the raw message generated by event $\mathbf{e}(t)$, $\tilde{t}$ is the instant of time of the last event involving each node, and $t^-$ the one immediately preceding $t$.

Table 4.C.1: Statistics of the datasets used in the experiments.

|  | Wikipedia | Reddit | Twitter |
| --- | --- | --- | --- |
| # Nodes | 9,227 | 11,000 | 8,861 |
| # Edges | 157,474 | 672,447 | 119,872 |
| # Edge features | 172 | 172 | 768 |
| # Edge features type | LIWC | LIWC | BERT |
| Timespan | 30 days | 30 days | 7 days |
| Chronological Split | 70%-15%-15% | 70%-15%-15% | 70%-15%-15% |
| # Nodes with dynamic labels | 217 | 366 | – |

respectively), and labels represent whether a user is banned. Both interactions and labels are time-stamped.

The Twitter dataset is a non-bipartite graph released as part of the 2020 RecSys Challenge [22]. Nodes are users and interactions are retweets. The features of an interaction are a BERT-based [257] vector representation of the text of the retweet.

Node features are not present in any of these datasets, and we therefore assign the same zero feature vector to all nodes. Moreover, While our framework is general and in Sec. 4.3.1 we showed how it can process any type of event, these three datasets only contain the edge creation (interaction) event type. Creating and evaluation of datasets with a wider variety of events is left as future work.

The statistics of the three datasets are reported in Tab. 4.C.1.

TWITTER DATASET GENERATION  To generate the Twitter dataset we started with the snapshot of the Recsys Challenge training data on 2020/09/06. We filtered the data to include only retweet edges (discarding other types of interactions) where the timestamp was present. This left approximately 10% of the edges in the original dataset. We then filtered the retweet multi-graph (users can be connected by multiple retweets) to only include the largest connected component. Finally, we filtered the graph to only the top 5,000 nodes in-degree and the top 5,000 by out-degree, ending up with 8,861 nodes since some nodes were in both sets.

## 4.D  Additional Experimental Settings and Results

HYPERPARAMETERS  For all datasets, we use the Adam optimizer with a learning rate of 0.0001, a batch size of 200 for both training, validation and testing and early stopping with patience of 5. We sample an equal amount of negatives to the positive interactions, and use *average precision* as reference metric. Additional hyperparameters used for both future edge prediction and dynamic node classification are reported in Tab. 4.D.1. For all the graph embedding modules we use neighbors sampling [101] (i.e. only aggregate from $k$ neighbors) since it improves the efficiency of the model without losing in accuracy. In particular, the sampled edges are the $k$ most recent ones, rather than the traditional approach of sampling them uniformly, since we found it to perform much better (see Fig. 4.D.1). All experiments and timings are conducted on an AWS p3.16xlarge machine and the results are averaged over 10 runs. The code will be made available for all our experiments to be reproduced.

BASELINES RESULTS  Our results for GAE [135], VGAE [135], DeepWalk [194], Node2Vec [98], GAT [244] and GraphSAGE [101], CTDNE [182] and TGAT [263] are taken directly from the TGAT paper [263]. For Jodie [142] and DyRep [238], in order to make the comparison as fair as possible, we implement our own version in PyTorch as a specific case of our tgn framework. For Jodie we simply use the time embedding module, while for DyRep we augment the messages with the result of a temporal graph attention performed on the destination's neighborhood. For both we use a vanilla RNN as the memory updater module.

Table 4.D.1: Model Hyperparameters.

|  | Value |
| --- | --- |
| Memory Dimension | 172 |
| Node Embedding Dimension | 100 |
| Time Embedding Dimension | 100 |
| # Attention Heads | 2 |
| Dropout | 0.1 |

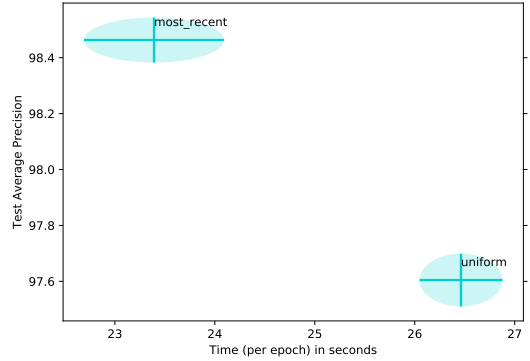

Figure 4.D.1: Comparison of two TGN-attn models using different neighbour sampling strategies (when sampling 10 neighbors). Sampling the most recent edges clearly outperforms uniform sampling. Means and standard deviations (visualized as ellipses) were computed over 10 runs.

### Neighbor Sampling: Uniform vs Most Recent

When performing neighbourhood sampling [102] in static graphs, nodes are usually sampled uniformly. While this strategy is also possible for dynamic graphs, it turns out that the most recent edges are often the most informative. In Fig. 4.D.1, we compare two TGN-attn models (see Tab. 1) with either uniform or most recent neighbour sampling, which shows that a model which samples the most recent edges obtains higher performances.

# 5 Directed Graph Neural Networks

> "Efforts and courage are not enough without purpose
> and direction." – John F. Kennedy

## 5.1 Introduction

Most Graph Neural Networks models assume that the input graph is undirected [102, 136, 244], despite the fact that many real-world networks, such as citation and social networks, are inherently directed. Applying GNNs to directed graphs often involves either converting them to undirected graphs or only propagating information over incoming (or outgoing) edges, both of which may discard valuable information crucial for downstream tasks.

We believe the dominance of undirected graphs in the field is rooted in two "original sins" of GNNs. First, undirected graphs have symmetric Laplacians which admit orthogonal eigendecomposition. Orthogonal Laplacian eigenvectors act as a natural generalization of the Fourier transform and allow to express graph convolution operations in the Fourier domain. Since some of the early graph neural networks originated from the field of graph signal processing [207, 217], the undirected graph assumption was necessary for spectral GNNs [35, 62, 136] to be properly defined. With the emergence of spatial GNNs, unified with the message-passing framework (MPNNs [88]), this assumption was not strictly required anymore, as MPNNs can easily be applied to directed graphs by propagating over the directed adjacency, resulting however in information being propagated only in a single direction, at the risk of discarding useful information from the opposite one. However, early works empirically observed that making the graph undirected consistently leads to better performance on established node-classification benchmarks, which historically have been mainly *homophilic* graphs such as Cora and Pubmed [212], where neighbors tend to share the same label. Consequently, converting input graphs to undirected ones has become a standard part of the dataset preprocessing pipeline, to the extent that the popular GNN library PyTorch-Geometric [72] includes a general utility function that automatically makes graphs undirected when loading datasets[1].

The key observation in this paper is that while accounting for edge directionality indeed does not help in homophilic graphs, it can bring extensive gains in *heterophilic* settings (Fig. 1), where neighbors tend to have different labels. In the rest of the paper, we study *why* and *how* to use directionality to improve learning on heterophilic graphs.

**Contributions.** Our contributions are the following:

- We show that considering the directionality of a graph substantially increases its *effective homophily* in heterophilic settings, with negligible or no impact on homophilic settings (see Sec. 5.3).
- We propose a novel and generic Directed Graph Neural Network framework (Dir-GNN) which extends *any* MPNN to work on directed graphs by performing separate aggregations of the

---

[1]This Pytorch-Geometric routine is used to load datasets stored in an npz format. It makes some directed datasets, such as Cora-ML and Citeseer-Full, automatically undirected without any option to get the directed version instead.

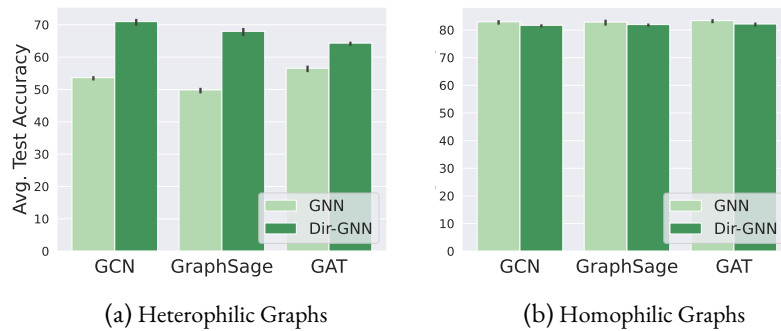

(a) Heterophilic Graphs  (b) Homophilic Graphs

Figure 1: Extending popular GNN architectures with our Dir-GNN framework to incorporate edge-directionality information brings large gains (10% to 15%) on heterophilic datasets (left), while leaving performance mostly unchanged on homophilic datasets (right). The plots illustrate the average performance over all datasets, while the full results are presented in Sec. 5.7.

*incoming* and *outgoing* edges. Moreover, we show that Dir-GNN leads to more homophilic aggregations compared to their undirected counterparts (see Sec. 5.4).
- Our theoretical analysis establishes that Dir-GNN is as expressive as the Directed Weisfeiler-Lehman test, while being *strictly more expressive than MPNNs* (see Sec. 5.4.1).
- We empirically validate that augmenting popular GNN architectures with the Dir-GNN framework yields *large improvements on heterophilic benchmarks*, achieving *state-of-the-art results* and outperforming even more complex methods specifically designed for such settings. Moreover, this enhancement does not negatively impact the performance on homophilic benchmarks (see Sec. 5.7).

## 5.2 NOTATION

We consider a directed graph $G = (V, E)$ with a node set $V$ of $n$ nodes, and an edge set $E$ of $m$ edges. We define its respective directed adjacency matrix $\mathbf{A} \in \{0,1\}^{n \times n}$ where $a_{ij} = 1$ if $(i, j) \in E$ and zero otherwise, its respective undirected adjacency matrix $\mathbf{A}_u$ where $(a_u)_{ij} = 1$ if $(i, j) \in E$ or $(j, i) \in E$ and zero otherwise.

## 5.3 HETEROPHILY IN DIRECTED GRAPHS

In this Section, we discuss how accounting for directionality can be particularly helpful for dealing with heterophilic graphs. By leveraging the directionality information, we argue that even standard MPNNs that are traditionally thought to struggle in the heterophilic regime, can in fact perform extremely well.

**Weighted homophily metrics.** First, we extend the homophily metrics introduced in Section 2.4 to account for directed edges and higher-order neighborhoods. Given a possibly directed and weighted $n \times n$ message-passing matrix $\mathbf{S}$, we define the *weighted node homophily* as

$$h(\mathbf{S}) = \frac{1}{|V|} \sum_{i \in V} \frac{\sum_{j \in V} s_{ij} I[y_i = y_j]}{\sum_{j \in V} s_{ij}} \quad (5.1)$$

5 Directed Graph Neural Networks

|  | | $\mathbf{A}_u$ | $\mathbf{A}_u^2$ | $h_u^{(\text{eff})}$ | $\mathbf{A}$ | $\mathbf{A}^\top$ | $\mathbf{A}^\top\mathbf{A}$ | $\mathbf{A}\mathbf{A}^\top$ | $h_d^{(\text{eff})}$ | $h_{\text{gain}}^{(\text{eff})}$ |
|---|---|---|---|---|---|---|---|---|---|---|
| Homophilic | CITESEER-FULL | 0.958 | 0.951 | 0.958 | 0.954 | 0.959 | 0.971 | 0.951 | 0.971 | 1.36% |
|  | CORA-ML | 0.810 | 0.767 | 0.810 | 0.808 | 0.833 | 0.803 | 0.779 | 0.833 | 2.84% |
|  | OGBN-ARXIV | 0.635 | 0.548 | 0.635 | 0.632 | 0.675 | 0.658 | 0.556 | 0.675 | 6.3% |
| Heterophilic | CHAMELEON | 0.248 | 0.331 | 0.331 | 0.249 | 0.274 | 0.383 | 0.335 | 0.383 | 15.71% |
|  | SQUIRREL | 0.218 | 0.252 | 0.252 | 0.219 | 0.210 | 0.257 | 0.258 | 0.258 | 2.38% |
|  | ARXIV-YEAR | 0.289 | 0.397 | 0.397 | 0.310 | 0.403 | 0.487 | 0.431 | 0.487 | 22.67% |
|  | SNAP-PATENTS | 0.221 | 0.372 | 0.372 | 0.266 | 0.271 | 0.478 | 0.522 | 0.522 | 40.32% |
|  | ROMAN-EMPIRE | 0.046 | 0.365 | 0.365 | 0.045 | 0.042 | 0.535 | 0.609 | 0.609 | 66.85% |

Table 1: Weighted node homophily for different diffusion matrices, and effective homophily for both undirected ($h_u^{(\text{eff})}$) and directed graph ($h_d^{(\text{eff})}$). The last column reports the gain in effective homophily obtained by using the directed graph as opposed to the undirected graph.

Accordingly, by taking $\mathbf{S} = \mathbf{A}$ and $\mathbf{S} = \mathbf{A}^\top$ respectively, we can compute the node homophily based on outgoing or incoming edges. Similarly, we can also take $\mathbf{S}$ to be any *weighted* 2-hop matrix associated with a directed graph (see details below) and compute its node homophily.

We can also extend the construction to edge-computations by defining the $C \times C$ *weighted compatibility matrix* $\mathbf{H}(\mathbf{S})$ of a message-passing matrix $\mathbf{S}$ as

$$h_{kl}(\mathbf{S}) = \frac{\sum_{i,j \in V : y_i = k \wedge y_j = l} s_{ij}}{\sum_{i,j \in V : y_i = k} s_{ij}} \tag{5.2}$$

As above, one can take $\mathbf{S} = \mathbf{A}$ or $\mathbf{S} = \mathbf{A}^\top$ to derive the compatibility matrix associated with the out and in-edges, respectively.

**Effective homophily.** Stacking multiple layers of a GNN effectively corresponds to taking powers of diffusion matrices, resulting in message propagation over higher-order hops. Zhu et al. [294] noted that for heterophilic graphs, the 2-hop tends to be more homophilic than the 1-hop. This phenomenon of similarity within "friends-of-friends" has been widely observed and is commonly referred to as monophily [7]. If higher-order hops exhibit increased homophily, exploring the graph through layers can prove beneficial for the task. Consequently, we introduce the concept of *effective homophily* as the maximum weighted node homophily observable at any hop of the graph.

For directed graphs, there exists an exponential number of $k$-hops. For instance, four 2-hop matrices can be considered: the squared operators $\mathbf{A}^2$ and $(\mathbf{A}^\top)^2$, which correspond to following the same forward or backward edge direction twice, as well as the *non*-squared operators $\mathbf{A}\mathbf{A}^\top$ and $\mathbf{A}^\top\mathbf{A}$, representing the forward/backward and backward/forward edge directions. Given a graph $G$, we define its effective homophily as follows:

$$h^{(\text{eff})} = \max_{k \geqslant 1} \max_{\mathbf{C} \in \mathcal{B}^k} h(\mathbf{C}) \tag{5.3}$$

where $\mathcal{B}^k$ denotes the set of all $k$-hop matrices for a graph. If $G$ is undirected, $\mathcal{B}^k$ contains only $\mathbf{A}^k$. In our empirical analysis, we will focus on the 2-hop matrices, as computing higher-order $k$-hop matrices becomes intractable for all but the smallest graphs [2].

**Leveraging directionality to enhance effective homophily.** We observe that converting a graph from directed to undirected results in lower effective homophily for heterophilic graphs, while the impact on homophilic graphs is negligible (refer to the last column of Tab. 1). Specifically, $\mathbf{A}\mathbf{A}^\top$ and $\mathbf{A}^\top\mathbf{A}$ emerge as the most homophilic diffusion matrices for heterophilic graphs. In fact, the average relative gain of effective homophily, $h_{\text{gain}}^{(\text{eff})}$, when using directed graphs compared to undirected ones is only around 3% for homophilic datasets, while it is almost 30% for heterophilic datasets. We further validate this observation on synthetic directed graphs exhibiting various levels of homophily, generated through a modified preferential attachment process (see Appendix 5.F.1 for more details). Fig. 2a displays the results: the directed graph consistently demonstrates higher effective homophily compared to its undirected counterpart, with the gap being particularly prominent for lower node homophily levels. The minimal effective homophily gain on homophilic datasets further substantiates the traditional practice of using undirected graphs for benchmarking GNNs, as the datasets were predominantly homophilic until recently.

---

[2]This is attributed to the fact that while $\mathbf{A}$ is typically quite sparse, $\mathbf{A}^k$ grows increasingly dense as $k$ increases, quickly approaching $n^2$ non-zero entries.

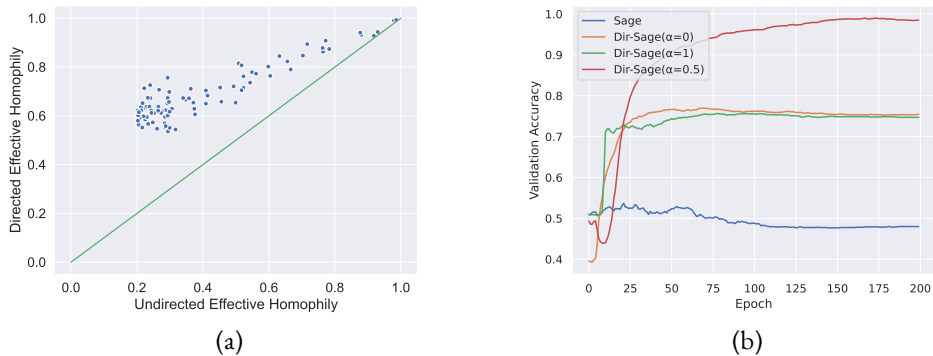

Figure 2: In our synthetic experiments, we observe the following: (a) the effective homophily of directed graphs is consistently higher compared to their undirected counterparts. Interestingly, this gap widens for graphs that are less homophilic. (b) When examining the performance of GraphSage and its Dir-GNN extensions on a synthetic task requiring directionality information, only Dir-Sage($\alpha$=0.5), which utilizes information from both directions, is capable of solving the task.

**Real-world example.** We illustrate the concept of effective homophily in heterophilic directed graphs by taking the concrete task of predicting the publication year of papers based on a directed citation network such as Arxiv-Year [154]. In this case, the different 2-hop neighbourhoods have very different semantics: the diffusion operator $(\mathbf{A}^2)_i$ represents papers that are cited by those papers that paper $i$ cites. As these 2-hop neighboring papers were published further in the past relative to paper $i$, they do not offer much information about the publication year of $i$. On the other hand, $(\mathbf{A}^\top \mathbf{A})_i$ represents papers that share citations with paper $i$. Papers cited by the same sources are more likely to have been published in the same period, so the diffusion operator $\mathbf{A}^\top \mathbf{A}$ is expected to be more homophilic. The undirected 2-hop operator $\mathbf{A}_u^2 = (\frac{1}{2}(\mathbf{A} + \mathbf{A}^\top))^2 = \frac{1}{4}(\mathbf{A}^2 + (\mathbf{A}^\top)^2 + \mathbf{A}\mathbf{A}^\top + \mathbf{A}\mathbf{A}^\top)$ is the average of the four directed 2-hops. Therefore, the highly homophilic matrix $\mathbf{A}^\top \mathbf{A}$ is diluted by the inclusion of $(\mathbf{A}^2)$, leading to a less homophilic operator overall.

**Harmless heterophily through directions.** Ma et al. [159] showed that heterophily is not necessarily harmful for GNNs as long as nodes with the same label share similar neighbourhood patterns and different classes have distinguishable patterns. We find that some directed datasets, such as Arxiv-Year and Snap-Patents, show this form of *harmless heterophily* when treated as directed, and instead manifest *harmful heterophily* when made undirected (see Fig. 5.A.2 in the Appendix). This suggests that using directionality can be beneficial also when using only one layer, as we confirm empirically (see Fig. 5.G.2 in the Appendix).

**Toy example.** We further illustrate the concepts presented in this Section with the toy example in Fig. 3, which shows a directed graph with three classes (blue, orange, green). Despite the graph being maximally heterophilic, it presents harmless heterophily since different classes have very different neighbourhood patterns that can be observed clearly from the compatibility matrix of $\mathbf{A}$ (b). When the graph is made undirected (c), we are corrupting this information, and the classes become less distinguishable, making the task harder. We also note that both $\mathbf{A}^\top \mathbf{A}$ and $\mathbf{A}\mathbf{A}^\top$ present perfect homophily (d), while $\mathbf{A}_u^2$ does not, in line with the discussion in previous paragraphs.

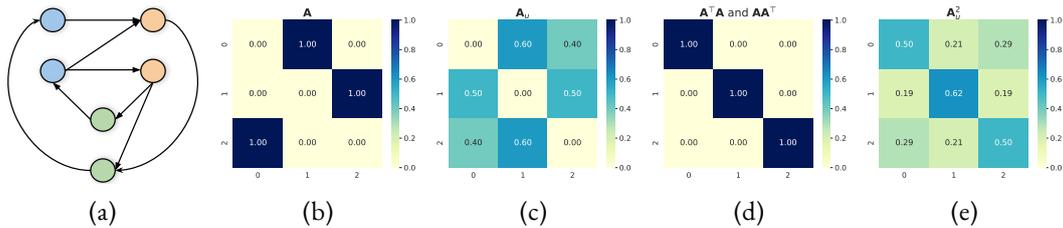

Figure 3: (a) A toy directed graph with three classes showcasing harmless heterophily. (b) Compatibility matrix of $\mathbf{A}$ showing that classes (blue, orange and green) have very different neighbourhoods and can be easily distinguished. (c) Making the graph undirected makes the classes harder to distinguish, making the task harder to solve. (d) The mixed directed 2-hops ($\mathbf{A}^\top \mathbf{A}$ and $\mathbf{A}\mathbf{A}^\top$) have perfect homophily, while (e) this is not the case for the undirected 2-hop.

## 5.4 Directed Graph Neural Network

In this Section, we extend the class of MPNNs to directed graphs, and refer to such generalization as Directed Graph Neural Network (Dir-GNN). We follow the scheme in Eq. (2.4), meaning that the update of a node feature is the result of a *combination* of its previous state and an *aggregation* over its neighbours. Crucially though, the characterization of neighbours now has to account for the edge-directionality. Accordingly, given a node $i \in V$, we perform separate aggregations over the in-neighbours ($j \to i$) and the out-neighbours ($i \to j$) respectively:

$$\begin{aligned}
\mathbf{m}_{i,\leftarrow}^{(k)} &= \mathrm{AGG}_{\leftarrow}^{(k)}\left(\{\!\!\{(\mathbf{x}_j^{(k-1)}, \mathbf{x}_i^{(k-1)}) : (j,i) \in E\}\!\!\}\right) \\
\mathbf{m}_{i,\rightarrow}^{(k)} &= \mathrm{AGG}_{\rightarrow}^{(k)}\left(\{\!\!\{(\mathbf{x}_j^{(k-1)}, \mathbf{x}_i^{(k-1)}) : (i,j) \in E\}\!\!\}\right) \\
\mathbf{x}_i^{(k)} &= \mathrm{COM}^{(k)}\left(\mathbf{x}_i^{(k-1)}, \mathbf{m}_{i,\leftarrow}^{(k)}, \mathbf{m}_{i,\rightarrow}^{(k)}\right).
\end{aligned} \quad (5.4)$$

The idea behind our framework is that given *any* MPNN, we can readily adapt it to the directed case by choosing how to aggregate over both directions. For this purpose, we replace the neighbour aggregator $\mathrm{AGG}^{(k)}$ with two separate in- and out-aggregators $\mathrm{AGG}_{\leftarrow}^{(k)}$ and $\mathrm{AGG}_{\rightarrow}^{(k)}$, which can have *independent* sets of parameters for the two aggregation phases and, possibly, different normalizations as discussed below – however the functional expression in both cases stays the same. As we will show, accounting for both directions separately is fundamental for both the expressivity of the model (see Sec. 5.4.1) and its empirical performance (see Sec. 5.6).

**Extension of common architectures.** To make our discussion more explicit, we describe extensions of popular MPNNs where the aggregation map is computed by $\mathbf{m}_i^{(k)} = (\mathbf{S}\mathbf{x}^{(k-1)})_i$, where $\mathbf{S} \in \mathbb{R}^{n \times n}$ is a message-passing matrix. In GCN [136], $\mathbf{S} = \mathbf{D}^{-1/2}\mathbf{A}_u\mathbf{D}^{-1/2}$, where $\mathbf{D}$ is the degree matrix of the undirected graph. In the case of directed graphs, two message-passing matrices $\mathbf{S}_{\leftarrow}$ and $\mathbf{S}_{\rightarrow}$ are required for in- and out-neighbours respectively. Additionally, the normalization is slightly more subtle, since we now have two different diagonal degree matrices

## 5 Directed Graph Neural Networks

$\mathbf{D}_{\leftarrow}$ and $\mathbf{D}_{\rightarrow}$ containing the in-degrees and out-degrees respectively. Accordingly, we propose a normalization of the form $\mathbf{S}_{\rightarrow} = \mathbf{D}_{\rightarrow}^{-1/2}\mathbf{A}\mathbf{D}_{\leftarrow}^{-1/2}$ i.e. $(\mathbf{S}_{\rightarrow})_{ij} = a_{ij}/\sqrt{d_i^{\rightarrow} d_j^{\leftarrow}}$. To motivate this choice, note that the normalization modulates the aggregation based on the out-degree of $i$ and the in-degree of $j$ as one would expect given that we are computing a message going from $i$ to $j$. We can then take $\mathbf{S}_{\leftarrow} = \mathbf{S}_{\rightarrow}^{\top}$ and write the update at layer $k$ of Dir-GCN as

$$\mathbf{X}^{(k)} = \sigma\left(\mathbf{S}_{\rightarrow}\mathbf{X}^{(k-1)}\mathbf{W}_{\rightarrow}^{(k)} + \mathbf{S}_{\rightarrow}^{\top}\mathbf{X}^{(k-1)}\mathbf{W}_{\leftarrow}^{(k)}\right), \tag{5.5}$$

for learnable channel-mixing matrices $\mathbf{W}_{\rightarrow}^{(k)}, \mathbf{W}_{\leftarrow}^{(k)}$ and with $\sigma$ a pointwise activation map. Finally, we note that in our implementation of Dir-GNN we use an additional learnable or tunable parameter $\alpha$ allowing the framework to weigh one direction more than the other (a convex combination), depending on the dataset. Dir-GNN extensions of GAT [244] and GraphSAGE [102] can be found in Appendix 5.B.

**Dir-GNN leads to more homophilic aggregations.** Since our main application amounts to relying on the graph-directionality to mitigate heterophily, here we comment on how information is iteratively propagated in a Dir-GNN, generally leading to an aggregation scheme beneficial on most real-world directed heterophilic graphs. We focus on the Dir-GCN formulation, however the following applies to any other Dir-GNN up to changing the message-passing matrices. Consider a 2-layer Dir-GCN as in Eq. (5.5), and let us remove the pointwise activation $\sigma$ [3]. Then, the node representation can be written as

$$\mathbf{X}^{(2)} = \mathbf{A}_{\rightarrow}^2 \mathbf{X}^{(0)} \mathbf{W}_{\rightarrow}^{(1)} \mathbf{W}_{\rightarrow}^{(2)} + (\mathbf{A}_{\rightarrow}^{\top})^2 \mathbf{X}^{(0)} \mathbf{W}_{\leftarrow}^{(1)} \mathbf{W}_{\leftarrow}^{(2)} + \mathbf{A}_{\rightarrow}\mathbf{A}_{\rightarrow}^{\top} \mathbf{X}^{(0)} \mathbf{W}_{\leftarrow}^{(1)} \mathbf{W}_{\rightarrow}^{(2)} + \mathbf{A}_{\rightarrow}^{\top}\mathbf{A}_{\rightarrow} \mathbf{X}^{(0)} \mathbf{W}_{\rightarrow}^{(1)} \mathbf{W}_{\leftarrow}^{(2)}.$$

We observe that when we aggregate information over multiple layers, the final node representation is derived by also computing convolutions over 2-hop matrices $\mathbf{A}_{\rightarrow}^{\top}\mathbf{A}_{\rightarrow}$ and $\mathbf{A}_{\rightarrow}\mathbf{A}_{\rightarrow}^{\top}$. From the discussion in Sec. 5.3, we deduce that this framework may be more suited to handle heterophilic graphs since generally such 2-hop matrices are more likely to encode similarity than $\mathbf{A}_{\rightarrow}^2$, $(\mathbf{A}_{\rightarrow}^{\top})^2$ or $\mathbf{A}_u^2$ – this is validated empirically on real-world datasets in Sec. 5.7.

**Advantages of two-directional updates.** We discuss the benefits of incorporating both directions in the layer update, as opposed to using a single direction. Although spatial MPNNs can be adapted to directed graphs by simply utilizing $\mathbf{A}$ instead of $\mathbf{A}_u$—resulting in message propagation only along out-edges—relying on a single direction presents three primary drawbacks. First, if the layer update only considers one direction, the exploration of the multi-hop neighbourhoods through powers of diffusion operators would not include the mixed terms $\mathbf{A}\mathbf{A}^{\top}$ and $\mathbf{A}^{\top}\mathbf{A}$, which have been shown to be particularly beneficial for heterophilic graphs in Sec. 5.3. Second, by using only one direction we disregard the graph entirely for nodes where the out-degree is zero [4]. This phenomenon frequently occurs in real-world graphs, as reported in Tab. 5.F.3. Incorporating both directions in the layer update helps mitigate this problem, as it is far less common for a node to have both in- and out-degree to be zero, as also illustrated in Tab. 5.F.3. Third, limiting the update to a single direction reduces expressivity, as we discuss in Sec. 5.4.1.

**Complexity.** The complexity of Dir-GNN depends on the specific instantiation of the framework. Dir-GCN, Dir-Sage, and Dir-GAT maintain the same per-layer computational complexity as their undirected counterparts ($\mathcal{O}(md + nd^2)$ for GCN and GraphSage, and $\mathcal{O}(md^2)$ for GAT).

---

[3] Note that this does not affect our discussion, in fact, any observation can be extended to the non-linear case by computing the Jacobian of node features as in Topping et al. [236].

[4] Or in-degree, depending on which direction is selected.

However, they have twice as many parameters, owing to their separate weight matrices for in- and out-neighbors.

### 5.4.1 Expressive power of Dir-GNN

It is a well-known result that MPNNs are bound in expressivity by the 1-WL test, and that it is possible to construct MPNN models which are as expressive as the 1-WL test [264]. In this section, we show that Dir-GNN is the optimal way to extend MPNNs to directed graphs. We do so by proving that Dir-GNN models can be constructed to be as expressive as an extension of the 1-WL test to directed graphs [97], referred to as *D-WL* (for a formal definition, see Appendix 5.C.1). Additionally, we illustrate its greater expressivity over more straightforward approaches, such as converting the graph to its undirected form and utilizing a standard MPNN (*MPNN-U*) or applying an MPNN that propagates solely along edge direction (*MPNN-D*)[5]. Formal statements for the theorems in this section along with their proofs can be found in Appendix 5.C.

**Theorem 5.4.1** (Informal). *Dir-GNN is as expressive as D-WL if* $\mathrm{AGG}_{\rightarrow}^{(k)}$, $\mathrm{AGG}_{\leftarrow}^{(k)}$, *and* $\mathrm{COM}^{(k)}$ *are injective for all k.*

A discussion of how a Dir-GNN can be parametrized to meet these conditions (similarly to what is done in Xu et al. [264]) can be found in Appendix 5.C.4.

**Theorem 5.4.2** (Informal). *Dir-GNN is strictly more expressive than both MPNN-U and MPNN-D.*

Intuitively, the theorem states that while all directed graphs distinguished by MPNNs are also separated by Dir-GNNs, there also exist directed graphs separated by the latter but not by the former. This holds true for MPNNs applied both on the directed and undirected graph. We observe these theoretical findings to be in line with the empirical results detailed in Appendix 5.G and Tab. 5.F.4, where Dir-GNN performs comparably or better (typically in the case of heterophily) than MPNNs.

## 5.5 Related Work

**GNNs for directed graphs.** While several papers have alluded to the extension of their spatial models to directed graphs, empirical validation has not been conducted [88, 210]. GatedGCN [152] deals with directed graphs; however, it aggregates information only from out-neighbours, neglecting potentially valuable information from in-neighbours. More recently, Vrček et al. [248] tackle the genome assembly problem by employing a GatedGCN with separate aggregations for in- and out-neighbors. Peach et al. [191] propose a model akin to running two independent GNNs on $\mathbf{A}$ and $\mathbf{A}^\top$ respectively, which, however, prevents mixing of the two directions, which we found to be highly important. On the other hand, various approaches have been developed to generalize spectral convolutions for directed graphs [160, 174, 235]. Of particular interest are DiGCN [234], which uses Personalized Page Rank matrix as a generalized Laplacian and incorporates $k$-hop diffusion matrices, and MagNet [287], which adopts a complex matrix for graph diffusion where the real and imaginary parts represent the undirected adjacency and edge direction respectively. The above spectral methods share the following limitations: 1) in-neighbors and out-neighbors share the same weight matrix, which restricts expressivity; 2) they are specialized models, often inspired by GCN, as opposed to broader frameworks; 3) their scalability is severely limited due to their spectral nature. Concurrently to our work Geisler et al. [85] extend transformers to directed graph for the task of graph classification, while Maskey et al. [166] generalize the concept of oversmoothing to directed graphs.

---

[5]The same results apply to a model which sends messages only along in-edges.

**GNNs for relational graphs.** Our Dir-GCN model can be considered as a Relational Graph Convolutional Network (R-GCN) [211] applied to an augmented relational graph that incorporates two relation types: one for the original edges and another for the inverse edges added to the graph. Several papers handled multi-relational directed graphs by adding inverse relations [122, 164, 211, 242]. Our Dir-GNN framework, however, extends these methods by introducing more versatile AGG and COMB functions. Moreover, we are also the first to perform an in-depth investigation of the role of directionality in graph learning and its relation with homophily of the graph.

**Heterophilic GNNs.** Several GNN architectures have been proposed to handle heterophily. One way amounts to effectively allow the model to enhance the high-frequency components by learning 'generalized' negative weights on the graph [29, 30, 50, 64, 157]. A different approach tries to enlarge the neighbourhood aggregation to take advantage of the fact that on heterophilic graphs, the likelihood of finding similar nodes increases beyond the 1-hop [2, 150, 154, 168, 294].

## 5.6 Experiments on Synthetic Data

**Setup.** In order to show the limits of current MPNNs, we design a synthetic task where the label of a node depends on both its in- and out-neighbours: it is one if the mean of the scalar features of their in-neighbours is greater than the mean of the features of their out-neighbours, or zero otherwise (more details in Appendix 5.F.2). We report the results using GraphSage as base MPNN, but similar results were obtained with GCN and GAT and reported in Fig. 5.G.1 of the Appendix. We compare GraphSage on the undirected version of the graph (Sage), with three Dir-GNN extensions of GraphSage using different convex combination coefficients $\alpha$: Dir-Sage($\alpha = 0$) (only considering in-edges), Dir-Sage($\alpha = 1$) (only considering out-edges) and Dir-Sage($\alpha = 0.5$) (considering both in- and out-edges equally).

**Results.** The results show that only *Dir-Sage($\alpha$=0.5)*, which accounts for both directions, is able to almost perfectly solve the task. Using only in- or out-edges results in around 75% accuracy, whereas GraphSage on the undirected graph is no better than a random classifier.

## 5.7 Experiments on Real-World Datasets

**Datasets.** We evaluate on the task of node classification on several directed benchmark datasets with varying levels of homophily: Citesee-Full, Cora-ML [32], OGBN-Arxiv [113], Chameleon, Squirrel [192], Arxiv-Year, Snap-Patents [154] and Roman-Empire [195] (refer to Tab. 5.F.1 for dataset statistics). While the first three are mainly homophilic (edge homophily greater than 0.65), the last five are highly heterophilic (edge homophily smaller than 0.24). Refer to Appendix 5.F.3 for more details on the experimental setup and on dataset splits.

### 5.7.1 Extending Popular GNNs with Dir-GNN

**Setup.** We evaluate the gain of extending popular undirected GNN architectures (GCN [136], GraphSage [102] and GAT [244]) with our framework. For this ablation, we use the same hyper-parameters (provided in Appendix 5.F.4) for all models and datasets. The aggregated results are plotted in Fig. 1, while the raw numbers are reported in Sec. 5.7. For Dir-GNN, we take the best results out of $\alpha \in \{0, 0.5, 1\}$ (see Tab. 5.F.4 for the full results).

|  | Homophilic | | | Heterophilic | | | | |
|---|---|---|---|---|---|---|---|---|
|  | CITESEER_FULL | CORA_ML | OGBN-ARXIV | CHAMELEON | SQUIRREL | ARXIV-YEAR | SNAP-PATENTS | ROMAN-EMPIRE |
| HOM. | 0.949 | 0.792 | 0.655 | 0.235 | 0.223 | 0.221 | 0.218 | 0.05 |
| HOM. GAIN | 1.36% | 2.84% | 6.30% | 15.71% | 2.38% | 22.67% | 40.32% | 66.85% |
| GCN | 93.37 ± 0.22 | 84.37 ± 1.52 | 68.39 ± 0.01 | 71.12 ± 2.28 | 62.71 ± 2.27 | 46.28 ± 0.39 | 51.02 ± 0.07 | 56.23 ± 0.37 |
| DIR-GCN | 93.44 ± 0.59 | 84.45 ± 1.69 | 66.66 ± 0.02 | **78.77 ± 1.72** | **74.43 ± 0.74** | **59.56 ± 0.16** | **71.32 ± 0.06** | **74.54 ± 0.71** |
| SAGE | **94.15 ± 0.61** | **86.01 ± 1.56** | **67.78 ± 0.07** | 61.14 ± 2.00 | 42.64 ± 1.72 | 44.05 ± 0.02 | 52.55 ± 0.10 | 72.05 ± 0.41 |
| DIR-SAGE | 94.14 ± 0.65 | 85.84 ± 2.09 | 65.14 ± 0.03 | **64.47 ± 2.27** | **46.05 ± 1.16** | **55.76 ± 0.10** | **70.26 ± 0.14** | **79.10 ± 0.19** |
| GAT | **94.53 ± 0.48** | **86.44 ± 1.45** | **69.60 ± 0.01** | 66.82 ± 2.56 | 56.49 ± 1.73 | 45.30 ± 0.23 | OOM | 49.18 ± 1.35 |
| DIR-GAT | 94.48 ± 0.52 | 86.21 ± 1.40 | 66.50 ± 0.16 | **71.40 ± 1.63** | **67.53 ± 1.04** | **54.47 ± 0.14** | OOM | **72.25 ± 0.04** |

Table 2: Ablation study comparing base MPNNs on the undirected graphs versus their Dir-GNN extension on the directed graphs. Homophilic datasets, located to the left of the dashed line, show little to no improvement when incorporating directionality, sometimes even experiencing a minor decrease in performance. Conversely, heterophilic datasets, found to the right of the dashed line, demonstrate large accuracy improvements when directionality is incorporated into the model.

|  | Squirrel | Chameleon | Arxiv-year | Snap-patents | Roman-Empire |
|---|---|---|---|---|---|
| MLP | 28.77 ± 1.56 | 46.21 ± 2.99 | 36.70 ± 0.21 | 31.34 ± 0.05 | 64.94 ± 0.62 |
| GCN | 53.43 ± 2.01 | 64.82 ± 2.24 | 46.02 ± 0.26 | 51.02 ± 0.06 | 73.69 ± 0.74 |
| H$_2$GCN | 37.90 ± 2.02 | 59.39 ± 1.98 | 49.09 ± 0.10 | OOM | 60.11 ± 0.52 |
| GPR-GNN | 54.35 ± 0.87 | 62.85 ± 2.90 | 45.07 ± 0.21 | 40.19 ± 0.03 | 64.85 ± 0.27 |
| LINKX | 61.81 ± 1.80 | 68.42 ± 1.38 | 56.00 ± 0.17 | 61.95 ± 0.12 | 37.55 ± 0.36 |
| FSGNN | 74.10 ± 1.89 | 78.27 ± 1.28 | 50.47 ± 0.21 | 65.07 ± 0.03 | 79.92 ± 0.56 |
| ACM-GCN | 67.40 ± 2.21 | 74.76 ± 2.20 | 47.37 ± 0.59 | 55.14 ± 0.16 | 69.66 ± 0.62 |
| GloGNN | 57.88 ± 1.76 | 71.21 ± 1.84 | 54.79 ± 0.25 | 62.09 ± 0.27 | 59.63 ± 0.69 |
| Grad. Gating | 64.26 ± 2.38 | 71.40 ± 2.38 | 63.30 ± 1.84 | 69.50 ± 0.39 | 82.16 ± 0.78 |
| DiGCN | 37.74 ± 1.54 | 52.24 ± 3.65 | OOM | OOM | 52.71 ± 0.32 |
| MagNet | 39.01 ± 1.93 | 58.22 ± 2.87 | 60.29 ± 0.27 | OOM | 88.07 ± 0.27 |
| Dir-GNN (Ours) | **75.31 ± 1.92** | **79.71 ± 1.26** | **64.08 ± 0.26** | **73.95 ± 0.05** | **91.23 ± 0.32** |

Table 3: Results on real-world directed heterophilic datasets. OOM indicates out of memory.

**Results.** On **heterophilic datasets**, using directionality brings exceptionally large gains (10% to 20% absolute) in accuracy across all three base GNN models. On the other hand, on **homophilic datasets** using directionality leaves the performance unchanged or slightly hurts. This is in line with the findings of Tab. 1, which shows that using directionality as in our framework generally increases the effective homophily of heterophilic datasets, while leaving it almost unchanged for homophilic datasets. The inductive bias of undirected GNNs to propagate information in the *same* way in both directions is beneficial on homophilic datasets where edges encode a notion of class similarity. Moreover, averaging information across *all* your neighbors, independent of direction, leads to a low-pass filtering effect that is indeed beneficial on homophilic graphs [184]. In contrast, Dir-GNN has to learn to align in- and out-convolutions since they have independent weights.

### 5.7.2 Comparison with State-of-the-Art Models

**Setup.** Given the importance of directionality on heterophilic tasks, we compare Dir-GNN with **simple baselines**: MLP and GCN [136], **heterophilic state-of-the-art models**: H$_2$GCN [294], GPR-GNN [50], LINKX [154], FSGNN [168], ACM-GCN [157], GloGNN [150], Gradient Gating [206], and **state-of-the-art models for directed graphs**: DiGCN [234] and MagNet [287]. Appendix 5.F.6 contains more details on how baseline results were obtained. Differently from the results in Sec. 5.7, we now tune the hyperparameters of our model using a grid search (see Appendix 5.F.5 for the exact ranges).

**Results.** We observe that Dir-GNN obtains **new state-of-the-art** results on all five heterophilic datasets, outperforming complex methods which were specifically designed to tackle heterophily. These results suggest that, when present, *using the edge direction can significantly improve learning on heterophilic graphs*, justifying the title of the paper. In contrast, discarding it is so harmful that not even complex architectures can make up for this loss of information. We further note that DiGCN and MagNet, despite being specifically designed for directed graphs, struggle on Squirrel and Chameleon. This is due to their inability to selectively aggregate from one direction while disregarding the other, a strategy that proves particularly advantageous for these two datasets (see Tab. 5.F.4). Our proposed Dir-GNN framework overcomes this limitation thanks to its distinct weight matrices and the flexibility provided by the $\alpha$ parameter, enabling selective directional aggregation.

## 5.8 Discussion

**Limitations.** Our research has several areas that could be further refined and explored. First, the theoretical exploration of the conditions that lead to higher effective homophily in directed graphs compared to their undirected counterparts is still largely unexplored. Furthermore, we have yet to investigate the expressivity advantage of Dir-GNN in the specific context of heterophilic graphs, where empirical gains were most pronounced. Finally, we haven't empirically investigated different functional forms for aggregating in- and out-edges. These aspects mark potential areas for future enhancements and investigations.

## 5.9 Conclusion

We introduced Dir-GNN, a generic framework to extend any spatial graph neural network to directed graphs, which we prove to be strictly more expressive than MPNNs. We showed that treating the graph as directed improves the effective homophily of heterophilic datasets, and validated empirically that augmenting popular GNN architectures with our framework results in large improvements on heterophilic benchmarks, while leaving performance almost unchanged on homophilic benchmarks. Surprisingly, we found simple instantiations of our framework to obtain state-of-the-art results on the five directed heterophilic benchmarks we experimented on, outperforming recent architectures developed specifically for heterophilic settings as well as previously proposed methods for directed graphs.

FAST-FORWARD TO TODAY  Despite having been published only a few months ago, Dir-GNN has already been incorporated into PyTorch-Geometric [72][6] as the default solution to extend any GNN to directed graphs.

---

[6] https://pytorch-geometric.readthedocs.io/en/latest/generated/torch_geometric.nn.conv.DirGNNConv.html#torch_geometric.nn.conv.DirGNNConv

# Appendices

## 5.A Compatibility Matrices

Fig. 5.A.1 shows the compatibility matrices for both Citeseer-Full (homophilic) and Chameleon (heterophilic). Additionally, Fig. 5.A.2 presents the weighted compatibility matrices of the undirected diffusion operator $\mathbf{A}_u$ and the two directed diffusion operators $\mathbf{A}$ and $\mathbf{A}^\top$ for Arxiv-Year. The last two have rows (classes) which are much more distinguishable then the first, despite still being heterophilic. This phenomen, called harmless heterophily, is discussed in Sec. 5.3.

## 5.B Extension of Popular GNNs

We first consider an extension of GraphSAGE [102] using our Dir-GNN framework. The main choice reduces to that of normalization. In the spirit of GraphSAGE, we require the message-passing matrices $\mathbf{A}_\leftarrow$ and $\mathbf{A}_\rightarrow$ to both be row-stochastic. This is done by taking $\mathbf{A}_\rightarrow = \mathbf{D}_\rightarrow^{-1}\mathbf{A}$ and $\mathbf{A}_\leftarrow = \mathbf{D}_\leftarrow^{-1}\mathbf{A}^\top$, respectively. In this case, the directed version of GraphSAGE becomes

$$\mathbf{X}^{(k)} = \sigma(\mathbf{X}^{(k-1)}\mathbf{\Omega}^{(k)} + \mathbf{D}_\rightarrow^{-1}\mathbf{A}\mathbf{X}^{(k-1)}\mathbf{W}_\rightarrow^{(k)} + \mathbf{D}_\leftarrow^{-1}\mathbf{A}^\top\mathbf{X}^{(k-1)}\mathbf{W}_\leftarrow^{(k)}).$$

Finally, we consider the generalization of GAT [244] to the directed case. Here we simply compute attention coefficients over the in- and out-neighbours separately. If we denote the attention coefficient over the edge $(i,j)$ by $\beta_{ij}$, then the update of node $i$ at layer $k$ can be computed as

$$\mathbf{h}_i^{(k)} = \sigma\Big(\sum_{(i,j)\in E} \beta_{ij}^\rightarrow \mathbf{W}_\rightarrow^{(k)} \mathbf{h}_j^{(k)} + \sum_{(j,i)\in E} \beta_{ji}^\leftarrow \mathbf{W}_\leftarrow^{(k)} \mathbf{h}_j^{(k)}\Big),$$

where $\beta^\rightarrow, \beta^\leftarrow$ are both row-stochastic matrices with support given by $\mathbf{A}$ and $\mathbf{A}^\top$, respectively.

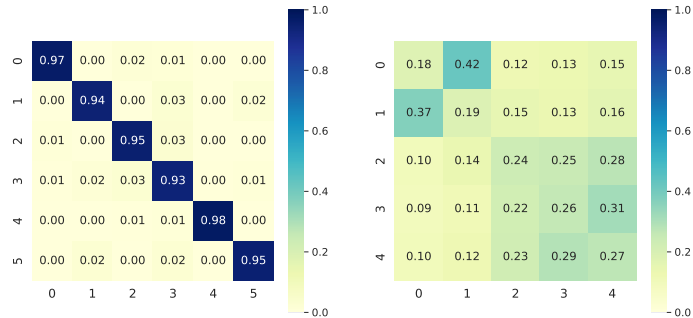

Figure 5.A.1: Compatibility matrices for the undirected version of Citeseer-Full (homophilic, left) and Chameleon (heterophilic, right).

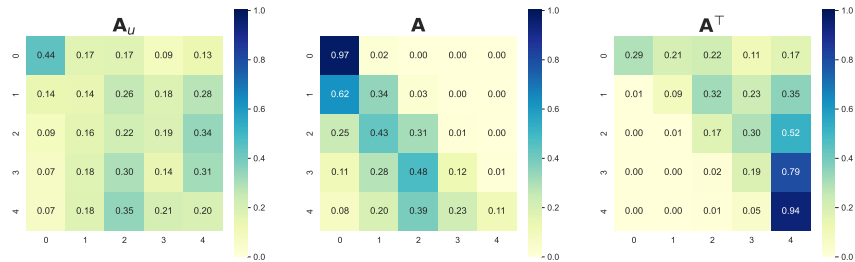

Figure 5.A.2: Weighted compatibility matrices of the undirected diffusion operator $\mathbf{A}_u$ and the two directed diffusion operators $\mathbf{A}$ and $\mathbf{A}^\top$ for Arxiv-Year. The last two have rows (classes) which are much more distinguishable then the first, despite still being heterophilic.

## 5.C Analysis of Expressivity

In this appendix we prove the expressivity results reported in Sec. 5.4.1, after restating them more formally. It is important to note that we cannot build on the expressivity results from Barcelo et al. [13], since their scope is limited to undirected relational graphs, and (perhaps surprisingly) it is not possible to equivalently represent a directed graph with an undirected relational graphs, as we show in Appendix 5.D. In our analysis, we will assume all nodes to have constant scalar node features $c$. We start by introducing useful concepts which will be instrumental to our discussion. As commonly done, we will assume in our analysis that all nodes have constant scalar node features $c$.

### 5.C.1 (Directed) Weisfeiler-Lehman Test

The 1-dimensional Weisfeiler-Lehman algorithm (1-WL), or color refinement, is a heuristic approach to the graph isomorphism problem, initially proposed by Weisfeiler et al. [254]. This algorithm is essentially an iterative process of vertex labeling or coloring, aimed at identifying whether two graphs are non-isomorphic.

Starting with an identical coloring of the vertices in both graphs, the algorithm proceeds through multiple iterations. In each round, vertices with identical colors are assigned different colors if their respective sets of equally-colored neighbors are unequal in number. The algorithm continues this process until it either reaches a point where the distribution of vertices colors is different in the two graphs or converges to the same distribution. In the former case, the algorithm concludes that the graphs are not isomorphic and halts. Alternatively, the algorithm terminates with an inconclusive result: the two graphs are 'possibly isomorphic'. It has been shown that this algorithm cannot distinguish all non-isomorphic graphs [36].

Formally, given an undirected graph $G = (V, E)$, the 1-WL algorithm calculates a node coloring $C^{(t)} : V(G) \to \mathbb{N}$ for each iteration $t > 0$, as follows:

$$C^{(t)}(i) = \text{RELABEL}\Big(C^{(t-1)}(i), \{\!\!\{C^{(t-1)}(j) : j \in N(i)\}\!\!\}\Big) \tag{5.6}$$

where RELABEL is a function that injectively assigns a unique color, not used in previous iterations, to the pair of arguments. The function $N(i)$ represents the set of neighbours of $i$.

Since we deal with directed graphs, it is necessary to extend the 1-WL test to accommodate directed graphs. We note that a few variants have been proposed in the literature [97, 122, 140]. Here, we focus on a variant whereby in- and out-neighbours are treated separately, as discussed in Grohe et al. [97]. This variant, which we refer to as *D-WL*, refines colours as follows:

$$D^{(t)}(i) = \text{RELABEL}\Big(D^{(t-1)}(i), \{D^{(t-1)}(j) : j \in N_\rightarrow(i)\}, \{D^{(t-1)}(j) : j \in N_\leftarrow(i)\}\Big)$$
(5.7)

where $N_\rightarrow(i)$ and $N_\leftarrow(i)$ are the set of out- and in-neighbors of $i$, respectively. Our first objective is to demonstrate that Dir-GNN is as expressive as D-WL. Establishing this will enable us to further show that Dir-GNN is strictly more expressive than an MPNN operating on either the directed or undirected version of a graph. Let us start by introducing some further auxiliary tools that will be used in our analysis.

5.C.2 EXPRESSIVENESS AND COLOR REFINEMENTS

A way to compare graph models (or algorithms) in their expressiveness is by contrasting their discriminative power, that is the ability they have to disambiguate between non-isomorphic graphs.

Two graphs are called *isomorphic* whenever there exists a graph isomorphism between the two:

**Definition 5.C.1** (Graph isomorphism). Let $G_1 = (V_1, E_1), G_2 = (V_2, E_2)$ be two (directed) graphs. An *isomorphism* between $G_1, G_2$ is a bijective map $\varphi : V_1 \rightarrow V_2$ which preserves adjacencies, that is: $\forall u, v \in V_1 : (u, v) \in E_1 \iff (\varphi(u), \varphi(v)) \in E_2$.

On the contrary, they are deemed non-isomorphic when such a bijection does not exist. A model that is able to discriminate between two non-isomorphic graphs assigns them distinct representations. This concept is extended to families of models as follows:

**Definition 5.C.2** (Graph discrimination). Let $G = (V, E)$ be any (directed) graph and $M$ a model belonging to some family $\mathcal{M}$. Let $G_1$ and $G_2$ be two graphs. We say $M$ discriminates $G_1$, $G_2$ iff $M(G_1) \neq M(G_2)$. We write $G_1 \neq_M G_2$. If there exists such a model $M \in \mathcal{M}$, then family $\mathcal{M}$ distinguishes between the two graphs and we write $G_1 \neq_\mathcal{M} G_2$.

Families of models can be compared in their expressive power in terms of graph disambiguation:

**Definition 5.C.3** (At least as expressive). Let $\mathcal{M}_1, \mathcal{M}_2$ be two model families. We say $\mathcal{M}_1$ is at least as expressive as $\mathcal{M}_2$ iff $\forall G_1 = (V_1, E_1), G_2 = (V_2, E_2), G_1 \neq_{\mathcal{M}_2} G_2 \implies G_1 \neq_{\mathcal{M}_1} G_2$. We write $\mathcal{M}_1 \sqsubseteq \mathcal{M}_2$.

Intuitively, $\mathcal{M}_1$ is at least as expressive as $\mathcal{M}_2$ if when $\mathcal{M}_2$ discriminates a pair of graphs, also $\mathcal{M}_1$ does. Additionally, a family can be *strictly* more expressive than another:

**Definition 5.C.4** (Strictly more expressive). Let $\mathcal{M}_1, \mathcal{M}_2$ be two model families. We say $\mathcal{M}_1$ is strictly more expressive than $\mathcal{M}_2$ iff $\mathcal{M}_1 \sqsubseteq \mathcal{M}_2 \wedge \mathcal{M}_2 \not\sqsubseteq \mathcal{M}_1$. Equivalently, $\mathcal{M}_1 \sqsubseteq \mathcal{M}_2 \wedge \exists G_1 = (V_1, E_1), G_2 = (V_2, E_2)$, s.t. $G_1 \neq_{\mathcal{M}_1} G_2 \wedge G_1 =_{\mathcal{M}_2} G_2$.

Intuitively, $\mathcal{M}_1$ is strictly more expressive than $\mathcal{M}_2$ if $\mathcal{M}_1$ is at least as expressive as $\mathcal{M}_2$ and there exist pairs of graphs that $\mathcal{M}_1$ distinguishes but $\mathcal{M}_1$ does not.

Many graph algorithms and models operate by generating *node colorings* or representations. These can be gathered into multisets of colors that are compared to assess whether two graphs are non-isomorphic. Other than convenient, in these cases it is interesting to characterise the discriminative power at the level of nodes by means of the concept of color refinement [26, 30, 176].

**Definition 5.C.5** (Color refinement). Let $G = (V, E)$ be a graph and $C, D$ two coloring functions. Coloring $D$ refines colouring $C$ when $\forall v, w \in V, D(v) = D(w) \implies C(v) = C(w)$.

Essentially, when $D$ refines $C$, if any two nodes are assigned the same color by $D$, the same holds for $C$. Equivalently, if two nodes are distinguished by $C$ (because they are assigned different colors), then they are also distinguished by $D$. When, for any graph, $D$ refines $C$, then we write $D \sqsubseteq C$ and, when also the opposite holds, ($C \sqsubseteq D$), we then write $D \equiv C$. As an example, for any $t \geqslant 0$ it can be shown that, on any graph, the coloring generated by the 1-WL algorithm at round $t+1$ refines that at round $t$, that is $C^{(t+1)} \sqsubseteq C^{(t)}$; this being essentially due to the injectivity property of the RELABEL function.

Importantly, as we were anticipating above, this concept can be directly translated into graph discrimination as long as graphs are represented by the multiset of their vertices' colors, or an *injection* thereof. This link, which explains the use of the same symbol to refer to the concepts of color refinement and discriminative power, is explicitly shown, for example, in Bevilacqua et al. [26] and Bodnar et al. [31]. More concretely, it can be shown that, if coloring $D$ refines coloring $C$, then the algorithm which generates $D$ is at least as expressive as the one generating $C$, as long as multisets of node colours are directly compared to discriminate between graphs, or they are first encoded by a multiset injection before the comparison is carried out. In the following we will resort to the concept of color refinement to prove some of our theoretical results. This approach is not only practically convenient for the required derivations, but it also informs us on the discriminative power models have at the level of nodes, something which is of relevance to us given our focus on node-classification tasks.

Furthermore, even though Dir-GNN outputs node-wise embeddings, it can be augmented with a global readout function to generate a single graph-wise embeddings $\mathbf{x}_G = \text{READOUT}\left(\{\!\!\{\mathbf{x}_i^{(K)} : i \in V\}\!\!\}\right)$. We will assume that all models discussed in this section are augmented with a global readout function.

### 5.C.3 MPNNs on Directed Graphs

Before moving forward to prove our expressiveness results, let us introduce the families of architectures we compare with. These embody straightforward approaches to adapt MPNNs to directed graphs.

Let MPNN-D be a model that performs message-passing by only propagating messages in accordance with the directionality of edges. Its layers can be defined as follows:

$$\begin{aligned}\mathbf{m}_i^{(k)} &= \text{AGG}^{(k)}\left(\{\!\!\{(\mathbf{x}_j^{(k-1)}, \mathbf{x}_i^{(k-1)}) : j \in N_\rightarrow(i)\}\!\!\}\right) \\ \mathbf{x}_i^{(k)} &= \text{COM}^{(k)}\left(\mathbf{x}_i^{(k-1)}, \mathbf{m}_i^{(k)}\right)\end{aligned} \quad (5.8)$$

Instead, let MPNN-U be a model which propagates messages equally along any incident edge, independent of their directionality. Its layers can be defined as follows:

$$\begin{aligned}\mathbf{m}_i^{(k)} &= \text{AGG}^{(k)}\left(\{\!\!\{(\mathbf{x}_j^{(k-1)}, \mathbf{x}_i^{(k-1)}) : j \in N_\rightarrow(i)\}\!\!\} \cup \{\!\!\{(\mathbf{x}_j^{(k-1)}, \mathbf{x}_i^{(k-1)}) : j \in N_\leftarrow(i)\}\!\!\}\right) \\ \mathbf{x}_i^{(k)} &= \text{COM}^{(k)}\left(\mathbf{x}_i^{(k-1)}, \mathbf{m}_i^{(k)}\right)\end{aligned}$$
$$(5.9)$$

Note that if there are no bi-directional edges, MPNN-U is equivalent to first converting the graph to its undirected form (where the edge set is redefined as $E^{(u)} = \{(i,j) : (i,j) \in E \vee (j,i) \in E\}$) and then running an undirected MPNN( Eq. (2.4)). In practice, we observe that the number of bi-directional edges is generally small on average, while *extremely* small on specific datasets (see Tab. 5.F.1). In these cases, we expect the empirical performance of the two approaches to be close to each other. We remark that, in our experiments, we opt for the latter strategy as it is easier and more efficient to implement.

We can now formally define families for the models we will be comparing.

**Definition 5.C.6** (Model families). Let $\mathcal{M}_{\text{MPNN-D}}$ be the family of Message Passing Neural Networks on the directed graph (Eq. (5.8)), $\mathcal{M}_{\text{MPNN-U}}$ that of Message Passing Neural Networks on the undirected form of the graph (Eq. (5.9)), and $\mathcal{M}_{\text{Dir-GNN}}$ that of Dir-GNN models (Eq. (5.4)).

### 5.C.4 COMPARISON WITH D-WL

We start by restating Theorem 5.4.1 more formally:

**Theorem 5.C.7.** *$\mathcal{M}_{Dir-GNN}$ is as expressive as D-WL if $\text{AGG}_{\rightarrow}^{(k)}$, $\text{AGG}_{\leftarrow}^{(k)}$, and $\text{COM}^{(k)}$ are injective for all $k$ and node representations are aggregated via an injective READOUT function.*

We now prove the theorem by showing that D-WL and Dir-GNN (under the hypotheses of the theorem) are equivalent in their expressive power. We will show this in terms of color refinement and, in particular, by showing that, not only the D-WL coloring at any round $t$ refines that induced by any Dir-GNN at the same time step, but also that, when Dir-GNN's components are injective, the opposite holds.

*Proof of Theorem 5.C.7.* Let us begin by showing that Dir-GNN is upper-bounded in expressive power by the D-WL test. We do this by showing that, at any $t \geq 0$, the D-WL coloring $D^{(t)}$ refines the coloring induced by the representations of any Dir-GNN, that is, on any graph $G = (V, E)$, $\forall v, w \in V, \quad D^{(t)}(v) = D^{(t)}(w) \implies h_v^{(t)} = h_w^{(t)}$, where $h_v^{(t)}$ refers to the representation of node $v$ in output from any Dir-GNN at layer $t > 0$. For $t = 0$ nodes are populated with a constant color: $\forall v \in V : D_v^{(0)} = \bar{c}$, or an appropriate encoding thereof in the case of the Dir-GNN $h_v^{(0)} = \text{enc}(\bar{c})$.

We proceed by induction. The base step trivially holds for $t = 0$ given how nodes are initialised. As for the recursion step, let us assume the thesis hold for $t > 0$; we seek to prove it also hold for $t + 1$, showing that $\forall v, w \in V, \quad D^{(t+1)}(v) = D^{(t+1)}(w) \implies h_v^{(t+1)} = h_w^{(t+1)}$. $D^{(t+1)}(v) = D^{(t+1)}(w)$ implies the equality of the inputs of the RELABEL function given it is injective. That is: $D^{(t)}(v) = D^{(t)}(w)$, $\{\!\!\{D^{(t)}(u) : u \in N_\rightarrow(v)\}\!\!\} = \{\!\!\{D^{(t)}(u) : u \in N_\rightarrow(w)\}\!\!\}$, and $\{\!\!\{D^{(t)}(u) : u \in N_\leftarrow(v)\}\!\!\} = \{\!\!\{D^{(t)}(u) : u \in N_\leftarrow(w)\}\!\!\}$. By the induction hypothesis, we immediately get $h_v^{(t)} = h_w^{(t)}$. Also, the induction hypothesis, along with Bevilacqua et al., Lemma 2, gives us: $\{\!\!\{h_u^{(t)} : u \in N_\rightarrow(v)\}\!\!\} = \{\!\!\{h_u^{(t)} : u \in N_\rightarrow(w)\}\!\!\}$, and $\{\!\!\{h_u^{(t)} : u \in N_\leftarrow(v)\}\!\!\} = \{\!\!\{h_u^{(t)} : u \in N_\leftarrow(w)\}\!\!\}$. Given that $h_v^{(t)} = h_w^{(t)} = \bar{h}$, we also have $\{\!\!\{(h_u^{(t)}, h_v^{(t)}) : u \in N_\rightarrow(v)\}\!\!\} = \{\!\!\{(h_u^{(t)}, h_w^{(t)}) : u \in N_\rightarrow(w)\}\!\!\}$, and $\{\!\!\{(h_u^{(t)}, h_v^{(t)}) : u \in N_\leftarrow(v)\}\!\!\} = \{\!\!\{(h_u^{(t)}, h_w^{(t)}) : u \in N_\leftarrow(w)\}\!\!\}$: it would be sufficient, for example, to construct the well-defined function $\varphi : h \mapsto (h, \bar{h})$ and invoke Bevilacqua et al., Lemma 3. These all represents the only inputs to a Dir-GNN layer – the two $\text{AGG}^{(t)}$ and the $\text{COM}^{(t)}$ function in particular. Being well defined functions, they must return equal outputs for equal inputs, so that $h_v^{(t+1)} = h_w^{(t+1)}$.

In a similar way, we show that, when AGG and COM functions are injective, the opposite hold, that is, $\forall v, w \in V, h_v^{(t)} = h_w^{(t)} \implies D^{(t)}(v) = D^{(t)}(w)$. The base step holds for $t = 0$ for the same motivations above. Let us assume the thesis holds for $t > 0$ and seek to show that for $t + 1$, $\forall v, w \in V$, $h_v^{(t+1)} = h_w^{(t+1)} \implies D^{(t+1)}(v) = D^{(t+1)}(w)$. If $h_v^{(t+1)} = h_w^{(t+1)}$, then $\text{COM}^{(t)}\left(\mathbf{h}_v^{(t)}, \mathbf{m}_{v,\rightarrow}^{(t)}, \mathbf{m}_{v,\leftarrow}^{(t)}\right) = \text{COM}^{(t)}\left(\mathbf{h}_w^{(t)}, \mathbf{m}_{w,\rightarrow}^{(t)}, \mathbf{m}_{w,\leftarrow}^{(t)}\right)$. As $\text{COM}^{(t)}$ is injective, it must also hold $\mathbf{h}_v^{(t)} = \mathbf{h}_w^{(t)}$, which, by the induction hypothesis, gives $D^{(t)}(v) = D^{(t)}(w)$. Furthermore, by the same argument, we must also have $\mathbf{m}_{v,\rightarrow}^{(t)} = \mathbf{m}_{w,\rightarrow}^{(t)}$, and $\mathbf{m}_{v,\leftarrow}^{(t)} = \mathbf{m}_{w,\leftarrow}^{(t)}$. At this point we recall that, for any node $v$, $\mathbf{m}_{v,\rightarrow}^{(t)} = \text{AGG}_{\rightarrow}^{(t)}(\{\!\!\{(h_u^{(t)}, h_v^{(t)}) : u \in N_{\rightarrow}(v)\}\!\!\})$ and $\mathbf{m}_{v,\leftarrow}^{(t)} = \text{AGG}_{\leftarrow}^{(t)}(\{\!\!\{(h_u^{(t)}, h_v^{(t)}) : u \in N_{\leftarrow}(v)\}\!\!\})$, where, by our assumption, $\text{AGG}_{\leftarrow}^{(t)}, \text{AGG}_{\rightarrow}^{(t)}$ are injective. This implies the equality between the multisets in input, i.e. $\{\!\!\{(h_u^{(t)}, h_v^{(t)}) : u \in N_{\rightarrow}(v)\}\!\!\} = \{\!\!\{(h_u^{(t)}, h_w^{(t)}) : u \in N_{\rightarrow}(w)\}\!\!\}$, and $\{\!\!\{(h_u^{(t)}, h_v^{(t)}) : u \in N_{\leftarrow}(v)\}\!\!\} = \{\!\!\{(h_u^{(t)}, h_w^{(t)}) : u \in N_{\leftarrow}(w)\}\!\!\}$. From these equalities it clearly follows $\{\!\!\{h_u^{(t)} : u \in N_{\rightarrow}(v)\}\!\!\} = \{\!\!\{h_u^{(t)} : u \in N_{\rightarrow}(w)\}\!\!\}$, and $\{\!\!\{h_u^{(t)} : u \in N_{\leftarrow}(v)\}\!\!\} = \{\!\!\{h_u^{(t)} : u \in N_{\leftarrow}(w)\}\!\!\}$ – one can invoke Bevilacqua et al., Lemma 3 with the well defined function $\varphi : (h_1, h_2) \mapsto h_1$. Again, by the induction hypothesis, and Bevilacqua et al., Lemma 2, we have $\{\!\!\{D^{(t)}(u) : u \in N_{\leftarrow}(v)\}\!\!\} = \{\!\!\{D^{(t)}(u) : u \in N_{\leftarrow}(w)\}\!\!\}$ and $\{\!\!\{D^{(t)}(u) : u \in N_{\rightarrow}(v)\}\!\!\} = \{\!\!\{D^{(t)}(u) : u \in N_{\rightarrow}(w)\}\!\!\}$. Finally, as all and only inputs to the RELABEL function are equal, $D^{(t+1)}(v) = D^{(t+1)}(w)$. The proof then terminates: as the READOUT function is assumed to be injective, having proved the refinement holds at the level of nodes, this is enough to also state that, if two graphs are distinguished by D-WL they are also distinguished by a Dir-GNN satisfying the injectivity assumptions above. □

As for the existence and implementation of these injective components, constructions can be found in Xu et al. [264] and Corso et al. [56]. In particular, in Xu et al., Lemma 5, the authors show that, for a countable $\mathcal{X}$, there exist maps $f : \mathcal{X} \rightarrow \mathbb{R}^n$ such that function $h : X \mapsto \sum_{x \in X} f(x)$ is injective for subsets $X \subset \mathcal{X}$ of bounded cardinality, and any multiset function $g$ can be decomposed as $g(X) = \varphi\left(\sum_{x \in X} f(x)\right)$ for some function $\varphi$. As $\mathcal{X}$ is countable, there always exists an injection $Z : x \rightarrow \mathbb{N}$, and function $f$ can be constructed, for example, as $f(x) = N^{-Z(x)}$, with $N$ being the maximum (bounded) cardinality of subsets $X \subset \mathcal{X}$. These constructions are used to build multiset aggregators in the GIN architecture Xu et al. [264] when operating on features from a countable set and neighbourhoods of bounded size. Under the same assumptions, the same constructions can be readily adapted to express the aggregators $\text{AGG}_{\rightarrow}^{(k)}, \text{AGG}_{\leftarrow}^{(k)}$ as well as READOUT in our Dir-GNN. Similarly to the above, under the same assumptions, injective maps for elements $(c, X), c \in \mathcal{X}, X \subset \mathcal{X}$ can be constructed as $h(c, X) = (1+\epsilon)f(c) + \sum_{x \in X} f(x)$ for infinitely many choice of $\epsilon$, including all irrational numbers, and any function $g$ on couples $(c, X)$ can be decomposed as $g(c, X) = \varphi\left((1+\epsilon)f(c) + \sum_{x \in X} f(x)\right)$ Xu et al., Corollary 6. The same approach can be extended to our use-case. In fact, for irrationals $\epsilon_X, \epsilon_Y$, an injection on triple $(c, X, Y)$ (with $c \in \mathcal{X}, X, Y \subset \mathcal{X}$ of bounded size and $\mathcal{X}$ countable) can be built as $h(c, X, Y) = \ell\left((1 + \epsilon_X)f_X(c) + \sum_{x \in X} f_X(x), (1 + \epsilon_Y)f_Y(c) + \sum_{y \in Y} f_Y(y)\right)$, where $\ell$ is an injection on a countable set and $(1 + \epsilon_X)f_X(c) + \sum_{x \in X} f_X(x), (1 + \epsilon_Y)f_Y(c) + \sum_{y \in Y} f_Y(y)$ realise injections over couples $(c, X), (c, Y)$ as described above. This construction can be used to express the $\text{COM}^{(k)}$ components of Dir-GNN. In practice, in view of the Universal Approximation Theorem (UAT) Hornik et al. [110], Xu et al. [264] propose to use Multi-Layer Perceptrons (MLPs) to learn the required components described above, functions $f$ and $\varphi$ in particular. We note that, in order to resort to the original statement of the UAT, this approach additionally requires boundedness of set $\mathcal{X}$ itself. Similar practical parameterizations can be used to build our desired Dir-GNN layers. Last, we refer readers to Corso et al. [56] for constructions which can be adopted in the case where initial node features have a continuous, uncountable support.

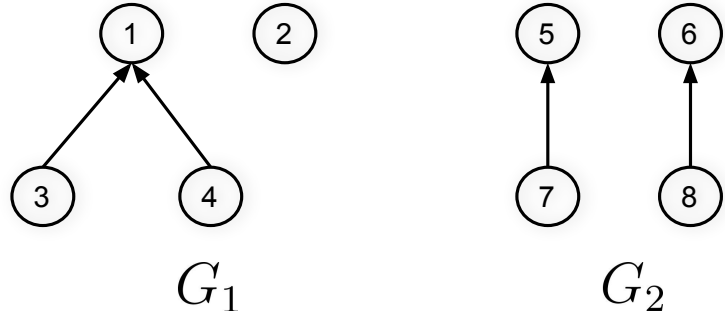

Figure 5.C.1: Two non-isomorphic directed graphs that cannot be distinguished by any MPNN-D model but can be distinguished by Dir-GNN.

### 5.C.5 Comparison with MPNNs

In this subsection we prove Theorem 5.4.2, which we restate more formally and split into two separate parts, one regarding MPNN-D and the other regarding MPNN-U. We start by proving that Dir-GNN is stricly more expressive than MPNN-D, i.e an MPNN which operates on directed graphs by only propagating messages according to the directionality of edges:

**Theorem 5.C.8.** $\mathcal{M}_{\text{Dir-GNN}}$ *is strictly more powerful than* $\mathcal{M}_{\text{MPNN-D}}$.

We begin by first proving the following lemmas:

**Lemma 5.C.9.** $\mathcal{M}_{\text{Dir-GNN}}$ *is at least as expressive as* $\mathcal{M}_{\text{MPNN-D}}$ *(*$\mathcal{M}_{\text{Dir-GNN}} \sqsubseteq \mathcal{M}_{\text{MPNN-D}}$*)*.

*Proof of Lemma 5.C.9.* We prove this Lemma by noting that the Dir-GNN architecture generalizes that of an MPNN-D, so that a Dir-GNN model can (learn to) simulate a standard MPNN-D by adopting particular weights. Specifically, Dir-GNN defaults to MPNN-D (which only sends messages along the out edges) if $\text{COM}^{(k)}\left(\mathbf{x}_i^{(k-1)}, \mathbf{m}_{i,\leftarrow}^{(k)}, \mathbf{m}_{i,\rightarrow}^{(k)}\right) = \text{COM}^{(k)}\left(\mathbf{x}_i^{(k-1)}, \mathbf{m}_{i,\rightarrow}^{(k)}\right)$, i.e. COM ignores in-messages, and the two readout modules coincide. Importantly, the direct implication of the above is that whenever an MPNN-D model distinguishes two graphs, then there exists a Dir-GNN which can implement such a model and then discriminate the two graphs as well. □

**Lemma 5.C.10.** *There exist graph pairs discriminated by a Dir-GNN model which are not discriminated by any MPNN-D model.*

*Proof of Lemma 5.C.10.* Let $G_1$ and $G_2$ be the non-isomorphic graphs illustrated in Fig. 5.C.1. To confirm that they are not isomorphic, simply note that node 1 in $G_1$ has an in-degree of two, while no node in $G_2$ has an in-degree of two.

To prove that no MPNN-D model can distinguish between the two graphs, we will show that any MPNN-D induce the same coloring for the two graphs. In particular, we will show that, if $C^{(t)}(v)$ refers to the representation a MPNN-D computes for node $v$ at time step $t$, then $C^{(t)}(1) = C^{(t)}(2) = C^{(t)}(5) = C^{(t)}(6)$ and $C^{(t)}(3) = C^{(t)}(4) = C^{(t)}(7) = C^{(t)}(8)$ for any $t \geq 0$.

We proceed by induction. The base step trivially holds for $t = 0$ given that nodes are all initialised with the same color. As for the inductive step, let us assume that the statement holds for $t$ and prove that it also holds for $t + 1$. Assume $C^{(t)}(1) = C^{(t)}(2) = C^{(t)}(5) = C^{(t)}(6)$ and $C^{(t)}(3) = C^{(t)}(4) = C^{(t)}(7) = C^{(t)}(8)$ (induction hypothesis). Then we have:

$$C^{(t+1)}(1) = \text{COM}^{(t)}\Big(C^{(t)}(1), \text{AGG}^{(t)}(\{\!\{\}\!\})\Big)$$
$$C^{(t+1)}(2) = \text{COM}^{(t)}\Big(C^{(t)}(2), \text{AGG}^{(t)}(\{\!\{\}\!\})\Big)$$
$$C^{(t+1)}(5) = \text{COM}^{(t)}\Big(C^{(t)}(5), \text{AGG}^{(t)}(\{\!\{\}\!\})\Big)$$
$$C^{(t+1)}(6) = \text{COM}^{(t)}\Big(C^{(t)}(6), \text{AGG}^{(t)}(\{\!\{\}\!\})\Big)$$

The induction hypothesis then gives us that $C^{(t+1)}(1) = C^{(t+1)}(2) = C^{(t+1)}(5) = C^{(t+1)}(6)$. As for the other nodes, we have:

$$C^{(t+1)}(3) = \text{COM}^{(t)}\Big(C^{(t)}(3), \text{AGG}^{(t)}(\{\!\{(C^{(t)}(3), C^{(t)}(1))\}\!\})\Big)$$
$$C^{(t+1)}(4) = \text{COM}^{(t)}\Big(C^{(t)}(4), \text{AGG}^{(t)}(\{\!\{(C^{(t)}(4), C^{(t)}(1))\}\!\})\Big)$$
$$C^{(t+1)}(7) = \text{COM}^{(t)}\Big(C^{(t)}(7), \text{AGG}^{(t)}(\{\!\{(C^{(t)}(7), C^{(t)}(5))\}\!\})\Big)$$
$$C^{(t+1)}(8) = \text{COM}^{(t)}\Big(C^{(t)}(8), \text{AGG}^{(t)}(\{\!\{(C^{(t)}(8), C^{(t)}(6))\}\!\})\Big)$$

The induction hypothesis then gives us that $C^{(t+1)}(3) = C^{(t+1)}(4) = C^{(t+1)}(7) = C^{(t+1)}(8)$. Importantly, the above holds for any parameters of the COM$^{(t)}$ and AGG$^{(t)}$ functions. As *any* MPNN-D will always compute the same set of node representations for the two graphs, it follows that no MPNN-D can disambiguate between the two graphs, no matter the way they are aggregated. To conclude our proof, we show that there exists Dir-GNN models that can discriminate the two graphs. In view of Theorem 5.4.1, it is enough to show that the two graphs are disambiguated by D-WL. Applying D-WL to the two graphs leads to different colorings after two iterations (see Tab. 5.C.1), so the D-WL algorithm terminates deeming the two graphs non-isomorphic. Then, by Theorem 5.4.1, there exist Dir-GNNs which distinguish them. In fact, it is easy to even construct simple 1-layer architecture that can assign the two graphs distinct representations, an exercise which we leave to the reader. Importantly, note how Dir-GNN can distinguish between the two graphs hinging on the discrimination of non-isomorphic nodes such as 1, 2, something no MPNN-D is capable of doing. □

With the two results above prove Theorem 5.C.8.

*Proof of Theorem 5.C.8.* The theorem follows directly from Lemmas 5.C.9 and 5.C.10. □

Next, we focus on the comparison with MPNN-U, i.e. an MPNN on the undirected form of the graph:

**Theorem 5.C.11.** $\mathcal{M}_{\text{Dir}-\text{GNN}}$ *is strictly more expressive than* $\mathcal{M}_{\text{MPNN}-\text{U}}$.

Instrumental to us is to consider a variant of the 1-WL test MPNN-U can be regarded as the neural counterpart of. In the following we will show that such a variant, which we call U-WL, generates colorings which refine the ones induced by *any* MPNN-U and that, in turn, are refined by the D-WL test. In view of Theorem 5.4.1, this will be enough to show that there exists Dir-GNNs refining any MPNN-U instantiation, so that, ultimately, $\mathcal{M}_{\text{Dir-GNN}} \sqsubseteq \mathcal{M}_{\text{MPNN-U}}$.

| Iteration | Node 1 | Node 2 | Node 3 | Node 4 | Node 5 | Node 6 | Node 7 | Node 8 |
|---|---|---|---|---|---|---|---|---|
| 1 | A | A | A | A | A | A | A | A |
| 2 | B | C | C | D | E | E | C | C |

| $M(v)$ | RELABEL(M(v)) |
|---|---|
| initialize | A |
| $(A, \{\}, \{(A, A), (A, A)\})$ | B |
| $(A, \{(A, A)\}, \{\})$ | C |
| $(A, \{\}, \{\})$ | D |
| $(A, \{\}, \{(A, A)\})$ | E |

Table 5.C.1: Node colorings at different iterations, as well as the RELABEL hash function, when applying D-WL to the two graphs in Fig. 5.C.1.

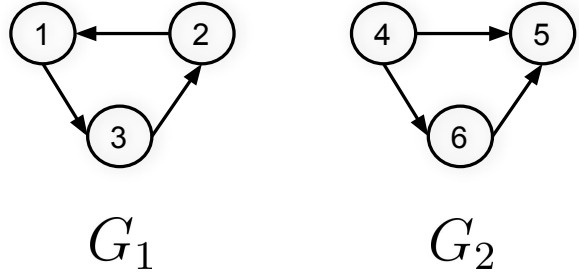

Figure 5.C.2: Two non-isomorphic directed graphs that cannot be distinguished by any MPNN-U model but can be distinguished by Dir-GNN.

**Lemma 5.C.12.** $\mathcal{M}_{\text{Dir-GNN}}$ *is at least as expressive as* $\mathcal{M}_{\text{MPNN-U}}$ *(*$\mathcal{M}_{\text{Dir-GNN}} \sqsubseteq \mathcal{M}_{\text{MPNN-U}}$*)*.

*Proof of Lemma 5.C.12.* Let us start by introducing the U-WL test, which, on an undirected graph, refines node colors as:

$$A^{(t+1)}(v) = \text{RELABEL}\left(A^{(t)}(v), \{\!\!\{A^{(t)}(u) : u \in N_{\rightarrow}(v)\}\!\!\} \cup \{\!\!\{A^{(t)}(u) : u \in N_{\leftarrow}(v)\}\!\!\}\right),$$

that is, by the gathering neighbouring colors from each incident edge, independent of its direction. It is easy to show that U-WL generates a coloring that, at any round $t \geq 0$ is refined by the coloring generated by D-WL, i.e., for any graph $G = (V, E)$ it holds that $\forall v, w \in V$, $D^{(t)}(v) = D^{(t)}(w) \implies A^{(t)}(v) = A^{(t)}(w)$, where $D$ refers to the coloring of D-WL. Again, proceeding by induction, we have the following. First, the base step hold trivially for $t = 0$. We assume the thesis holds true for $t$ and seek to show it also holds for $t + 1$. If $D^{(t+1)}(v) = D^{(t+1)}(w)$ then, by the injectivity of RELABEL we must have $D^{(t)}(v) = D^{(t)}(w)$, which implies $A^{(t)}(v) = A^{(t)}(w)$ via the induction hypothesis. Additionally, we have $\{\!\!\{D^{(t)}(u) : u \in N_{\rightarrow}(v)\}\!\!\} = \{\!\!\{D^{(t)}(u) : u \in N_{\rightarrow}(w)\}\!\!\}$, and $\{\!\!\{D^{(t)}(u) : u \in N_{\leftarrow}(v)\}\!\!\} = \{\!\!\{D^{(t)}(u) : u \in N_{\leftarrow}(w)\}\!\!\}$ which, by the induction hypothesis and Bevilacqua et al., Lemma 2, gives $\mathcal{A}^{(t)}_{\rightarrow,v} = \{\!\!\{A^{(t)}(u) : u \in N_{\rightarrow}(v)\}\!\!\} = \{\!\!\{A^{(t)}(u) : u \in N_{\rightarrow}(w)\}\!\!\} = \mathcal{A}^{(t)}_{\rightarrow,w}$, and $\mathcal{A}^{(t)}_{\leftarrow,v} = \{\!\!\{A^{(t)}(u) : u \in N_{\leftarrow}(v)\}\!\!\} = \{\!\!\{A^{(t)}(u) : u \in$ $N_\leftarrow(w)\}\!\} = \mathcal{A}^{(t)}_{\leftarrow,w}$. From these equalities we then derive $\mathcal{A}^{(t)}_{\rightarrow,v} \cup \mathcal{A}^{(t)}_{\leftarrow,v} = \mathcal{A}^{(t)}_{\rightarrow,w} \cup \mathcal{A}^{(t)}_{\leftarrow,w}$. Indeed, let us suppose that, instead, $\mathcal{A}^{(t)}_{\rightarrow,v} \cup \mathcal{A}^{(t)}_{\leftarrow,v} \neq \mathcal{A}^{(t)}_{\rightarrow,w} \cup \mathcal{A}^{(t)}_{\leftarrow,w}$ and that, w.l.o.g., this is due by the existence of a color $\bar{a}$ such that its number of appearances in $\mathcal{A}^{(t)}_{\rightarrow,w} \cup \mathcal{A}^{(t)}_{\leftarrow,w}$ is larger than that in $\mathcal{A}^{(t)}_{\rightarrow,v} \cup \mathcal{A}^{(t)}_{\leftarrow,v}$. We write $\#^{\mathcal{A}^{(t)}_{\rightarrow,w} \cup \mathcal{A}^{(t)}_{\leftarrow,w}}(\bar{a}) > \#^{\mathcal{A}^{(t)}_{\rightarrow,v} \cup \mathcal{A}^{(t)}_{\leftarrow,v}}(\bar{a})$. Then, as these are all multisets, we can rewrite $\#^{\mathcal{A}^{(t)}_{\rightarrow,w}}(\bar{a}) + \#^{\mathcal{A}^{(t)}_{\leftarrow,w}}(\bar{a}) > \#^{\mathcal{A}^{(t)}_{\rightarrow,v}}(\bar{a}) + \#^{\mathcal{A}^{(t)}_{\leftarrow,v}}(\bar{a})$. Since $\mathcal{A}^{(t)}_{\rightarrow,w} = \mathcal{A}^{(t)}_{\rightarrow,v}$, we must have that $\#^{\mathcal{A}^{(t)}_{\rightarrow,w}}(\bar{a}) = \#^{\mathcal{A}^{(t)}_{\rightarrow,v}}(\bar{a})$, which leads to necessarily having $\#^{\mathcal{A}^{(t)}_{\leftarrow,w}}(\bar{a}) \neq \#^{\mathcal{A}^{(t)}_{\leftarrow,v}}(\bar{a})$. However, this entails a contradiction, because by hypothesis we had that $\mathcal{A}^{(t)}_{\leftarrow,w} = \mathcal{A}^{(t)}_{\leftarrow,v}$. Last, given that $A^{(t)}(v) = A^{(t)}(w)$, and $\mathcal{A}^{(t)}_{\rightarrow,v} \cup \mathcal{A}^{(t)}_{\leftarrow,v} = \mathcal{A}^{(t)}_{\rightarrow,w} \cup \mathcal{A}^{(t)}_{\leftarrow,w}$, being the only inputs to the RELABEL function in U-WL, then we also have that $A^{(t+1)}(v) = A^{(t+1)}(w)$, concluding the proof of the refinement.

| Iteration | Node 1 | Node 2 | Node 3 | Node 4 | Node 5 | Node 6 |
|---|---|---|---|---|---|---|
| 1 | A | A | A | A | A | A |
| 2 | B | B | B | C | B | D |

| M(v) | RELABEL(M(v)) |
|---|---|
| initialize | A |
| $(A, \{\!\{(A, A)\}\!\}, \{\!\{(A, A)\}\!\})$ | B |
| $(A, \{\!\{(A, A), (A, A)\}\!\}, \{\!\{\}\!\})$ | C |
| $(A, \{\!\{\}\!\}, \{\!\{(A, A), (A, A)\}\!\})$ | D |

Table 5.C.2: Node colorings at different iterations, , as well as the RELABEL hash function, when applying D-WL to the two graphs in Fig. 5.C.2.

Now, in view of Theorem 5.4.1, it is sufficient to show that U-WL refines the coloring induced by any MPNN-U; the lemma will then follow by transitivity. We want to show that $t \geq 0$, the U-WL coloring $A^{(t)}$ refines the coloring induced by the representations of any MPNN-U, that is, on any graph $G = (V, E)$, $\forall v, w \in V$, $A^{(t)}(v) = A^{(t)}(w) \implies h^{(t)}_v = h^{(t)}_w$, with $h^{(t)}_v$ referring to the representation an MPNN-U assigns to node $v$ at time step $t$. The thesis is easily proved. It clearly holds for $t = 0$ if initial node representations are produced by a well-defined function enc : $\bar{c} \mapsto \text{enc}(\bar{c})$. Then, if we assume the thesis holds for $t > 0$, we can show it also holds for $t + 1$. Indeed, if $A^{(t+1)}(v) = A^{(t+1)}(w)$, from the injectivity of RELABEL, it follows that $A^{(t)}(v) = A^{(t)}(w)$ and $\mathcal{A}^{(t)}_{\rightarrow,v} \cup \mathcal{A}^{(t)}_{\leftarrow,v} = \mathcal{A}^{(t)}_{\rightarrow,w} \cup \mathcal{A}^{(t)}_{\leftarrow,w}$. By the induction hypothesis, we have $h^{(t)}_v = h^{(t)}_w$ and, jointly due to Bevilacqua et al., Lemma 2, $\{\!\{h^{(t)}_u : u \in N_\rightarrow(v)\}\!\} \cup \{\!\{h^{(t)}_u : u \in N_\leftarrow(v)\}\!\} = \{\!\{h^{(t)}_u : u \in N_\rightarrow(w)\}\!\} \cup \{\!\{h^{(t)}_u : u \in N_\leftarrow(w)\}\!\}$. Given that $h^{(t)}_v = h^{(t)}_w = \bar{h}$, it also clearly holds that $\{\!\{(h^{(t)}_u, h^{(t)}_v) : u \in N_\rightarrow(v)\}\!\} \cup \{\!\{(h^{(t)}_u, h^{(t)}_v) : u \in N_\leftarrow(v)\}\!\} = \{\!\{(h^{(t)}_u, h^{(t)}_w) : u \in N_\rightarrow(w)\}\!\} \cup \{\!\{(h^{(t)}_u, h^{(t)}_w) : u \in N_\leftarrow(w)\}\!\}$ – it is sufficient to construct the well-defined function $\varphi : h \mapsto (h, \bar{h})$ and invoke Bevilacqua et al., Lemma 3. These are the inputs to the well-defined functions constituting the update equations of an MPNN-U architecture, eventually entailing $h^{(t+1)}_v = h^{(t+1)}_w$. □

Now, with the following lemma we show that, not only $\mathcal{M}_{\text{Dir-GNN}}$ is at least as expressive as $\mathcal{M}_{\text{MPNN-U}}$, there actually exist pairs of graphs distinguished by former family but not by the latter.

| Iteration | Node 1 | Node 2 | Node 3 | Node 4 | Node 5 | Node 6 | Node 7 | Node 8 |
|---|---|---|---|---|---|---|---|---|
| 1 | $A$ | $A$ | $A$ | $A$ | $A$ | $A$ | $A$ | $A$ |
| 2 | $E$ | $E$ | $E$ | $E$ | $E$ | $E$ | $E$ | $E$ |

| $M(v)$ | RELABEL(M(v)) |
|---|---|
| INITIALIZE | $A$ |
| $(A, \{\!\!\{(A, A)\}\!\!\}, \{\!\!\{(A, A), (A, A)\}\!\!\})$ | $E$ |

Table 5.C.3: Node colorings at different iterations, as well as the RELABEL hash function, when applying U-WL to the two graphs in Fig. 5.C.2.

**Lemma 5.C.13.** *There exist graph pairs distinguished by a Dir-GNN model which are not distinguished by any MPNN-U model.*

*Proof of Lemma 5.C.13.* Let $G_1$ and $G_2$ be the non-isomorphic graphs illustrated in Fig. 5.C.2. From Tab. 5.C.3 we observe that U-WL is not able to distinguish between the two graphs, as after the first iteration all nodes still have the same color: the U-WL is at convergence and terminates concluding that the two graphs are possibly isomorphic. From Lemma 5.C.12, we conclude that no MPNN-U can distinguish between the two graphs. On the other hand, applying D-WL to the two graphs leads to different colorings after two iterations (see Tab. 5.C.2, so the D-WL algorithm terminates deeming the two graphs non-isomorphic. Then, by Theorem 5.4.1, there exists Dir-GNNs which distinguish them. In fact, simple Dir-GNN architectures which distinguish the two graphs are easy to construct. Importantly, we note, again, how these architectures distinguish between the two graphs by disambiguating non-isomorphic nodes such as 4, 6, something no MPNN-U is capable of doing. □

Last, the two results above are sufficient to prove Theorem 5.C.11.

*Proof of Theorem 5.C.11.* The theorem follows directly from Lemmas 5.C.12 and 5.C.13. □

## 5.D Exploring Alternative Representations for Directed Graphs

One might intuitively consider the possibility of representing a directed graph equivalently through an undirected relational graph, having two relations for *original* and *inverse* edges. However, this assumption proves to be erroneous, as illustrated by the following counterexample. Take the two non-isomorphic directed graphs illustrated in Fig. 5.D.1. Surprisingly, these two directed graphs share an identical representation as an undirected relational graph, as depicted in Fig. 5.D.1. Kollias et al. [140] showed that it is however possible to equivalently represent a directed graph with an undirected graph having two nodes for each node in the original graph, representing the source and destination role of the node respectively.

## 5.E MPNN with Binary Edge Features

A possible alternative approach to address directed graphs involves using an MPNN [88] combined with binary edge features. In this case, the original directed graph is augmented with inverse edges, which are assigned a distinct binary feature compared to the original ones. Given the original directed graph $G = (V, E)$, we define the augmented graph as $G_a = (V, E_a)$ with $E_a = \{((i,j), [0\ 1]) : (i,j) \in E\} \cup \{((j,i), [1\ 0]) : (i,j) \in E\}$. An MPNN with edge features can be defined as:

$$\mathbf{m}_i^{(k)} = \text{AGG}^{(k)}\left(\{\!\!\{(\mathbf{x}_i^{(k)}, \mathbf{x}_j^{(k)}, \mathbf{e}_{ij}) : (i,j) \in E\}\!\!\}\right)$$
$$\mathbf{x}_i^{(k)} = \text{COM}^{(k)}\left(\mathbf{x}_i^{(k-1)}, \mathbf{m}_i^{(k)}\right)$$

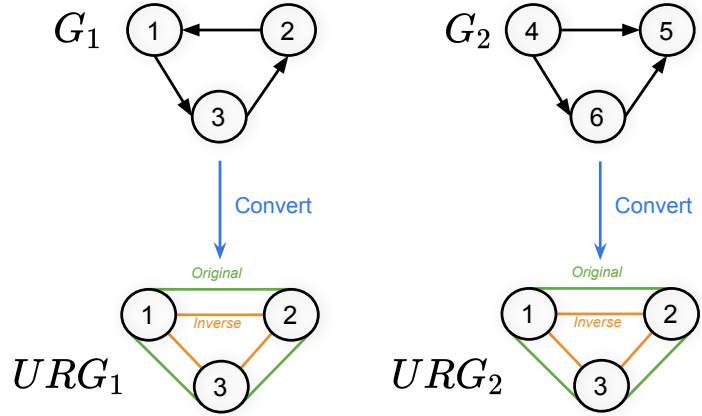

Figure 5.D.1: The two non-isomorphic directed graphs $G_1$ and $G_2$ become isomorphic when converted to an undirected relational graph. This shows that is not possible to represent directed graphs with undirected relational graphs without losing information.

However, since messages do not depend only on the source nodes but also on the destination node (through the edge features), this approach necessitates the materialization of explicit edge messages [231]. Given that there are $m$ such messages (one per edge) with dimension $d$, the memory complexity amounts to $\mathcal{O}(m \times d)$. This is significantly higher than the $\mathcal{O}(n \times d)$ memory complexity achieved by instantiations of Dir-GNN such as Dir-GCN or Dir-SAGE, and could lead to out-of-memory issues for even moderately sized datasets. Therefore, while having the same expressivity, our model has better memory complexity.

## 5.F Experimental Details

### 5.F.1 Effective Homophily of Synthetic Graphs

For the results in Fig. 2a, we generate directed synthetic graphs with various homophily levels using a modified preferential attachment process [11], inspired by Zhu et al. [294]. New nodes are incrementally added to the graphs until the desired number of nodes is achieved. Each node is assigned a class label, chosen uniformly at random among $C$ classes, and forms out-edges with exactly $m$ pre-existing nodes, where $m$ is a parameter of the process. The $m$ out-neighbors are sampled without replacement from a distribution that is proportional to both their in-degree and the class compatibility of the two nodes. Consequently, nodes with higher in-degree are more likely to receive new edges, leading to a "rich get richer" effect where a small number of highly connected "hub" nodes emerge. This results in the in-degree distribution of the generated graphs following a power-law, with heterophily controlled by the class compatibility matrix $\mathbf{H}$. In our experiments, we generate graphs comprising 1000 nodes and set $C = 5, m = 2$. Note that by construction, the generated graphs will not have any bidirectional edge.

| Dataset | # Nodes | # Edges | # Feat. | # C | Unid. Edges | Edge hom. |
|---|---|---|---|---|---|---|
| citeseer-full | 4,230 | 5,358 | 602 | 6 | 99.61% | 0.949 |
| cora-ml | 2,995 | 8,416 | 2,879 | 7 | 96.84% | 0.792 |
| ogbn-arxiv | 169,343 | 1,166,243 | 128 | 40 | 99.27% | 0.655 |
| chameleon | 2,277 | 36,101 | 2,325 | 5 | 85.01% | 0.235 |
| squirrel | 5,201 | 217,073 | 2,089 | 5 | 90.60% | 0.223 |
| arxiv-year | 169,343 | 1,166,243 | 128 | 40 | 99.27% | 0.221 |
| snap-patents | 2,923,922 | 13,975,791 | 269 | 5 | 99.98% | 0.218 |
| roman-empire | 22,662 | 44,363 | 300 | 18 | 65.24% | 0.050 |

Table 5.F.1: Statistics of the datasets used in this paper.

| Dataset | model_type | lr | # hidden_dim | # num_layers | jk | norm | dropout | $\alpha$ |
|---|---|---|---|---|---|---|---|---|
| chameleon | Dir-GCN | 0.005 | 128 | 5 | max | True | 0 | 1. |
| squirrel | Dir-GCN | 0.01 | 128 | 4 | max | True | 0 | 1. |
| arxiv-year | Dir-GCN | 0.005 | 256 | 6 | cat | False | 0 | 0.5 |
| snap-patents | Dir-GCN | 0.01 | 32 | 5 | max | True | 0 | 0.5 |
| roman-empire | Dir-SAGE | 0.01 | 256 | 5 | cat | False | 0.2 | 0.5 |

Table 5.F.2: Best hyperparameters for each dataset, determined through grid search, for our model.

### 5.F.2 Synthetic Experiment

For the results in Fig. 2b, we construct an Erdos-Renyi graph with 5000 nodes and edge probability of 0.001, where each node has a scalar feature sampled uniformly at random from $[-1, 1]$. The label of a node is set to 1 if the mean of the features of their in-neighbors is greater than the mean of the features of their out-neighbors, or zero otherwise.

### 5.F.3 Experimental Setup

Real-World datasets statistics are reported in table 5.F.1. All experiments are conducted on a GCP machine with 1 NVIDIA V100 GPU with 16GB of memory, apart from experiments on snap-patents which have been performed on a machine with 1 NVIDIA A100 GPU with 40GB of memory. The total GPU time required to conduct all the experiments presented in this paper is approximately two weeks. In all experiments, we use the Adam optimizer and train the model for 10000 epochs, using early stopping on the validation accuracy with a patience of 200 for all datasets apart from Chameleon and Squirrel, for which we use a patience of 400. We do not use regularization as it did not help on heterophilic datasets. For Citeseer-Full and Cora-ML we use random 50/25/25 splits, for OGBN-Arxiv we use the fixed split provided by OGB [113], for Chameleon and Squirrel we use the fixed GEOM-GCN splits [192], for Arxiv-Year and Snap-Patents we use the splits provided in Lim et al. [154], while for Roman-Empire we use the splits from Platonov et al. [195]. We report the mean and standard deviation of the test accuracy, computed over 10 runs in all experiments.

|  | IN_DEGREE | OUT_DEGREE | TOTAL_DEGREE |
|---|---|---|---|
| CORA_ML | 41.70% | 11.65% | 0.00% |
| CITESEER_FULL | 63.45% | 21.35% | 0.00% |
| OGBN-ARXIV | 36.62% | 10.30% | 0.00% |
| CHAMELEON | 62.06% | 0.00% | 0.00% |
| SQUIRREL | 57.60% | 0.00% | 0.00% |
| ARXIV-YEAR | 36.62% | 10.30% | 0.00% |
| SNAP-PATENTS | 23.38% | 30.16% | 6.09% |
| DIRECTED-ROMAN-EMPIRE | 0.00% | 0.00% | 0.00% |

Table 5.F.3: Percentage of nodes with either in-, out- or total-degree equal to zero.

### 5.F.4 Directionality Ablation Hyperparameters

For the ablation study in Sec. 5.7.1, we use the same hyperparameters for all models and datasets: learning_rate = 0.001, hidden_dimension = 64, num_layers = 3, norm = $True$, jk = $max$. norm refers to applying L2 normalization after each convolutional layer, which we found to be generally useful, while jk refers to the type of jumping knowledge [266] used.

### 5.F.5 Comparison with State-of-the-Art Results

To obtain the results for Dir-GNN in Tab. 3, we perform a grid search over the following hyperparameters: model_type ∈ {Dir-GCN, Dir-SAGE}, learning_rate ∈ {0.01, 0.005, 0.001, 0.0005}, hidden_dimension ∈ {32, 64, 128, 256, 512}, num_layers ∈ {2, 3, 4, 5, 6}, jk ∈ {$max, cat, none$}, norm ∈ {$True, False$}, dropout ∈ {0, 0.2, 0.4, 0.6, 0.8, 1} and $\alpha$ ∈ {0, 0.5, 1}. The best hyperparameters for each dataset are reported in table 5.F.2.

### 5.F.6 Baseline Results

GNNs for Heterophily   Results for MLP, GCN, H$_2$GCN, GPR-GNN and LINKX were taken from Lim et al. [154]. Results for Gradient Gating are taken from their paper [206]. Results for FSGNN are taken from their paper [168] for Actor, Squirrel and Chameleon, whereas we re-implement it to generate results on Arxiv-year and Snap-Patents, performing the same gridsearch outlined in Appendix 5.F.5. Results for GloGNN as well as MLP and GCN are taken from Li et al. [150]. Results on Roman-Empire are taken from Platonov et al. [195] for GCN, H$_2$GCN, GPR-GNN, FSGNN and GloGNN whereas we re-implement and generate results for MLP, LINKX, ACM-GCN and Gradient Gating performing the same gridsearch outlined in Appendix 5.F.5.

Directed GNNs   For DiGCN and MagNet, we used the classes provided by PyTorch Geometric Signed Directed library [105]. For MagNet, we tuned the learning_rate ∈ {0.01, 0.005, 0.001, 0.0005}, the hidden_dim ∈ {32, 64, 128, 256, 512}, the num_layers ∈ {2, 3, 4, 5, 6}, the $K$ parameter for its chebyshev convolution to ∈ {1, 2}, and its $q$ hyperparameter ∈ {0, 0.05, 0.10, 0.15, 0.20}. For DiGCN, we tune the learning_rate ∈ {0.01, 0.005, 0.001, 0.0005}, the hidden_dim ∈ {32, 64, 128, 256, 512}, the num_layers ∈ {2, 3, 4, 5, 6}, and the $\alpha$ ∈ {0.05, 0.10, 0.15, 0.20}.

| | CITESEER_FULL | CORA_ML | OGBN-ARXIV | CHAMELEON | SQUIRREL | ARXIV-YEAR | SNAP-PATENTS | ROMAN-EMPIRE |
|---|---|---|---|---|---|---|---|---|
| HOM. | 0.949 | 0.792 | 0.655 | 0.235 | 0.223 | 0.221 | 0.218 | 0.050 |
| GCN | 93.37±0.22 | 84.37±1.52 | 68.39±0.01 | 71.12±2.28 | 62.71±2.27 | 46.28±0.39 | 51.02±0.07 | 56.23±0.37 |
| DIR-GCN($\alpha$=0.0) | 93.21±0.41 | 84.45±1.69 | 23.70±0.20 | 29.78±1.27 | 33.03±0.78 | 50.51±0.45 | 51.71±0.06 | 42.69±0.41 |
| DIR-GCN($\alpha$=1.0) | 93.44±0.59 | 83.81±1.44 | 62.93±0.21 | 78.77±1.72 | 74.43±0.74 | 50.52±0.09 | 62.24±0.04 | 45.52±0.14 |
| DIR-GCN($\alpha$=0.5) | 92.97±0.31 | 84.21±2.48 | 66.66±0.02 | 72.37±1.50 | 67.82±1.73 | 59.56±0.16 | 71.32±0.06 | 74.54±0.71 |
| SAGE | 94.15±0.61 | 86.01±1.56 | 67.78±0.07 | 61.14±2.00 | 42.64±1.72 | 44.05±0.02 | 52.55±0.10 | 72.05±0.41 |
| DIR-SAGE($\alpha$=0.0) | 94.05±0.25 | 85.84±2.09 | 52.08±0.17 | 48.33±2.40 | 35.31±0.52 | 47.45±0.32 | 52.53±0.03 | 76.47±0.14 |
| DIR-SAGE($\alpha$=1.0) | 93.97±0.67 | 85.73±0.35 | 65.14±0.03 | 64.47±2.27 | 46.05±1.16 | 50.37±0.09 | 61.59±0.05 | 68.81±0.48 |
| DIR-SAGE($\alpha$=0.5) | 94.14±0.65 | 85.81±1.18 | 65.06±0.28 | 60.22±1.16 | 43.29±1.04 | 55.76±0.10 | 70.26±0.14 | 79.10±0.19 |
| GAT | 94.53±0.48 | 86.44±1.45 | 69.60±0.01 | 66.82±2.56 | 56.49±1.73 | 45.30±0.23 | OOM | 49.18±1.35 |
| DIR-GAT($\alpha$=0.0) | 94.48±0.52 | 86.13±1.58 | 52.57±0.05 | 40.44±3.11 | 28.28±1.02 | 46.01±0.06 | OOM | 53.58±2.51 |
| DIR-GAT($\alpha$=1.0) | 94.08±0.69 | 86.21±1.40 | 66.50±0.16 | 71.40±1.63 | 67.53±1.04 | 51.58±0.19 | OOM | 56.24±0.41 |
| DIR-GAT($\alpha$=0.5) | 94.12±0.49 | 86.05±1.71 | 66.44±0.41 | 55.57±1.02 | 37.75±1.24 | 54.47±0.14 | OOM | 72.25±0.04 |

Table 5.F.4: Ablation study comparing base MPNNs on the undirected graph versus their Dir-GNN extensions on the directed graph. We conducted experiments with $\alpha = 0$ (only in-edges), $\alpha = 1$ (only out-edges), and $\alpha = 0.5$ (both in- and out-edges, but with different weight matrices). For homophilic datasets (to the left of the dashed line), incorporating directionality does not significantly enhance or may slightly impair performance. However, for heterophilic datasets (to the right of the dashed line), the inclusion of directionality substantially improves accuracy.

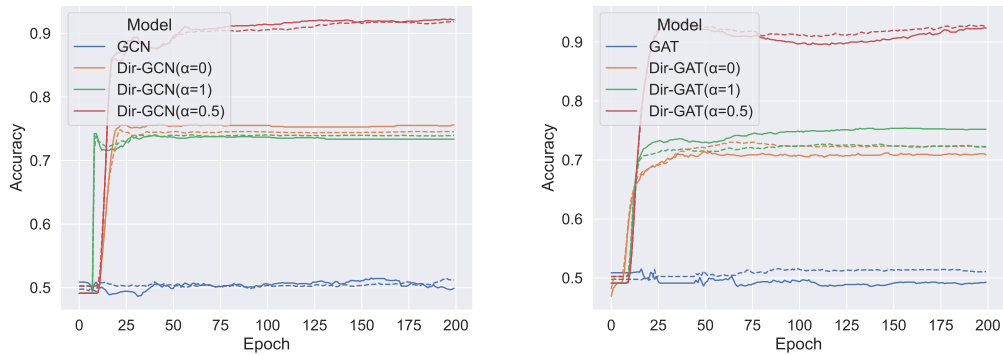

Figure 5.G.1: Validation accuracy (solid lines) and training accuracy (dashed lines) of GCN (left) and GAT (right), as well as their respective extensions using our Dir-GNN framework, on a synthetic task which requires directionality information in order to be solved.

## 5.G Additional Results

### 5.G.1 Synthetic Experiment

We also evaluate GCN and GAT (and their Dir-GNN extensions) on the synthetic task outlined in Appendix 5.F.2. Similarly to what observed for GraphSage (Fig. 2b), the Dir-GNN variant using both directions ($\alpha = 0.5$) significantly outperforms the other configurations, despite not reaching 100% accuracy. The undirected models are akin to a random classifier, whereas the models using only one directions obtain between 70% and 75% of accuracy.

### 5.G.2 Ablation Study on Using Directionality

Table 5.F.4 compares using the undirected graph vs using the directed graph with our framework with different $\alpha$. We observe that only on Chameleon and Squirrel, using only one direction of the edges, in particular the out direction, performs better than using both direction. Moreover, for these two datasets, the gap between the two directions ($\alpha = 0$ vs $\alpha = 1$) is extremely large

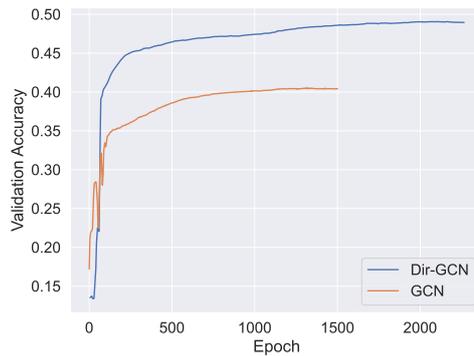

Figure 5.G.2: Performance of GCN (on the undirected version of the graph) and Dir-GCN on Arxiv-Year when using only one layer. Remarkably, directionality yields significant benefits, even in the absence of access to the homophilic directed 2-hop. This is largely attributable to the harmless heterophily exhibited by the directed graph.

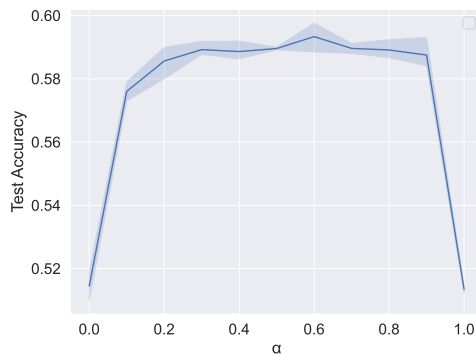

Figure 5.G.3: Dir-GNN test accuracy on Arxiv-Year for different values of the hyperparameter $\alpha$.

(more than 40% absolute accuracy). We find that this is likely due to the high number of nodes with zero in neighbors, as reported in Table 5.F.3. Chameleon and Squirrel have respectively about 62% and 57% of nodes with no in-edges: when propagating only over in edges, these nodes would get zero features. We observe a similar trend for other datasets, where $\alpha = 1$ performs generally better than $\alpha = 0$, in line with the fact that all these datasets have more nodes with zero in edges than out edges (Table 5.F.3). In general, using both in- and out- edges is the preferred solution.

### 5.G.3 Ablation Study on Using a Single Layer

In Sec. 5.3 we discuss how Arxiv-Year and Snap-Patents exhibit harmless heterophily when treating as directed. This suggest that even a 1-layer Dir-GNN model should be able to perform much better of its undirected counterpart, despite not being able to access the much more homophilic 2-hop. We verify this empirically by comparing a 1-layer GCN (on the undirected version of the graph) with a 1-layer Dir-GCN on Arxiv-Year. Fig. 5.G.2 presents the results, showing that Dir-GCN does indeed significantly outperform GCN.

5.G.4 Ablation Study on Different Values of $\alpha$

We train Dir-GNN models on Arxiv-Year with varying values of $\alpha$, using the hyperparameters outlined in Appendix 5.F.4. Fig. 5.G.3 presents the results: while a large drop is observed for $\alpha = 0$ and $\alpha = 1$, i.e. propagating messages only along one direction, the results for other values of $\alpha$ are largely similar.

# 6 Graph Neural Networks with Partially Missing Node Features

> "If I had nine of my fingers missing I wouldn't type
> any slower." – Mitch Hedberg

## 6.1 Introduction

GNNs typically operate by a message-passing mechanism [17, 89], where at each layer, nodes send their feature representations ("messages") to their neighbors. The feature representation of each node is initialized to their original features and updated by repeatedly aggregating incoming messages from neighbors. Being able to combine the topological information with feature information is what distinguishes GNNs from other purely topological learning approaches such as random walks [98, 194] or label propagation [295], and arguably what leads to their success.

GNN models typically assume a fully observed feature matrix, where rows represent nodes and columns feature channels. However, in real-world scenarios, each feature is often only observed for a subset of the nodes. For example, demographic information can be available for only a small subset of social network users, while content features are generally only present for the most active users. In a co-purchase network, not all products may have a full description associated with them. With the rising awareness around digital privacy, data is increasingly available only upon explicit user consent. In all the above cases, the feature matrix contains missing values and most existing GNN models cannot be directly applied.

While classic imputation methods [133, 155, 270] can be used to fill the missing values of the feature matrix, they are unaware of the underlying graph structure. Graph Signal Processing, a field attempting to generalize classical Fourier analysis to graphs, offers several methods that reconstruct signals on graphs [178]. However, they do not scale beyond graphs with a few thousand nodes, making them infeasible for practical applications. More recently, SAT [48], GCNMF [230] and PaGNN [123] have been proposed to adapt GNNs to the case of missing features. However, they are not evaluated at high missing feature rates ($> 90\%$), which occur in many real-world scenarios, and where we find them to suffer. Moreover, they are unable to scale to graphs with more than a few hundred thousand nodes. At the time of writing, PaGNN is the state-of-the-art method for node classification with missing features.

CONTRIBUTIONS   We present a general approach for handling missing node features in graph machine learning tasks. The framework consists of an initial diffusion-based feature reconstruction step followed by a downstream GNN. The reconstruction step is based on Dirichlet energy minimization, which leads to a diffusion-type differential equation on the graph. Discretization of this differential equation leads to a very simple, fast, and scalable iterative algorithm which we call Feature Propagation (FP). FP outperforms state-of-the-art methods on six standard node-classification benchmarks and presents the following advantages:

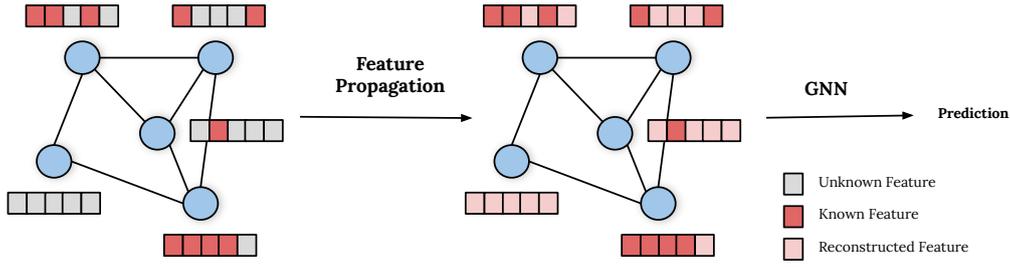

Figure 1: A diagram illustrating our Feature Propagation framework. On the left is a graph with missing node features. In the initial reconstruction step, Feature Propagation reconstructs the missing features by iteratively diffusing the known features in the graph. Subsequently, the graph and the reconstructed node features are fed into a downstream GNN model, which then produces a prediction.

- **Theoretically Motivated**: FP emerges naturally as the gradient flow minimizing the Dirichlet energy and can be interpreted as a diffusion equation on the graph with known features used as boundary conditions. This contributes to the promising direction of building continuous-time models on graphs.

- **Robust to high rates of missing features**: FP can withstand surprisingly high rates of missing features. In our experiment, we observe, on average, around 4% relative accuracy drop when up to 99% of the features are missing. In comparison, GCNMF and PaGNN have an average drop of 53.33% and 21.25% respectively. This finding has important implications, especially in scenarios where the cost of sampling (observing features on nodes) is high or sampling is not possible altogether.

- **Generic**: FP can be combined with any GNN model to solve the downstream task; in contrast, GCNMF and PaGNN are specific GCN-type models.

- **Fast and Scalable**: FP takes only around 10 seconds for the reconstruction step on OGBN-Products (a graph with ∼2.5M nodes and ∼123M edges) on a single GPU. GCNMF and PaGNN run out-of-memory on this dataset.

## 6.2 Preliminaries

Denote by $V_k := V$ the set of nodes on which the features are *known*, and by $V_u = V_k^c = V \setminus V_k$ the *unknown* ones. We further assume the ordering of the nodes such that we can write

$$\mathbf{x} = \begin{bmatrix} \mathbf{x}_k \\ \mathbf{x}_u \end{bmatrix} \quad \mathbf{A} = \begin{bmatrix} \mathbf{A}_{kk} & \mathbf{A}_{ku} \\ \mathbf{A}_{uk} & \mathbf{A}_{uu} \end{bmatrix} \quad \mathbf{\Delta} = \begin{bmatrix} \mathbf{\Delta}_{kk} & \mathbf{\Delta}_{ku} \\ \mathbf{\Delta}_{uk} & \mathbf{\Delta}_{uu} \end{bmatrix}.$$

Because the graph is undirected, $\mathbf{A}$ is symmetric and thus $\mathbf{A}_{ku}^\top = \mathbf{A}_{uk}$ and $\mathbf{\Delta}_{ku}^\top = \mathbf{\Delta}_{uk}$. We will tacitly assume this in the following discussion.

GRAPH FEATURE INTERPOLATION is the problem of reconstructing the unknown features $\mathbf{x}_u$ given the graph structure $G$ and the known features $\mathbf{x}_k$. The interpolation task requires some prior on the behavior of the features of the graph, which can be expressed in the form of an energy function $\ell(\mathbf{x}, G)$. The most common assumption is feature *homophily* (i.e., that the features of every node are similar to those of the neighbours), quantified using a criterion of *smoothness* such as the Dirichlet energy. Since, in many cases, the behaviour of the features is not known, the energy can possibly be learned from the data.

LEARNING ON A GRAPH WITH MISSING FEATURES  is a transductive learning problem (typically node-wise classification or regression using some GNN architecture) where the structure of the graph $G$ is known while the labels and node features are only partially known on the subsets $V_l$ and $V_k$ of nodes, respectively (that might be different and even disjoint). Specifically, we try to learn a function $\mathbf{f}(\mathbf{x}_k, G)$ such that $f_i \approx y_i$ for $i \in V_l$. Learning with missing features can be done by a pre-processing step of graph signal interpolation (reconstructing an estimate $\tilde{\mathbf{x}}$ of the full feature vector $\mathbf{x}$ from $\mathbf{x}_k$) independent of the learning task, followed by the learning task of $\mathbf{f}(\tilde{\mathbf{x}}, G)$ on the inferred fully-featured graph. In some settings, we are not interested in recovering the features *per se*, but rather ensuring that the output of the *function* $\mathbf{f}$ on these features is correct – arguably a more 'forgiving' setting.

## 6.3 FEATURE PROPAGATION

We assume to be given $\mathbf{x}_k$ and attempt to find the missing node features $\mathbf{x}_u$ by means of interpolation that minimizes some energy $\ell(\mathbf{x}, G)$. In particular, we consider the *Dirichlet energy* $\ell(\mathbf{x}, G) = \frac{1}{2}\mathbf{x}^\top \mathbf{\Delta} \mathbf{x} = \frac{1}{2}\sum_{ij} \tilde{a}_{ij}(x_i - x_j)^2$, where $\tilde{a}_{ij}$ are the individual entries of the normalized adjacency $\tilde{\mathbf{A}}$. The Dirichlet energy is widely used as a smoothness criterion for functions defined on the nodes of the graph and thus promotes homophily. Functions minimizing the Dirichlet energy are called *harmonic*; without boundary conditions, it is minimized by a constant function.

While the Dirichlet energy is convex and it is possible to derive its minimizer in a closed-form, as shown in Sec. 6.A, its computational complexity makes it unfeasible for graphs with many nodes with missing features. Instead, we consider the associated *gradient flow* $\dot{\mathbf{x}}(t) = -\nabla \ell(\mathbf{x}(t))$ as a differential equation with boundary condition $\mathbf{x}_k(t) = \mathbf{x}_k$ whose solution at the missing nodes, $\mathbf{x}_u = \lim_{t \to \infty} \mathbf{x}_u(t)$, provides the desired interpolation.

GRADIENT FLOW.  For the Dirichlet energy, $\nabla_\mathbf{x} \ell = \mathbf{\Delta} \mathbf{x}$ and the gradient flow takes the form of the standard isotropic heat diffusion equation on the graph,

$$\dot{\mathbf{x}}(t) = -\mathbf{\Delta}\mathbf{x}(t) \qquad \text{(IC)} \ \mathbf{x}(0) = \begin{bmatrix} \mathbf{x}_k \\ \mathbf{x}_u(0) \end{bmatrix} \qquad \text{(BC)} \ \mathbf{x}_k(t) = \mathbf{x}_k$$

where IC and BC stand for initial conditions and boundary conditions respectively. By incorporating the boundary conditions, we can compactly express the diffusion equation as

$$\begin{bmatrix} \dot{\mathbf{x}}_k(t) \\ \dot{\mathbf{x}}_u(t) \end{bmatrix} = -\begin{bmatrix} \mathbf{0} & \mathbf{0} \\ \mathbf{\Delta}_{uk} & \mathbf{\Delta}_{uu} \end{bmatrix} \begin{bmatrix} \mathbf{x}_k \\ \mathbf{x}_u(t) \end{bmatrix} = -\begin{bmatrix} \mathbf{0} \\ \mathbf{\Delta}_{uk}\mathbf{x}_k + \mathbf{\Delta}_{uu}\mathbf{x}_u(t) \end{bmatrix}. \tag{6.1}$$

As expected, the gradient flow of the observed features is $\mathbf{0}$, given that they do not change during the diffusion.

The evolution of the missing features can be regarded as a heat diffusion equation with a constant heat source $\mathbf{\Delta}_{uk}\mathbf{x}_k$ coming from the boundary (known) nodes. Since the graph Laplacian matrix is positive semi-definite, the Dirichlet energy $\ell$ is convex. Its global minimizer is given by the solution to the closed-form equation $\nabla_{\mathbf{x}_u} \ell = \mathbf{0}$ and by rearranging the final $|V_u|$ rows of Eq. (6.1) we get

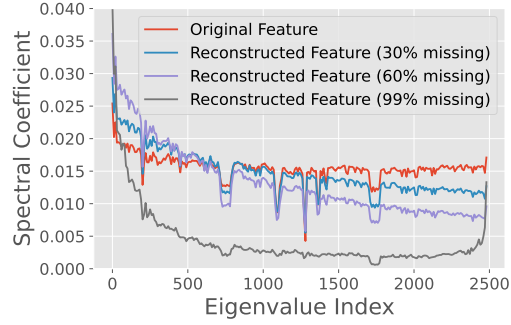

Figure 2: Graph Fourier transform magnitudes of the original Cora features (red) and those reconstructed by FP for varying rates of missing rates (we take the average over feature channels). Since FP minimizes the Dirichlet energy, it can be interpreted as a low-pass filter, which is stronger for a higher rate of missing features.

the solution $\mathbf{x}_u = -\mathbf{\Delta}_{uu}^{-1}\mathbf{\Delta}_{ku}^\top \mathbf{x}_k$. This solution always exists as $\mathbf{\Delta}_{uu}$ is non-singular, by virtue of the following:

**Proposition 6.3.1** (The sub-Laplacian matrix of an undirected connected graph is invertible)**.**
*Take any undirected, connected graph with adjacency matrix $\mathbf{A} \in \{0,1\}^{n \times n}$, and its Laplacian $\mathbf{\Delta} = \mathbf{I} - \mathbf{D}^{-1/2}\mathbf{A}\mathbf{D}^{-1/2}$, with $\mathbf{D}$ being the degree matrix of $\mathbf{A}$. Then, for any principle sub-matrix $\mathbf{L}_u \in \mathbb{R}^{b \times b}$ of the Laplacian, where $1 \leq b < n$, $\mathbf{L}_u$ is invertible.*

Proof: See Sec. 6.A. Also, while the proposition assumes that the graph is connected, our analysis and method generalize straightforwardly in the case of a disconnected graph as we can simply apply Feature Propagation to each connected component independently.

However, solving a system of linear equations is computationally expensive (incurring $\mathcal{O}(|V_u|^3)$ complexity for matrix inversion) and thus intractable for anything but only small graphs.

ITERATIVE SCHEME. As an alternative, we can discretize the diffusion equation( Eq. (6.1)) and solve it by an iterative numerical scheme. Approximating the temporal derivative as forward difference with the time variable $s$ discretized using a fixed step ($s = ht$ for step size $h > 0$ and $t = 1, 2, \ldots$), we obtain the *explicit Euler scheme:*

$$\mathbf{x}^{(t+1)} = \mathbf{x}^{(t)} - h \begin{bmatrix} \mathbf{0} & \mathbf{0} \\ \mathbf{\Delta}_{uk} & \mathbf{\Delta}_{uu} \end{bmatrix} \mathbf{x}^{(t)} = \left( \mathbf{I} - \begin{bmatrix} \mathbf{0} & \mathbf{0} \\ h\mathbf{\Delta}_{uk} & h\mathbf{\Delta}_{uu} \end{bmatrix} \right) \mathbf{x}^{(t)} = \begin{bmatrix} \mathbf{I} & \mathbf{0} \\ -h\mathbf{\Delta}_{uk} & \mathbf{I} - h\mathbf{\Delta}_{uu} \end{bmatrix} \mathbf{x}^{(t)}$$

For the special case of $h = 1$, we can use the following observation

$$\tilde{\mathbf{A}} = \mathbf{I} - \mathbf{\Delta} = \begin{bmatrix} \mathbf{I} & \mathbf{0} \\ \mathbf{0} & \mathbf{I} \end{bmatrix} - \begin{bmatrix} \mathbf{\Delta}_{kk} & \mathbf{\Delta}_{ku} \\ \mathbf{\Delta}_{uk} & \mathbf{\Delta}_{uu} \end{bmatrix} = \begin{bmatrix} \mathbf{I} - \mathbf{\Delta}_{kk} & -\mathbf{\Delta}_{ku} \\ -\mathbf{\Delta}_{uk} & \mathbf{I} - \mathbf{\Delta}_{uu} \end{bmatrix}, \tag{6.2}$$

to write the iteration formula as

$$\mathbf{x}^{(t+1)} = \begin{bmatrix} \mathbf{I} & \mathbf{0} \\ \tilde{\mathbf{A}}_{uk} & \tilde{\mathbf{A}}_{uu} \end{bmatrix} \mathbf{x}^{(t)}. \tag{6.3}$$

The Euler scheme is the gradient descent of the Dirichlet energy. Thus, applying the scheme decreases the Dirichlet energy and results in the features becoming increasingly smooth. The iteration in Eq. (6.3) can be interpreted as successive low-pass filtering. Fig. 2 depicts the magnitude of the graph Fourier coefficients of the original and reconstructed features on the Cora dataset, indicating that the higher the rate of missing features, the stronger the low-pass filtering effect.

The following results show that the iterative scheme with $h = 1$ always converges and its steady state is equal to the closed-form solution. Importantly, the solution does not depend on the initial values $\mathbf{x}_u^{(0)}$ given to the unknown features.

**Proposition 6.3.2.** *Take any undirected and connected graph with adjacency matrix* $\mathbf{A} \in \{0,1\}^{n \times n}$, *and normalised Adjacency* $\tilde{\mathbf{A}} = \mathbf{D}^{-1/2}\mathbf{A}\mathbf{D}^{-1/2}$, *with* $\mathbf{D}$ *being the degree matrix of* $\mathbf{A}$. *Let* $\mathbf{x} = \mathbf{x}^{(0)} \in \mathbf{R}^n$ *be the initial feature vector and define the following recursive relation*

$$\mathbf{x}^{(t)} = \begin{bmatrix} \mathbf{I} & \mathbf{0} \\ \tilde{\mathbf{A}}_{uk} & \tilde{\mathbf{A}}_{uu} \end{bmatrix} \mathbf{x}^{(t-1)}.$$

*Then this recursion converges, and the steady state is given to be*

$$\lim_{t \to \infty} \mathbf{x}^{(t)} = \begin{bmatrix} \mathbf{x}_k \\ -\mathbf{\Delta}_{kk}^{-1}\tilde{\mathbf{A}}_{uk}\mathbf{x}_k \end{bmatrix}.$$

Proof: See Sec. 6.B.

FEATURE PROPAGATION ALGORITHM. We can notice that the update in Eq. (6.3) is equivalent to first multiplying the feature vector $\mathbf{x}$ by the original diffusion matrix $\tilde{\mathbf{A}}$, and then resetting the known features to their true value. This gives us Algorithm 2, an extremely simple and scalable iterative algorithm to reconstruct the missing features on a graph, which we refer to as *Feature Propagation* (FP). While $\mathbf{x}_u$ can be initialized to any value, in practice we initialize $\mathbf{x}_u$ to zero and find 40 iterations to be enough to provide convergence for all datasets we experimented on. At each iteration, the diffusion occurs from the nodes with known features to the nodes with unknown features as well as among the nodes with unknown features.

**Algorithm 2** Feature Propagation

1: **Input:** feature vector $\mathbf{x}$, diffusion matrix $\tilde{\mathbf{A}}$
2: $\mathbf{y} \leftarrow \mathbf{x}$
3: **while** $\mathbf{x}$ has not converged **do**
4: $\quad \mathbf{x} \leftarrow \tilde{\mathbf{A}}\mathbf{x}$ ▷ Propagate features
5: $\quad \mathbf{x}_k \leftarrow \mathbf{y}_k$ ▷ Reset known features
6: **end while**

EXTENSION TO VECTOR-VALUED FEATURES. Algorithm 2 extends seamlessly to vector-valued features by simply replacing the feature vector $\mathbf{x}$ with a $n \times d$ feature matrix $\mathbf{X}$, where $d$ is the number of features. Multiplying the diffusion matrix $\tilde{\mathbf{A}}$ by the feature matrix $\mathbf{X}$ diffuses each feature channel independently. Interestingly, it would not be trivial to extend Eq. (6.3) to vector-valued features without noticing its equivalence with Algorithm 2, as each node could have different missing features, leading to different sub-matrices $\tilde{\mathbf{A}}_{uk}$ and $\tilde{\mathbf{A}}_{uu}$ for each feature channel.

LEARNING. One significant advantage of FP is that it can be easily combined with any graph learning model to generate predictions for the downstream task. Moreover, FP is not aimed at merely reconstructing the node features. Instead, by only reconstructing the lower frequency components of the signal, it is by design very well suited to be combined with GNNs, which are known to mainly leverage these lower frequency components [258]. Our approach is generic and can be used for any graph-related task for missing features, such as node classification, link prediction and graph classification. In this paper, we focus on node classification.

OVERSMOOTHING. Fig. 2 shows that the more features are missing, the smoother the reconstruction produced by FP is. Despite this, FP does not suffer from oversmoothing [185], a term used when node representations converge to similar values. Oversmoothing is caused by repeated diffusion and occurs widely when stacking more than a few layers of the most popular GNNs such as GCN [136], GAT [244] or SGC [258]. However, the boundary conditions in the Feature Propagation diffusion equation prevent the reconstructed features from becoming overly smooth, even when using an extremely high number of diffusion steps. This has also been studied by CGNN [260] and GRAND++ [233], which require soft boundary conditions in the form of a source term to prevent oversmoothing, although not in the context of missing features.

## 6.4 RELATED WORK

LABEL PROPAGATION. The proposed algorithm bears some similarity with Label Propagation [295] (LP), which predicts a class for each node by propagating the known labels in the graph. Differently from our setting of diffusion of continuous node features, they deal with discrete label classes directly, resulting in a different diffusion operator. However, the key difference between them lies in how they are used. Importantly, LP is used to directly perform node classification, taking into account only the graph structure and being unaware of node features. On the other hand, FP is used to reconstruct missing features, which are then fed into a downstream GNN classifier. FP allows a GNN model to effectively combine features and graph structures, even when most of the features are missing. Our experiments show that FP+GNN always outperforms LP, even in cases of extremely high rates of missing features, suggesting the effectiveness of FP. Also, the derived scheme is a special case of Neural Graph PDEs [42], which are in turn related to the iterative scheme presented in [290].

MATRIX COMPLETION. Several optimization-based approaches [38, 114] as well as learning-based approaches [133, 155, 270] have been proposed to solve the matrix completion problem. However, they are unaware of the underlying graph structure. Graph matrix completion [23, 127, 172, 198] extends the above approaches to make use of an underlying graph. Similarly, Graph Signal Processing offers several methods to interpolate signals on graphs. [178] prove the necessary conditions for a graph signal to be recovered perfectly, and provide a corresponding algorithm. However, due to the optimisation problems involved, most of the above approaches are too computationally intensive and cannot scale to graphs with more than ∼1,000 nodes. Moreover, the goal of all the above approaches is to reconstruct the missing entries of the matrix rather than solving a downstream task.

EXTENDING GNNS TO MISSING NODE FEATURES. SAT [48] consists of a Transformer-like model for feature reconstruction and a GNN model to solve the downstream task. GCNMF [230] adapts GCN [136] to the case of missing node features by representing the missing data with a Gaussian mixture model. PaGNN [123] is a GCN-like model which uses a partial message-passing scheme to only propagate observed features. While showing a reasonable performance for low rates of missing features, these methods suffer in regimes of high rates of missing features and do not scale to large graphs.

OTHER RELATED GNN WORKS. Several papers investigate how to augment GNNs when no node features are available [59], as well as investigating the performance of GNNs with random features [1, 209]. Dirichlet energy minimization has been widely used as a regularizer in several graph-related tasks [255, 290, 296]. Discretization of continuous diffusion on graphs has already been explored in [42] and [260]. Propagation on the graph has also been studied as a solution to the different problem of node regression on multi-relational graphs [20]. Other methods have investigated propagating node features [47, 83, 258]; however, not in the scenario of missing features. The boundary conditions given by the available features in FP's diffusion equation (enforced by resetting the known feature after each iteration in the algorithm) make it different from other propagation approaches and an effective solution to the missing features problem. While [47, 83, 258] assume to observe all features, and then modify all features, FP assumes to observe only a subset of the features and modifies only the unobserved ones.

## 6.5 EXPERIMENTS AND DISCUSSION

DATASETS. We evaluate on the task of node classification on several benchmark datasets: Cora, Citeseer and PubMed [212], Amazon-Computers, Amazon-Photo [215] and OGBN-Arxiv [113]. To test the scalability of our method, we also test it on OGBN-Products (2,449,029 nodes, 123,718,280 edges). We report dataset statistics in Tab. 6.B.1.

BASELINES. We compare to two strong feature-agnostic baselines: Label Propagation [295], which only makes use of the graph structure by propagating labels on the graph, and Graph Positional Encodings [69], which consist in computing the top $k$ eigenvectors of the Laplacian matrix and treating them as node features in input to a GNN. We additionally compare to feature-imputation methods that are graph-agnostic, such as setting the missing features to 0 (Zero), a random value from a standard Gaussian (Random), or the global mean of that feature over the graph (Global Mean) [1]. We also compare to a simple graph-based imputation baseline, which sets a missing feature to the mean (of that same feature) over the neighbors of a node (Neighbor Mean). We additionally experiment with MGCNN [172], a geometric graph completion method which learns how to reconstruct the missing features by making use of the observed features and the graph structure. For all the above baselines, as well as for our Feature Propagation, we experiment with both GCN [136] and GraphSage with mean aggregator [102] as downstream GNNs. We also compare to recently state-of-the-art methods for learning in the missing features setting (GCNMF [230] and PaGNN [123]). For GCNMF we use the publicly available code.[2] We could not find publicly available code for PaGNN so use our own implementation for this comparison. We do not compare to other commonly used imputation based methods such as VAE [133] or GAIN [270], nor to the Transformer-based method SAT [48], as they have previously been shown to consistently underperform GCNMF and PaGNN [123, 230].

EXPERIMENTAL SETUP. We report the mean and standard error of the test accuracy, computed over 10 runs, in all experiments. Each run has a different train/validation/test split (apart from OGBN datasets where we use the provided splits) and mask of missing features[3]. The splits are generated at random by assigning 20 nodes per class to the training set, 1500 nodes in total to the validation set and the rest to the test set, similar to [138]. For a fair comparison, we use the same standard hyperparameters for all methods across all experiments. We train using the Adam [132] optimizer with a learning rate of 0.005 for a maximum of 10000 epochs, combined with early stopping with a patience of 200. Downstream GNN models (as well as GCNMF and PaGNN) use 2 layers with a hidden dimension of 64 and a dropout rate of 0.5 for all datasets, apart from OGBN datasets where 3 layers and a hidden dimension of 256 are used. For OGBN-Arxiv we also employ the Jumping Knowledge scheme [266] with max aggregation. Feature Propagation uses 40 iterations to diffuse the features, as we found this to be enough to reach convergence on all datasets. We want to emphasize that we did not perform any hyperparameter tuning, and FP proved to perform consistently with any reasonable choice of hyperparameters. We use neighbor sampling [102] when training on OGBN-Products. All experiments are conducted on an AWS p3.16xlarge machine with 8 NVIDIA V100 GPUs with 16GB of memory each, and took around 4 GPU days in total to perform.

---

[1] If a feature is not observed for any of the node's neighbors, we set it to zero.
[2] https://github.com/marblet/GCNmf
[3] Each entry of the feature matrix is independently missing with a probability equal to the missing rate.

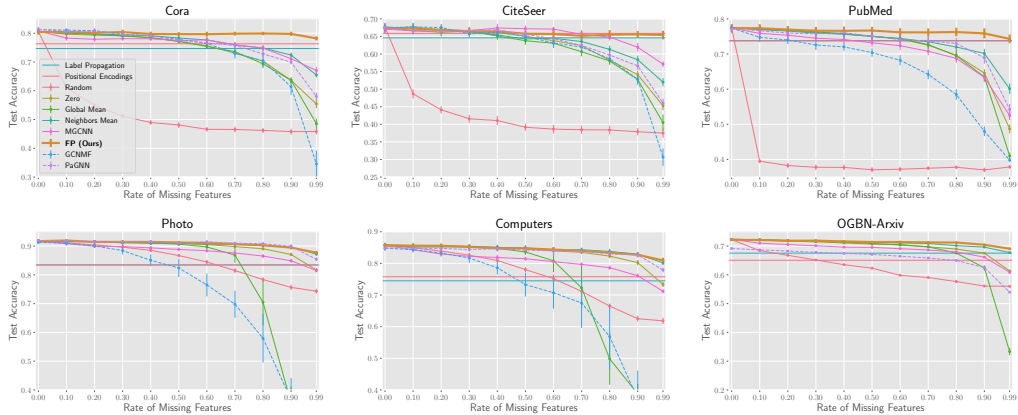

Figure 3: Test accuracy for a varying rate of missing features on six common node-classification benchmarks. For methods that require a downstream GNN, a 2-layer GCN [136] is used. On OGBN-Arxiv, GCNMF goes out-of-memory and is not reported.

NODE CLASSIFICATION RESULTS. Fig. 3 shows the results for different rates of missing features (x-axis), when using GCN as a downstream GNN (results with GraphSAGE are reported in Fig. 6.B.1). FP matches or outperforms other methods in all scenarios. Both GCNMF and PaGNN are consistently outperformed by the simple Neighbor Mean baseline. This is not completely unexpected, as Neighbor Mean can be seen as a first-order approximation of Feature Propagation, where only one step of propagation is performed (and with a slightly different normalization of the diffusion operator). We elaborate on the relation between Neighbor Mean and Feature Propagation as well as on the results of the other baselines in Sec. 6.D. Interestingly, most methods perform extremely well up to 50% of missing features, suggesting that, in general, node features are redundant, as replacing half of them with zeros (*Zero* baseline) has little effect on the performance. The gap between methods opens up from around 60% of missing features, and is particularly large for extremely high rates of missing features (90% or 99%): FP is the only feature-aware method which is robust to these high rates on all datasets (see Tab. 2). Moreover, FP outperforms or matches Label Propagation and Positional Encodings on all datasets, even in the extreme case of 99% missing features. On some datasets, such as Cora, Photo, and Computers, the gap is especially significant. We conclude that reconstructing the missing features using FP is indeed useful for the downstream task. We highlight the surprising results that, on average, FP with 99% missing features performs only 4.12% worse (in relative accuracy terms) than the same GNN model used with no missing features, compared to 58.33% and 21.25% worse for GCNMF and PaGNN respectively.

| Dataset | Full Features | 50.0% Missing | 90.0% Missing | 99.0% Missing |
| --- | --- | --- | --- | --- |
| Cora | 80.39% | 79.70%(-0.86%) | 79.77%(-0.77%) | 78.22%(-2.70%) |
| CiteSeer | 67.48% | 65.74%(-2.57%) | 65.57%(-2.82%) | 65.40%(-3.08%) |
| PubMed | 77.36% | 76.68%(-0.89%) | 75.85%(-1.96%) | 74.29%(-3.97%) |
| Photo | 91.73% | 91.29%(-0.48%) | 89.48%(-2.46%) | 87.73%(-4.36%) |
| Computers | 85.65% | 84.77%(-1.04%) | 82.71%(-3.43%) | 80.94%(-5.51%) |
| OGBN-Arxiv | 72.22% | 71.42%(-1.10%) | 70.47%(-2.43%) | 69.09%(-4.33%) |
| OGBN-Products | 78.70% | 77.16%(-1.96%) | 75.94%(-3.51%) | 74.94%(-4.78%) |
| Average | 79.08% | 78.11%(-1.27%) | 77.11%(-2.48%) | 75.80%(-4.10%) |

Table 1: Performance of Feature Propagation (combined with a GCN model) for 50%, 90% and 99% of missing features, and relative drop compared to the performance of the same model when all features are present. On average, our method loses only 2.50% of relative accuracy with 90% of missing features, and 4.12% with 99% of missing features.

| Dataset | GCNMF | PaGNN | Label Prop. | Pos. Enc. | FP (Ours) |
| --- | --- | --- | --- | --- | --- |
| Cora | 34.54±2.07 | 58.03±0.57 | 74.68±0.36 | 76.33±0.26 | **78.22±0.32** |
| CiteSeer | 30.65±1.12 | 46.02±0.58 | 64.60±0.40 | **65.87±0.37** | 65.40±0.54 |
| PubMed | 39.80±0.25 | 54.25±0.70 | 73.81±0.56 | 73.70±0.29 | **74.29±0.55** |
| Photo | 29.64±2.78 | 85.41±0.28 | 83.45±0.94 | 83.45±0.26 | **87.73±0.27** |
| Computers | 30.74±1.95 | 77.91±0.33 | 74.48±0.61 | 75.77±0.47 | **80.94±0.37** |
| OGBN-Arxiv | OOM | 53.98±0.08 | 67.56±0.00 | 65.08±0.04 | **69.09±0.06** |
| OGBN-Products | OOM | OOM | 74.42±0.00 | OOM | **74.94±0.07** |

Table 2: Performance of GCNMF, PaGNN and FP(+GCN) with 99% of features missing, as well as Label Propagation and Positional Encodings (which are feature-agnostic). GCNMF and PaGNN perform respectively 58.33% and 21.25% worse in terms of relative accuracy in this scenario compared to when all the features are present. In comparison, FP has only a 4.12% drop.

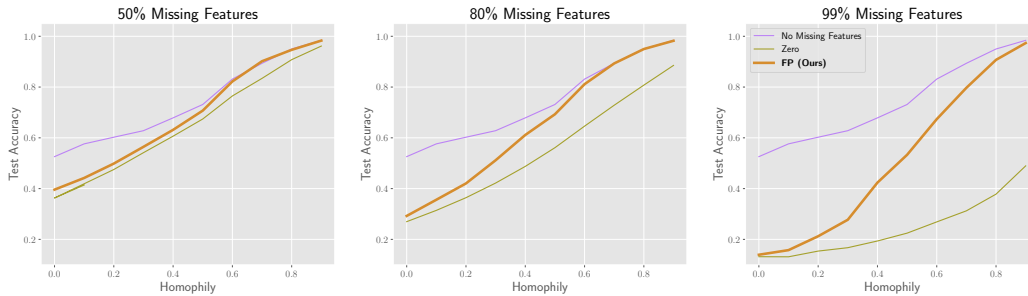

Figure 5: Test accuracy on the synthetic datasets from [3] with different levels of homophily. We use GraphSage as downstream model as it is preferable to GCN on low homophily data [294].

RUN-TIME ANALYSIS. Feature Propagation scales to extremely large graphs, as it only consists of repeated sparse-to-dense matrix multiplications. Moreover, it can be regarded as a pre-processing step, and performed only once, separately from training. In Fig. 4 we compare the run-time to complete the training of the model for FP, PaGNN and GCNMF. The time for FP includes both the feature propa-gation step to reconstruct the missing features and training of a downstream GCN model. FP is around 3x faster than PaGNN and GCNMF. The propagation step of FP takes only a fraction of the total running time, and the vast majority of the time is spent in training the downstream model. The feature propagation step takes only ∼0.6s for Computers, ∼0.8s for OGBN-Arxiv and ∼10.5s for OGBN-Products using a single GPU. Both PaGNN and GCNMF go out-of-memory on OGBN-Products.

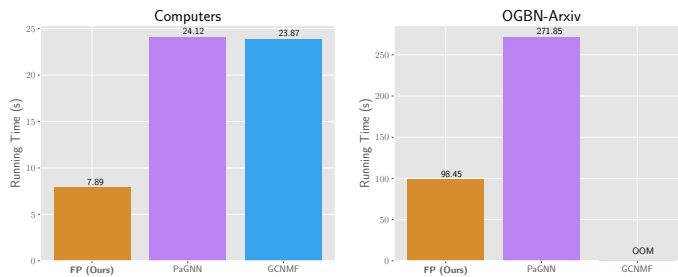

Figure 4: Run-time (in seconds) of FP, PaGNN and GCNMF. FP is 3x faster than both other methods. GCNMF goes out-of-memory (OOM) on OGBN-Arxiv.

WHEN DOES FEATURE PROPAGATION WORK? Since FP can be interpreted as a low-pass filter that smoothes the features on the graph, we expect it to be suitable in the case of homophilic graph data (where neighbors tend to have similar attributes), and, conversely, to suffer in scenarios of low homophily. To verify this, we experiment on the synthetic dataset from [3], which consists of 10 graphs with different levels of homophily. Fig. 5 confirms our hypothesis: when the homophily is high, Feature Propagation with 99% of features missing performs similarly to the case when all the features are known. As the homophily decreases, the gap between the two widens to become extremely large in the case of zero homophily. In such scenarios, FP is only slightly better than setting the missing features to zero (Zero baseline). This observation calls for a different kind of non-homogeneous diffusion dependent on the features that can potentially be made learnable for low-homophily data. We leave this as future work.

## 6.6 Conclusion

We have introduced a novel approach for handling missing node features in graph-learning tasks. Our Feature Propagation model can be directly derived from energy minimization, and can be implemented as an efficient iterative algorithm where the features are multiplied by a diffusion matrix, before resetting the known features to their original value. Experiments on a number of datasets suggest that FP can reconstruct the missing features in a way that is useful for the downstream task, even when 99% of the features are missing. FP outperforms previously proposed methods by a significant margin on common benchmarks, while also being extremely scalable.

LIMITATIONS.   While our method is designed for homophilic graphs, a more general learnable diffusion could be adapted to perform well in low homophily scenarios, as discussed in Sec. 6.5. Feature Propagation is designed for graphs with only one node and edge type, however it could be extended to heterogenous graphs by having separate diffusions for different types of edges and nodes. Finally, Feature Propagation treats feature channels independently. To account for dependencies, diffusion with channel mixing should be used.

FAST FORWARD TO TODAY.   Since the initial introduction of Feature Propagation in our earlier work [203], the methodology has undergone various enhancements and adaptations. These include extensions to spatio-temporal graphs [165, 251], heterogeneous graphs [146], and multi-level attributes [281]. Notably, the work by Um et al. [240] employs a feature correlation matrix to facilitate channel mixing.

In addition to theoretical advancements, Feature Propagation has been practically applied across a range of domains. For instance, it has been leveraged to obtain embeddings for new users within recommender systems [37], as well as for the imputation of single-cell RNA-Seq data [279].

Feature Propagation techniques are also gaining traction in addressing the broader issue of imputing missing data in tabular formats. In such scenarios, artificial graphs are often constructed based on feature similarity between individual samples [225], or they may be generated through learned processes [232]. As an alternative approach, You et al. [272] propose a bipartite graph model that represents both samples and features as nodes, thereby obviating the need for generating or learning an artificial graph between samples.

# Appendices

## 6.A Closed-Form Solution for Harmonic Interpolation

Given the *Dirichlet energy* $\ell(\mathbf{x}, G) = \frac{1}{2}\mathbf{x}^\top \mathbf{\Delta} \mathbf{x}$, we want to solve for missing features $\mathbf{x}_u = argmin_{\mathbf{x}_u} \ell$, leading to the optimality condition $\nabla_{\mathbf{x_u}} \ell = \mathbf{0}$. From Eq. (6.1) we find $\nabla_{\mathbf{x_u}} \ell = \mathbf{0}$ to be the solution of $\mathbf{\Delta}_{uk}\mathbf{x}_k + \mathbf{\Delta}_{uu}\mathbf{x}_u = \mathbf{0}$. The unique solution to this system of linear equations is $\mathbf{x}_u = -\mathbf{\Delta}_{uu}^{-1}\mathbf{\Delta}_{uk}\mathbf{x}_k$. We show this solution always exists by proving $\mathbf{\Delta}_{uu}$ is non-singular ( Proposition 6.3.1). The proof of this result follows from the following Lemma.

**Lemma 6.A.1.** *Take any undirected and connected graph with adjacency matrix $\mathbf{A} \in \{0, 1\}^{n \times n}$, and normalised Adjacency $\tilde{\mathbf{A}} = \mathbf{D}^{-1/2}\mathbf{A}\mathbf{D}^{-1/2}$, with $\mathbf{D}$ being the degree matrix of A. Let $\tilde{\mathbf{A}}_{uu}$ be the bottom right submatrix of $\tilde{\mathbf{A}}$ where $1 \leqslant b < n$. Then $\rho(\tilde{\mathbf{A}}_{uu}) < 1$ where $\rho(\cdot)$ denotes spectral radius.*

*Proof.* Define
$$\tilde{\mathbf{A}}_{up} = \begin{bmatrix} \mathbf{0}_u & \mathbf{0}_{uk} \\ \mathbf{0}_{ku} & \tilde{\mathbf{A}}_{uu} \end{bmatrix},$$
to be the matrix equal to $\tilde{\mathbf{A}}_{uu}$ in the lower right $b \times b$ sub-matrix and padded with zero entries elsewhere. Clearly $\tilde{\mathbf{A}}_{up} \leqslant \tilde{\mathbf{A}}$ elementwise and $\tilde{\mathbf{A}}_{up} \neq \tilde{\mathbf{A}}$. Furthermore, $\tilde{\mathbf{A}}_{up} + \tilde{\mathbf{A}}$ represents an adjacency matrix of some strongly connected graph and is therefore irreducible [25, Theorem 2.2.7]. These observations allow us to deduce that $\rho(\tilde{\mathbf{A}}_{up}) < \rho(\tilde{\mathbf{A}})$ [25, Corollary 2.1.5]. Note that $\rho(\tilde{\mathbf{A}}_{up}) = \rho(\tilde{\mathbf{A}}_{uu})$ as $\tilde{\mathbf{A}}_{up}$ and $\tilde{\mathbf{A}}_{uu}$ share the same non-zero eigenvalues. Furthermore, $\rho(\tilde{\mathbf{A}}) \leqslant 1$ as we can write $\tilde{\mathbf{A}} = \mathbf{I} - \mathbf{\Delta}$ and $\mathbf{\Delta}$ is known to have eigenvalues in the range $[0, 2]$ [53]. Combining these inequalities gives the result $\rho(\tilde{\mathbf{A}}_{uu}) = \rho(\tilde{\mathbf{A}}_{up}) < \rho(\tilde{\mathbf{A}}) \leqslant 1$. □

**Proposition 6.A.2** (The sub-Laplacian matrix of a undirected connected graph is invertible)**.** *Take any undirected, connected graph with adjacency matrix $\mathbf{A} \in \{0, 1\}^{n \times n}$, and its Laplacian $\mathbf{\Delta} = \mathbf{I} - \mathbf{D}^{-1/2}\mathbf{A}\mathbf{D}^{-1/2}$, with $\mathbf{D}$ being the degree matrix of $\mathbf{A}$. Then, for any principle sub-matrix $\mathbf{L}_u \in \mathbb{R}^{b \times b}$ of the Laplacian, where $1 \leqslant b < n$, $L_u$ is invertible.*

*Proof.* To prove $\mathbf{\Delta}_{uu}$ is non-singular it is enough to show 0 is not an eigenvalue. Note that $\mathbf{\Delta}_{uu} = \mathbf{I} - \tilde{\mathbf{A}}_{uu}$ so 0 is not an eigenvalue if and only if $\tilde{\mathbf{A}}_{uu}$ does not have an eigenvalue equal to 1, which follows from Lemma 6.A.1. □

## 6.B Closed-Form Solution for the Euler scheme

**Proposition 6.B.1.** *Take any undirected and connected graph with adjacency matrix $\mathbf{A} \in \{0, 1\}^{n \times n}$, and normalised Adjacency $\tilde{\mathbf{A}} = \mathbf{D}^{-1/2}\mathbf{A}\mathbf{D}^{-1/2}$, with $\mathbf{D}$ being the degree matrix of $\mathbf{A}$. Let $\mathbf{x} = \mathbf{x}^{(0)} \in \mathbf{R}^n$ be the initial feature vector and define the following recursive relation*

$$\mathbf{x}^{(t)} = \begin{bmatrix} \mathbf{I} & \mathbf{0} \\ \tilde{\mathbf{A}}_{uk} & \tilde{\mathbf{A}}_{uu} \end{bmatrix} \mathbf{x}^{(t-1)}.$$

| Dataset | Nodes | Edges | Features | Classes |
|---|---|---|---|---|
| Cora | 2,485 | 5,069 | 1,433 | 7 |
| CiteSeer | 2,120 | 3,679 | 3,703 | 6 |
| PubMed | 19,717 | 44,324 | 500 | 3 |
| Photo | 7,487 | 119,043 | 745 | 8 |
| Computers | 13,381 | 245,778 | 767 | 10 |
| OGBN-Arxiv | 169,343 | 1,166,243 | 128 | 40 |
| OGBN-Products | 2,449,029 | 123,718,280 | 100 | 47 |

Table 6.B.1: Dataset statistics.

*Then this recursion converges and the steady state is given to be*

$$\lim_{t\to\infty} \mathbf{x}^{(t)} = \begin{bmatrix} \mathbf{x}_k \\ -\boldsymbol{\Delta}_{kk}^{-1}\tilde{\mathbf{A}}_{uk}\mathbf{x}_k \end{bmatrix}.$$

*Proof.* The recursive relation can be written in the following form

$$\begin{bmatrix} \mathbf{x}_k^{(t)} \\ \mathbf{x}_u^{(t)} \end{bmatrix} = \begin{bmatrix} \mathbf{I}_l & \mathbf{0}_{ku} \\ \tilde{\mathbf{A}}_{uk} & \tilde{\mathbf{A}}_{uu} \end{bmatrix} \begin{bmatrix} \mathbf{x}_k^{(t-1)} \\ \mathbf{x}_u^{(t-1)} \end{bmatrix} = \begin{bmatrix} \mathbf{x}_k^{(t-1)} \\ \tilde{\mathbf{A}}_{uk}\mathbf{x}_k^{(t-1)} + \tilde{\mathbf{A}}_{uu}\mathbf{x}_u^{(t-1)} \end{bmatrix}.$$

The first $l$ rows remain the same so we can write $\mathbf{x}_k^{(t)} = \mathbf{x}_k^{(t-1)} = \mathbf{x}_k$ and consider just the convergence of the last $u$ rows

$$\mathbf{x}_u^{(t-1)} = \tilde{\mathbf{A}}_{uk}\mathbf{x}_k + \tilde{\mathbf{A}}_{uu}\mathbf{x}_u^{(t-1)}.$$

We can look at the stationary behaviour by unrolling this recursion and taking the limit to find stationary state

$$\lim_{t\to\infty}\mathbf{x}_u^{(t)} = \lim_{t\to\infty}\tilde{\mathbf{A}}_{uu}^t \mathbf{x}_u^{(0)} + \left(\sum_{i=1}^t \tilde{\mathbf{A}}_{uu}^{(i-1)}\right)\tilde{\mathbf{A}}_{uk}\mathbf{x}_k.$$

Using Lemma 6.A.1 we find $\lim_{t\to\infty}\tilde{\mathbf{A}}_{uu}^t\mathbf{x}_u^{(0)} = \mathbf{0}$ and the geometric series converges giving us the following limit

$$\lim_{t\to\infty}\mathbf{x}_u^{(t)} = \left(\mathbf{I}_u - \tilde{\mathbf{A}}_{uu}\right)^{-1}\tilde{\mathbf{A}}_{uk}\mathbf{x}_k = \boldsymbol{\Delta}_{uu}^{-1}\tilde{\mathbf{A}}_{uk}\mathbf{x}_k = -\boldsymbol{\Delta}_{uu}^{-1}\boldsymbol{\Delta}_{uk}\mathbf{x}_k$$

where in the last step we make use of Eq. (6.2). □

## 6.C Baselines' Implementation and Tuning

**Label Propagation**  We use the label propagation implementation provided in Pytorch-Geometric [72]. Since the method is quite sensitive to the value of the $\alpha$ hyperparameter, we perform a gridsearch separately on each dataset over the following values: $[0.1, 0.2, 0.3, 0.4, 0.5, 0.6, 0.7, 0.8, 0.9, 0.95, 0.99]$.

**Positional Encodings**  We compute the laplacian eigenvectors using SciPy [247] sparse eigenvectors routines. We use the top twenty eigenvectors as positional encodings.

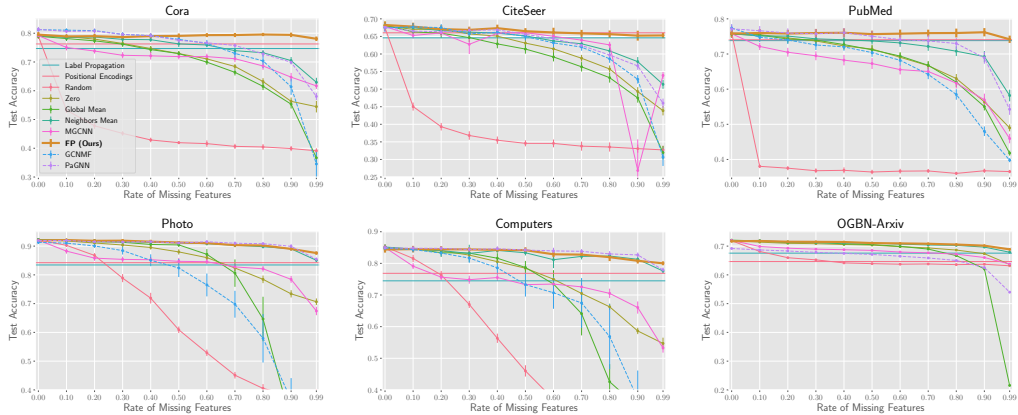

Figure 6.B.1: Test accuracy for varying rate of missing features on six common node-classification benchmarks. For methods that require a downstream GNNs, a 2-layer GraphSAGE [102] is used. On OGBN-Arxiv, GCNMF goes out-of-memory and is not reported.

MGCNN   We re-implement MGCNN [172] in Pytorch by taking inspiration from the authors' public TensorFlow code [4]. For simplicity, we use the version of the model with only graph convolutional layers and without an LSTM. For the matrix completion training process, we split the observed features into 50% input data, 40% training targets and 10% validation data. Once the MGCNN model is trained, we feed it the matrix with all the observed features to predict the whole feature matrix. This reconstructed features matrix is then used as input for a downstream GNN (as for the feature-imputation baselines).

## 6.D  Discussion Over Baselines' Performance

Neighborhood Averaging   As for some intuition as to why the simple Neighborhood Averaging performs competitively, let us assume to have a single feature channel and this feature to be homophilous over the graph. When a node has enough neighbors, the average of their features is a good estimate for the feature of the given node. However, as the rate of missing features increases, the feature may be present for only a few neighbors (or none at all), causing the estimate to have a higher variance. On the other hand, Feature Propagation allows information to travel longer distances in the graph by repeatedly multiplying by the diffusion matrix. Even if we do not observe the feature for any of a node's neighbors, it is still possible to estimate it from nodes further away in the graph. This can be observed empirically: the gap between Neighborhood Averaging and Feature Propagation becomes increasingly significant for higher rates of missing features.

Zero vs Random   In models such as GCN and GraphSage, where node embeddings are computed as the (weighted) average of neighbors embeddings, the effect of the Zero baseline is simply to reduce the norm of the average embeddings of all nodes (since all nodes have the same expected proportion of neighbors with missing features). On the other hand, the Random baseline corrupts this weighted average. More generally, while for a GNN model it could be relatively easy to learn to ignore features set to zero, and only focus on known (non-zero) features, it would be basically impossible for the model to do the same when setting the missing features to a random value.

---
[4]https://github.com/fmonti/mgcnn

However, we find Random to perform better than Zero when all features are missing. This is in line with findings in the literature [1, 209], where Random features have been shown to work well in conjunction with GNNs as they act as signatures for the nodes. On the other hand, if all nodes have zero vectors, it becomes basically impossible to distinguish them. After applying a GNN, all nodes will still have very similar embeddings, and the task performance will be close to a random guess.

# 7 Game-Theoretic Structural Inference

> "We cannot change the cards we are dealt, just how we play the game." – Randy Pausch

## 7.1 Introduction

Individuals or organisations cooperate with or compete against each other in a wide range of practical situations. Strategic interactions between individuals are often modelled as games played on networks [121], where an individual's utility depends not only on their actions (and characteristics) but also on those of their neighbours. In a network game (and games in general), the utility function is of fundamental importance, as rational agents maximise their utilities when making real-life decisions. The current literature on network games has primarily focused on the scenarios where the utility function is predefined, and the structure of the network (represented by a graph) is known beforehand. However, in practical settings, while it is common to observe the actions of the players, the underlying interaction network often remains hidden or uncertain due to privacy reasons or the dynamic nature of interactions. Furthermore, the utility function is often unobservable. This makes it challenging to exploit network information and utility functions for behavioural predictions and network-based interventions [241], e.g., marketing campaigns or information diffusion.

In this chapter, we focus on the problem of inferring the structure of the interaction network from the observed equilibrium actions of a network game. A few recent studies have tackled similar problems [14, 15, 80, 81, 82, 86, 87, 109, 120, 145]; however, several major limitations remain. First, all these methods require explicit knowledge of the utility function to infer the underlying network structure, which may be impractical to assume and may also change over time. The work of [109] considers a more general hypothesis space of games for linear influence games, but they only focus on binary actions and linear payoffs. Second, the methodology in each of these studies has been designed for the specific game under consideration, thereby limiting its scope in handling a wide range of strategic interactions in real-world scenarios.

CONTRIBUTIONS. First, we propose a unified parameterisation to summarise three common network games, which helps reveal the different nature of these games and interpret the strategic interactions they represent. This motivates us to propose a transformer-like data-driven model where we learn the functional mapping between the equilibrium actions and network structure. To our knowledge, our framework is one of the first that is able to infer the network structure behind the games *without* explicit knowledge about the utility function of the game. This capability is important in real-world scenarios where the nature of the interactions remains hidden or may even evolve over time. Synthetic and real-world experiments demonstrate the superiority of our method against several state-of-the-art approaches.

## 7.2 Related Work

*Network games*, a class of problems in game theory, have been studied extensively in computer science and economics. The majority of works in the network game literature study the characteristics of games on a known and static graph [9, 33]. While these studies are useful in understanding collective actions and designing interventions [78], it is increasingly acknowledged that networks are difficult to obtain in practice. Furthermore, the utility function associated with the game is usually unknown as well. We are interested in the *inverse problem* of inferring the network structure based on observed actions. This inverse setting is related to *graph* or *network inference*, a problem that has attracted interest in statistics [75, 139], physics [91, 92], signal processing [65, 167]. Our study differs from these works in accounting for the strategic interactions and the game theoretical framework underlying the observed data.

Deep learning models have been recently proposed for latent graph inference in a number of settings. Kipf et al. [134] proposes a graph neural network (GNN) model to infer the interactions of an underlying dynamical system from observational data. Differently from this work, their model is trained on predictions of the future state of the system, where there is a lack of validation for the learned network interactions. Similarly, in latent graph learning [57, 58, 253], the graph is learned jointly with a downstream task, conversely to our scenario where the structure itself is the learning objective. Methods for link prediction [283] predict edges in a graph, but they typically require part of the true links to be provided as input (and only predict the missing ones), whereas in our scenario, we are interested in inferring networks without observing any link in the test data. Among the few works where the network structure is the learning objective, Yu et al. [278] and Zheng et al. [289] propose to infer a DAG from the observed actions. However, these approaches are limited to predicting acyclic graphs, whereas the graphs we are interested in are often cyclic. The method of Belilovsky et al. [21] is the most related to ours, since they propose a supervised model to infer an undirected graphical model from observed covariates using a series of dilated convolutions. However, their model is not permutation-equivariant w.r.t. the order of nodes, and the number of layers depends on the number of nodes. Both issues cause the statistical efficiency of the model to scale poorly with the size of the graph. Different from the above studies, our framework aims to learn game-theoretical relationships in a supervised manner, while leveraging the structural symmetries of this problem.

Finally, there has been a recent stream of literature in learning network games from actions of players [14, 15, 80, 81, 82, 86, 87, 109, 120, 145]. Most of these methods focus on either a binary or a finite discrete action space. For continuous actions, Leng et al. [145] formulate an optimisation problem to learn the structure and marginal benefits of linear quadratic games, while Barik et al. [14] aim at inferring the network structure from an action-conforming graphical game. Garg et al. [82] learn a mapping from observed actions to interaction structures where individuals repeatedly update their strategies based on a weighted aggregate of other players' choices. Our work differs from existing methods in the literature in that it does not assume a specific game-theoretic structure (e.g., utility functions). Instead, we build a transformer-like model that learns a mapping from the equilibrium actions to the network structure of the games without explicit knowledge of the utility functions.

## 7.3 Setting

In this section, we start by analysing three commonly studied network games ( Sec. 7.3.1 and Sec. 7.3.2). Based on the specific utility function of these games, we establish a generic relationship between the equilibrium actions and network structure ( Sec. 7.3.3). This eventually motivates the proposed framework that learns network structure without any knowledge of the utility function[1] ( Sec. 7.4).

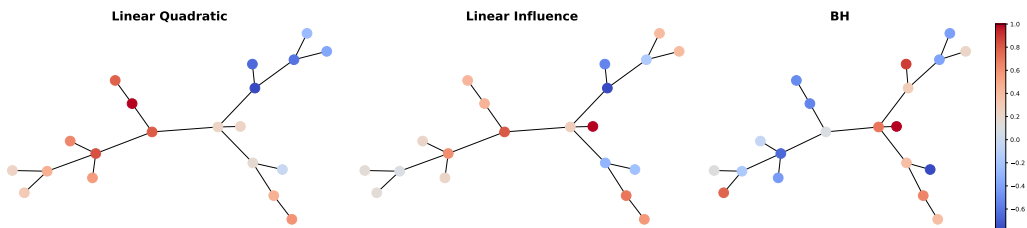

Figure 1: Example actions for different types of games on a Barabási-Albert graph. Actions are normalised to be in $[-1, 1]$ and are displayed as colors on nodes.

### 7.3.1 CONTINUOUS-ACTION NETWORK GAMES

In a network game, the payoff $u_i$ of an individual $i$ depends on their action $x_i$ as well as the actions of neighbours $j \in \mathcal{N}_i$. We consider three commonly studied network games: linear quadratic games [9], a variation of the linear influence games developed in Irfan et al. [120], and the graphical game studied in Barik et al. [14] (which we will refer to as Barik-Honorio or BH graphical game).
**Linear quadratic games.** Linear quadratic games are widely studied in the economics literature [9, 121]. In this game, a player $i$ chooses their action by maximising the following utility function:

$$\max_{\{x_i\}} u_i = b_i x_i - \frac{1}{2} x_i^2 + \beta \sum_{j \in \mathcal{N}_i} a_{ij} x_i x_j, \quad (7.1)$$

where $b_i$ represents the marginal benefit of $i$ by taking action $x_i$ and $\beta$ is the strength of dependencies between actions of neighbours in the network, respectively. Note that this utility function can also be thought of as a second-order approximation to non-linear utility functions of more complex games. The pure-strategy Nash equilibrium (PSNE) of this game is

$$\mathbf{x}^* = (\mathbf{I} - \beta \mathbf{A})^{-1} \mathbf{b}, \quad (7.2)$$

where $\mathbf{x}^*$ and $\mathbf{b}$ are $N$-dimensional vectors collecting actions and marginal benefits for all individuals, and $\mathbf{I}$ is the $N \times N$ identity matrix. Under the assumption that $|\beta| < 1$, the matrix inverse is guaranteed to exist as the spectral radius of $\mathbf{A}$ is 1. Furthermore, when $\beta > 0$, the game corresponds to a strategic complement relationship (i.e., intuitively, the incentive of a player to take a higher action is increasing in the number of their neighbours also taking a higher action); when $\beta < 0$, it corresponds to strategic substitute (i.e., intuitively, the incentive of a player to take a higher action is decreasing in the number of their neighbours also taking a higher action).
**Linear influence games.** Inspired by the threshold model [96], Irfan et al. [120] proposed the linear influence games, where an individual chooses the action that maximises the following utility function[2]:

$$\max_{\{x_i\}} u_i = \sum_{j \in \mathcal{N}_i} a_{ij} x_i x_j - b_i x_i, \quad (7.3)$$

---

[1]The utility function is used only in analysing three classical games in a universal framework as well as generating data in synthetic experiments. It is not used by the proposed method itself.
[2]The actions are discrete in the originally proposed game. We adapt the game to a continuous setting.

Table 1: Parameterisation of three network games.

|  | $\mathcal{F}(\mathbf{A})$ | $\mathcal{H}(\mathbf{b})$ |
| --- | --- | --- |
| Linear quadratic | $(\mathbf{I} - \beta \mathbf{A})^{-1}$ | $\mathbf{b}$ |
| Linear influence | $\mathbf{A}^{-1}$ | $\mathbf{b}$ |
| Barik-Honorio | $\mathbf{u}_1$ | 1 |

where $b_i$ can be understood as a threshold parameter for $i$'s level of tolerance for negative effects. Under the assumption that $\mathbf{A}$ is invertible, the PSNE satisfies the following condition:

$$\mathbf{x}^* = \mathbf{A}^{-1}\mathbf{b}. \tag{7.4}$$

**BH graphical game.** Finally, in the specific graphical game introduced in Barik et al. [14], an individual maximises the following utility function:

$$\max_{\{x_i\}} u_i = -\left\| x_i - \sum_{j \in \mathcal{V}} a_{ij} x_j \right\|_2. \tag{7.5}$$

This utility can be used to model games where an individual prefers to conform to the social norm (actions of their neighbours), since their own utility decreases as their action deviates from those of their neighbours. The PSNE for this game satisfies the following condition:

$$\mathbf{x}^* = \mathbf{A}\mathbf{x}^*. \tag{7.6}$$

This suggests that $\mathbf{x}^*$ is the eigenvector $\mathbf{u}_1$ of $\mathbf{A}$ which is associated with the largest eigenvalue (which is 1). We consider actions from the set of $\epsilon$-PSNE [14], which are obtained by adding noise independently per player. The observed actions $\mathbf{x}$ are: $\mathbf{x} = \mathbf{x}^* + \mathbf{e}$, where $\mathbf{e}$ is Gaussian noise.

Example actions for the three types of games on a Barabási-Albert graph have been illustrated in Fig. 1. By investigating Eq. (7.2), (7.4), and (7.6), we can write the condition for the equilibrium actions $\mathbf{x}^*$ in a generic form[3] (see Tab. 1):

$$\mathbf{x}^* = \mathcal{F}(\mathbf{A})\mathcal{H}(\mathbf{b}), \tag{7.7}$$

where $\mathcal{F}(\mathbf{A})$ is a function of the network structure and $\mathcal{H}(\mathbf{b})$ is a function of additional parameters (if any) associated with the game. That is, $\mathcal{F}(\mathbf{A})$ accounts for the influence of the actions of one's neighbours in the network. Conversely, $\mathcal{H}(\mathbf{b})$ is only affected by one's idiosyncratic (individual) characteristics.

### 7.3.2 Modeling of Individual Idiosyncratic Characteristics

Under the linear quadratic or linear influence games, the parameter $\mathbf{b}$ captures the marginal benefits or tolerance levels as idiosyncratic characteristics of players in the corresponding games. In the presence of the homophily effect [169], we may assume that this parameter is associated with the network structure. To this end, we propose to model $\mathbf{b}$ as follows:

$$\mathbf{b} \sim \mathcal{N}(\mathbf{0}, \mathbf{L}_\alpha^\dagger), \qquad \mathbf{L}_\alpha = (1-\alpha)\mathbf{I} + \alpha\mathbf{L} \tag{7.8}$$

---

[3] We tacitly assume that $|\beta| < 1$ and $\mathbf{A}$ is invertible.

where $\mathbf{L} = \mathbf{I} - \mathbf{A}$ is the normalised graph Laplacian matrix, and $^\dagger$ represents pseudoinverse in case the matrix is not invertible (this happens when $\alpha = 1$ as $\mathbf{L}$ has a smallest eigenvalue of 0). The parameter $\alpha \in [0, 1]$ controls the relation of the individual idiosyncratic characteristics to the network structure. The two corner cases $\alpha = 0$ (for which we have $\mathbf{L}_0 = \mathbf{I}$) and $\alpha = 1$ (when $\mathbf{L}_1 = \mathbf{L}$) correspond to independent idiosyncratic characteristics and homophilous idiosyncratic characteristics (individuals with similar characteristics tend to be connected), respectively. By varying $\alpha$ from 0 to 1 we can achieve increasing levels of homophily or smoothness (see Sec. 7.3.3) of $\mathbf{b}$ on the graph.

### 7.3.3 Analysis of Equilibrium Actions

With the conditions for equilibrium actions in Sec. 7.3.1 and the modelling of individual idiosyncratic characteristics in Sec. 7.3.2, we can analyse the characteristics of these actions explicitly.

**Linear quadratic games.** Assuming $\mathbf{b}$ of the form (7.8) and using Eq. (7.2) and $\mathbf{L} = \mathbf{I} - \mathbf{A}$, we have that the equilibrium actions $\mathbf{x}^*$ follow a multivariate Gaussian distribution:

$$\begin{aligned}
\mathbf{x}^* &\sim \mathcal{N}\left(\mathbf{0}, (\mathbf{I} - \beta\mathbf{A})^{-1}(\mathbf{I} - \alpha\mathbf{A})^\dagger(\mathbf{I} - \beta\mathbf{A})^{-1}\right) \\
&= \mathcal{N}\left(\mathbf{0}, \mathbf{U}[(\mathbf{I} - \beta\mathbf{\Lambda})^2(\mathbf{I} - \alpha\mathbf{\Lambda})]^\dagger\mathbf{U}^\top\right),
\end{aligned} \quad (7.9)$$

with the eigendecomposition $\mathbf{A} = \mathbf{U}\mathbf{\Lambda}\mathbf{U}^\top$. Eq. (7.9) illustrates the relationship between the actions $\mathbf{x}$ and the network structure $\mathbf{A}$ and motivates the learning framework proposed in the next section. Furthermore, the covariance in Eq. (7.9) may be interpreted as a graph filter whose frequency response $\frac{1}{(1-\beta\lambda)^2(1-\alpha\lambda)}$ may shed light on the behaviour of the actions. For $\beta \to 1$ and $\alpha \to 1$, the action vector $\mathbf{x}$ tends to behave like the leading eigenvectors of $\mathbf{A}$ ('low frequency'), which are smooth on the graph[4]. An example of the filter response (with $\beta = 0.8$ and $\alpha = 0.8$) applied on the eigenvalues of an instance of a 20-node Erdős-Rényi graph is shown in Fig. 2 (blue). This shows that in this case the actions of linear-quadratic games are dominated by the leading eigenvector of A which is smooth.

**Linear influence games.** Similarly, from Eq. (7.4) and (7.8), we have that the equilibrium actions $\mathbf{x}$ follow a multivariate Gaussian distribution:

$$\begin{aligned}
\mathbf{x}^* &\sim \mathcal{N}\left(\mathbf{0}, \mathbf{A}^{-1}(\mathbf{I} - \alpha\mathbf{A})^\dagger\mathbf{A}^{-1}\right) \\
&= \mathcal{N}\left(\mathbf{0}, \mathbf{U}[\mathbf{\Lambda}^2(\mathbf{I} - \alpha\mathbf{\Lambda})]^\dagger\mathbf{U}^\top\right).
\end{aligned} \quad (7.10)$$

Interpreting Eq. (7.10) as a spectral filter of the form $\frac{1}{\lambda^2(1-\alpha\lambda)}$ we can also conclude that an $\alpha \to 1$ tends to lead to smoother actions on the graph. However, the exact behaviour of actions in this case depends on the magnitude of the eigenvalue of $\mathbf{A}$ closest to 0. Given that the spectrum of $\mathbf{A}$ lies in the range of $[-1, 1]$, the actions are likely to behave like mid-spectrum eigenvectors, which are not necessarily smooth signals on the graph. Similarly, this can be seen from the filter response, under $\alpha = 0.8$, shown in Fig. 2 (red). This shows that in linear influence games, the actions are dominated by mid-spectrum eigenvectors, which are not necessarily smooth.

**BH graphical game.** We see from Eq. (7.6) that the equilibrium actions correspond to the largest eigenvector $\mathbf{u}_1$ of $\mathbf{A}$. Although in this setting the observed actions are $\epsilon$-PSNE, they would still tend to be smooth on the graph.

---

[4] By 'smoothness' here we mean the Dirichlet energy of the features, i.e., trace($\mathbf{X}^\top\mathbf{L}\mathbf{X}$).

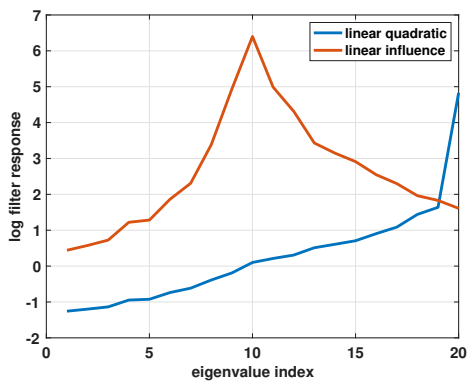

Figure 2: Interpreting actions of linear quadratic/influence games: the equilibrium actions for the two games are dominated by different sets of eigenvectors.

In summary, our analysis in this section shows that the smoothness of the equilibrium actions on the graph depends on $\beta$ and $\alpha$ in the linear quadratic game. The actions in linear influence games are likely to be nonsmooth, whereas those in the BH graphical games are likely to be smooth. We empirically validate this analysis in Sec. 7.5.1. Regardless of the smoothness, the relationship between the equilibrium actions and the network structure demonstrated in this section motivates us to propose a learning framework in the next section to infer the network structure from the observed actions.

## 7.4 Proposed Approach

Motivated by the analysis in Sec. 7.3.3, we propose a model that learns a direct mapping from the observed actions to the network structure. Such a model is agnostic to the utility function of the game and avoids strong assumptions on it (as long as it conforms to the broad class of games whose equilibrium actions can be parameterised by Eq. (7.7)). Specifically, we consider the scenario where the social network and decision data exist for a small population. The objective is to learn the mapping from decisions to the network structure on this small population such that it can be used to infer a large-scale unobserved network. For example, for cost-effective data collection, government, public agencies, and researchers can collect social network data on a small population (by asking individuals to nominate their friends) and then use the proposed method to learn the network interactions for a larger population.

In our setting, we assume to have a training set $\mathcal{D}$ of action-graph pairs $(\mathbf{X}^{(\ell)}, \mathbf{A}^{(\ell)})$ coming from games with the same (but unknown) utility function. For each $\ell$, the model takes as input an $N \times K$ matrix $\mathbf{X}^{(\ell)}$ containing the actions of $N$ players over $K$ independent games as columns, and outputs a predicted $N \times N$ adjacency matrix $\hat{\mathbf{A}}^{(\ell)} = g_\Theta(\mathbf{X}^{(\ell)})$ for the network game (we drop $\ell$ from now on for simplicity). Once trained, our model can then infer the network structure corresponding to previously unseen actions, as long as they are generated following a similar utility function. Moreover, this framework makes our model generalisable to learning networks with the number of nodes different from what was observed during training.

The model parameters are learned such that they minimise the cross-entropy between the binary ground truth adjacency matrix $\mathbf{A}$ and the predicted continuous one $\hat{\mathbf{A}}$. Binary cross-entropy is a standard loss function for link prediction tasks using graph neural networks [102], of which our problem is an instance. We also experimented with weighting the loss based on the proportion of edges in the graph, but it did not change the performance of the model.

Our model follows an encoder-decoder architecture $g_\Theta(\mathbf{X}) = \text{dec}(\text{enc}(\mathbf{X}))$, which is a standard solution for link-prediction problems in graph neural networks [135] since the resulting number of parameters of the model is independent on the size of the graph, allowing for more statistical efficiency, and for the same model to work on graphs of different sizes. The *encoder* outputs an $N \times K \times F$ tensor $\mathbf{Z} = \text{enc}(\mathbf{X})$, mapping each node $i$ in game $k$ to an $F$-dimensional latent embedding $\mathbf{z}_{ik}$. The *decoder* outputs the predicted $N \times N$ adjacency matrix $\hat{\mathbf{A}} = \text{dec}(\mathbf{Z})$, where the score $\hat{a}_{ij}$ for each edge $(i, j)$ is computed using the $K \times F$-dimensional embeddings $\mathbf{Z}_i, \mathbf{Z}_j$ of the respective nodes.

Given that the ordering of nodes in the graph is arbitrary, the model has to be a *permutation-equivariant* [34] function over the set of nodes: $g_\Theta(\mathbf{P}_1 \mathbf{X}) = \mathbf{P}_1 g_\Theta(\mathbf{X}) \mathbf{P}_1^\top$, where $\mathbf{P}_1$ is an $N \times N$ permutation matrix interpreted as reordering of the nodes of the graph. Additionally, in the case where there is no correspondence between game $k$ in one graph and game $k$ in another one (e.g., actions corresponding to user rating where however users in different graphs have rated different items), the model should also be *permutation-invariant* over the set of games, i.e., $g_\Theta(\mathbf{X}\mathbf{P}_2) = g_\Theta(\mathbf{X})$, where $\mathbf{P}_2$ is another $K \times K$ permutation matrix on the games. Overall, this combined symmetry condition can be written as $g_\Theta(\mathbf{P}_1 \mathbf{X} \mathbf{P}_2) = \mathbf{P}_1 g_\Theta(\mathbf{X}) \mathbf{P}_1^\top$.

ENCODER  The input to our model is a variable-length set (of actions), which excludes using a multi-layer perceptron (since it would not be able to handle the variable length) and sequence models such as LSTMs [108] or GRUs [51] (since they treat the input as an ordered sequence, while it is an unordered set). Moreover, since the ground truth graph between the players is not known a priori, but it is however important for the model to exchange information between players (as the values of agents' actions are meaningless if not compared to the ones of the others), the encoder needs to perform message passing on the fully connected graph of players. Therefore, we propose a Transformer-like [243] encoder (which we refer to as Network Game Transformer or *NuGgeT*). *NuGgeT* processes a *set* of games happening on a given network and outputs for each node a *set* of $K$ different game-specific embeddings (one for each game that happens on the same network). In this architecture, the value of action $x_{ik}$ of player $i$ in game $k$ is never mixed with the actions of other players in a different game $k'$, if not for the computation of an aggregated attention score $\alpha_{ij}$ that captures the overall similarity between players $i$ and $j$. Intuitively, in settings where we are given the outcomes of multiple games played on multiple graphs, there is no correspondence between such games. Thus, the value of an action does not bring any information about the role of a node in a graph and the only useful information that needs to be exchanged across games is the similarity of the nodes.

Specifically, for each node $i = 1, \ldots, N$ and game $k = 1, \ldots, K$, the embedding $\mathbf{z}_{ik}$ is computed as follows:

$$\mathbf{y}_{ik} = \text{ReLU}(x_{ik}\boldsymbol{w} + \mathbf{b}) \tag{7.11}$$

$$\alpha_{ij}^{(h)} = \text{softmax}_j\Big( \bigsqcup_{k=1}^{K} \mathbf{y}_{ik}^\top \mathbf{W}_Q^{(h)} \mathbf{W}_K^{(h)} \mathbf{y}_{jk} \Big)$$

$$\mathbf{z}_{ik} = \phi\Big(\mathbf{y}_{ik}, \sum_{j=1}^{N} \alpha_{ij}^{(1)} \mathbf{y}_{jk}, \ldots, \sum_{j=1}^{N} \alpha_{ij}^{(H)} \mathbf{y}_{jk}\Big)$$

where $\square$ denotes a general permutation-invariant aggregation operator (e.g max, mean $\frac{1}{K} \sum_{k=1}^{K}$ or sum $\sum_{k=1}^{K}$), $\psi$ is some learnable function, $\boldsymbol{w} \in \mathbb{R}^F$, $\mathbf{b} \in \mathbb{R}^F$, $\mathbf{W}_K^{(h)} \in \mathbb{R}^{F \times F'}$, and $\mathbf{W}_Q^{(h)} \in \mathbb{R}^{F' \times F}$ are learnable parameters, and $h = 1, \ldots, H$ denotes the attention heads. For a generic node $i$ in game $k$, NuGgeT first expands $i$'s action $x_{ik}$ into a vector $\mathbf{y}_{ik}$ of $F$ features, then computes $H$ attention scores (one for each head) for each pair of nodes $(i, j)$ via multiplicative attention on the expanded actions $\mathbf{y}_{\cdot k}$[5] (aggregating the unnormalised scores across games), and finally refines vector $\mathbf{y}_{ik}$ via a function $\psi$ that processes the $H$ aggregated representation of the neighbours obtained from the attention mechanism.

DECODER   In the case of directed graphs, there is no specific requirement for the decoder. If the graph is undirected, and therefore the adjacency matrix is symmetric, the decoder should be symmetric w.r.t. pairs of nodes (i.e., $(i, j)$ and $(j, i)$ treated the same way). Additionally, we impose invariance w.r.t. ordering of the games. *NuGgeT*'s decoder computes the predicted adjacency $\hat{a}_{ij}$ by aggregating the game-specific embeddings computed in Eq. (7.12) for a pair of nodes $(i, j)$ using a general permutation-invariant aggregation operator $\boxed{\cdot}$, which is then passed through a learnable function $\psi$:

$$\hat{a}_{ij} = \psi\Big(\boxed{\cdot}_{k=1}^{K} \mathbf{z}_{ik} \odot \mathbf{z}_{jk}\Big). \tag{7.12}$$

Here $\odot$ denotes element-wise product, whose use ensures symmetry w.r.t. node pairs (since $\mathbf{z}_{ik} \odot \mathbf{z}_{jk} = \mathbf{z}_{jk} \odot \mathbf{z}_{ik}$), while the permutation invariant operator over $k$ ensures invariance to the ordering of games. We empirically observe this approach to work better compared to simpler permutation-invariant functions such as the dot product of the concatenation of the embeddings from multiple games. In the SM we prove that *NuGgeT* satisfies the symmetry conditions outlined above.

## 7.5 EXPERIMENTS

Our implementation of *NugGeT* uses the sum $\sum_{k=1}^{K}$ as the permutation-invariant functions $\boxed{\phantom{x}}$ and $\boxed{\cdot}$, and two different 2-layer MLPs for $\phi$ and $\psi$. We use the Adam optimiser [132] with a learning rate of $0.001$, a batch size of $100$ and a patience of $50$ epochs. We did not perform any particular hyperparameter tuning for our method, since we found it to be quite robust to the choice of hyperparameters and perform well with standard choices. In all our experiments, we report the mean and the standard error of the mean over the test graphs. Note that we ignore diagonal elements of the adjacency matrix both for training and evaluation. Since they are always zero, the model could easily memorise them, influencing the metrics. We use an AWS p3.16xlarge machine with 8 GPUs. While the training of our model takes between 5 and 10 minutes on a single GPU, the whole set of experiments conducted in the paper necessitates roughly 4 days of GPU time.

### 7.5.1 SYNTHETIC DATA

DATA GENERATION   We follow the setup in Leng et al. [145] for generating the synthetic graphs using three different random graph models: Erdős-Rényi (ER), Watts-Strogatz (WS), and Barabási-Albert (BA). More details and exact parameters for the synthetic data are provided in Sec. 7.C. All the graphs in our experiments have $N = 20$ vertices. For each type of graph above, we simulate equilibrium actions for linear quadratic, linear influence, and BH graphical games using their respective utility function. For linear quadratic games, once the graphs are constructed, we compute $\beta > 0$ such that the spectral radius $\rho(\beta A)$ varies between 0 and 1. For linear influence games, when the adjacency matrix is not invertible, we take its pseudoinverse. For BH games, we set std($\mathbf{e}$) $= 1$ and ensure the resulting actions are $\epsilon$-PSNE with $\epsilon = 0.2$. The generated actions, together with the ground truth network structure, are used to train the model. We use 850 graphs in the training set, 50 graphs for validation and 100 graphs for testing and verify that there is no overlap between them. It is important to notice that while $\alpha$, $\beta$ and $\epsilon$ are used to generate the synthetic data, they are not used or known by the model.

---

[5] For a given player $i$, using the expanded actions $\mathbf{y}_{\cdot k}$ rather than input scalars $x_{\cdot k}$ allows the attention mechanism to produce attention scores which are not necessarily linearly dependent on the value of neighbours' action $x_{jk}$, thus producing richer attention scores

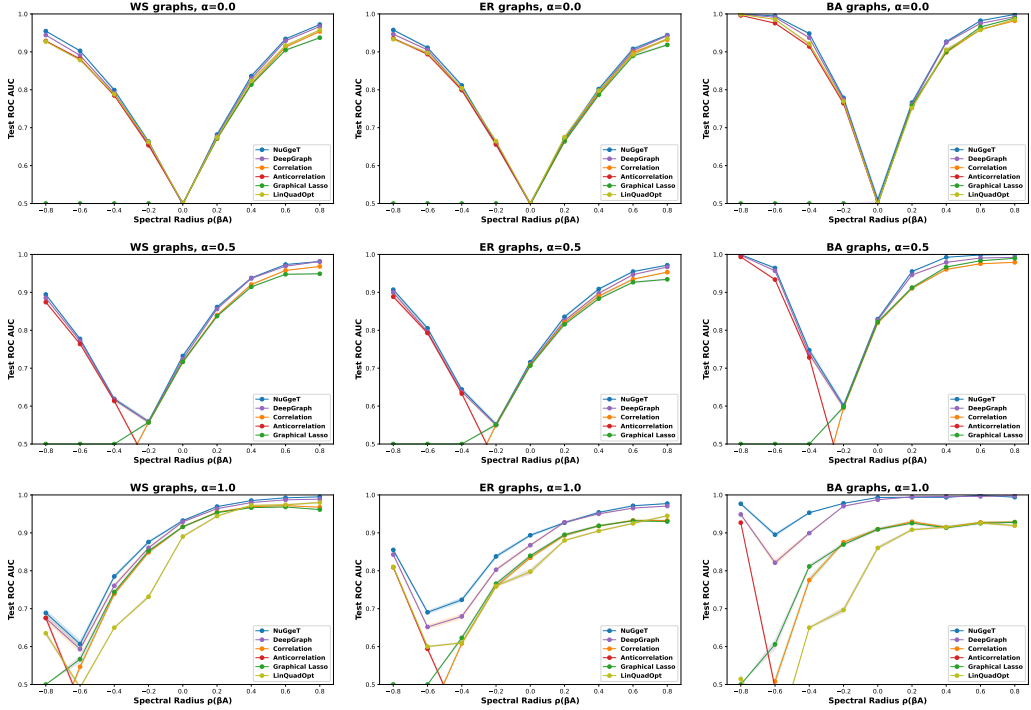

Figure 3: Results for linear quadratic games with varying $\alpha$ and spectral radius $\rho(\beta A)$.

BASELINES We compare with the following general baselines: *Correlation*, *Anticorrelation*, *Graphical Lasso* [76] and *DeepGraph* [21]. We also compare to game-specific baselines: *LinQuadOpt (independent)* [145] and *LinQuadOpt (homophilous)* [145] for linear quadratic games, and *BlockRegression* [14] for BH Graphical Games. A more detailed description of the baselines and how they are tuned is provided in Sec. 7.E.

RESULTS Results for Linear Quadratic Games are reported in Fig. 3. Columns are different types of graphs (ER, WS, BA), rows are different values of $\alpha$ that controls the smoothness of the marginal benefits ($\alpha = 0$, $\alpha = 0.5$ and $\alpha = 1$), while the x-axis represents the spectral radius $\rho(\beta A)$. *NuGgeT* is on par or superior to other methods in all scenarios, with *DeepGraph* being the runner-up competitor. We observe larger performance gap with $\alpha = 1$, i.e., when the distribution of marginal benefits largely depends on graph structure. Interestingly, it can be observed that in this case *NuGgeT* and *DeepGraph* perform well in both cases of strategic complements ($\rho(\beta A) > 0$, neighbours take similar actions) and strategic substitutes ($\rho(\beta A) < 0$, neighbours take opposite actions), whereas other baselines only perform well in one of the two cases. This is due to the former two methods learning directly a mapping between actions and graph structure that may correspond to different characteristics of the equilibrium actions.

For Linear Influence Games, we report the results in Fig. 4. Each plot corresponds to a different type of graph, and the x-axis represents the benefit smoothness $\alpha$. As expected, all methods improve their performance as $\alpha$ grows, since the actions generally become smoother over the graph ( Sec. 7.3.3). Again, *NuGgeT* outperforms the baselines in all scenarios. Interestingly, the performance on WS and ER graphs seems to be much lower than on BA graphs. This can be understood empirically by analysing the eigenvalues of the normalised adjacency matrix $\mathbf{A}$ for different graphs. For WS and ER graphs, the smallest absolute (non-zero) eigenvalue of $\mathbf{A}$ is, on average, much smaller than for BA graphs ( Fig. 7.G.1). As explained in Sec. 7.3.3, eigenvalues with very small absolute values will result in actions behaving like mid-spectrum eigenvectors (which are not necessarily smooth). We verify this further in Fig. 7.G.2, which shows the spectral coefficients of the actions for all combinations of graphs and games. Linear Influence Games on ER and WS graphs are indeed the only scenarios where we empirically observe large graph Fourier coefficients for mid-spectrum eigenvectors. On BA graphs we do not observe this behaviour; mid-spectrum eigenvectors are most often not represented at all by the actions (the BA graphs we generate are trees that have zero eigenvalues associated with mid-spectrum eigenvectors which are discarded when taking the pseudoinverse), and actions tend to be smoother in this case. The results in Fig. 4 suggest that when actions are less smooth (weaker association of actions with graph structure), all methods tend to perform less well.

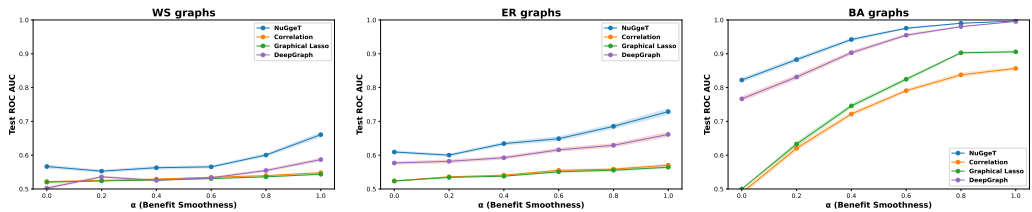

Figure 4: Results on linear influence games when varying the smoothness $\alpha$ of the marginal benefits.

Results for BH Graphical Games are reported in Fig. 5. Since there are no parameters controlling the game, we only have 3 configurations corresponding to different graph types. Again, *NuGgeT* outperforms other methods in all configurations, with the gap being largest on BA graphs. We also perform ablation studies on the number of games, number of training graphs and size of the graphs. The results are presented and discussed in Sec. 7.H.

In many real-world scenarios, the observed actions will not be exactly at equilibrium but close to it, or trying to re-converge to the equilibrium after some perturbation. We, therefore, investigate how the performance of our model degrades with "noisy" samples where the sampled actions are only near equilibrium. Tab. 2 shows the results for BA graphs and signals with unit norm. We can see performance remains satisfactory given reasonable amount of noise.

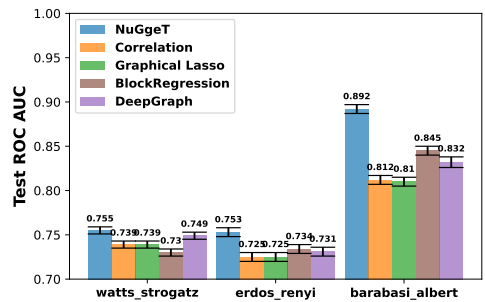

Figure 5: Results for BH graphical games.

Table 2: Test accuracy on Barabási-Albert graphs for NuGgeT given different levels of Gaussian noise added to player actions.

| Noise Std | Linear Quadratic | Linear Influence | BH |
|---|---|---|---|
| no noise | 99.87±0.02 | 99.87±0.02 | 99.87±0.02 |
| 0.10 | 99.33±0.10 | 99.33±0.10 | 99.33±0.10 |
| 0.20 | 95.50±0.38 | 95.50±0.38 | 95.50±0.38 |

To conclude, our method outperforms the baselines in all three types of games, most significantly in the scenarios where the actions are not necessarily smooth on the network and our method was able to infer such relationship from the observed data. While learning the mapping from actions to network, the model will learn not only the utility function of the games represented in the data, but also the typical structure of the networks. If all the networks in the training data are Barabási-Albert graphs, the prediction of the model will likely also be a Barabási-Albert graph. On the other hand, optimisation methods such as LinQuadOpt are not able to automatically bias their prediction to a particular class of graphs.

### 7.5.2 Real-World Data

Players in real-world scenarios often act according to strategic interactions. It has been shown in the sociology and game theory literature that people have an incentive to conform to social norms [170, 273] or be influenced by their social network neighbours [145]. That is, individual utilities are higher if their behaviours are similar to those of their neighbours in the social network. Such a mechanism will lead to strategic complement relationships. Following this assumption, we validate our model on two datasets, comparing with the baseline methods introduced in Sec. 7.5.1.

INDIAN VILLAGES   The *Indian Villages* dataset[6] [10] is a well-known dataset in the economics literature that contains data from a survey of social networks in 75 villages in rural southern Karnataka, a state in India. Each village constitutes a social network graph, where nodes are households and edges are self-reported friendships. Following the setup in Leng et al. [145], we consider as actions the number of rooms, number of beds and other decisions families have to make related to their household. The reasoning is that if neighbours adopt a specific facility, villagers tend to gain higher payoff by doing the same, i.e., complying with social norms. We only consider the 48 villages for which we have the ground truth actions and network. 40 are used for training, 3 for validation and 5 for testing. Categorical actions are one-hot encoded, while numerical actions are treated as continuous features. The resulting dataset has graphs with 10 actions and a number of nodes ranging between 77 and 356. It can be seen from Tab. 3 that *NuGgeT* outperforms all other methods by at least 5.01%. *DeepGraph* fails to learn altogether on this dataset.

YELP RATINGS   The *Yelp Ratings* dataset[7] consists of rating of users to business, as well as the social connectivity between users. Similarly to the previous case, on deciding to review a local business, people may have an incentive to conform to the social norms they perceive, which are formed by the ratings from their neighbours. Yelpers tend to gain higher payoffs with similar ratings due to social conformity, i.e., strategic complements in a game-theoretic context. From the raw data, we extracted 5000 sub-graphs representing communities, where the actions are the average rating of users for 22 categories of businesses. The task is to reconstruct the social connectivity of the users given their actions (ratings). 4250 graphs are used for training, 250 for validation and 500 for testing. More details of the dataset construction are provided in the Sec. 7.D. On this dataset, NuGgeT outperforms all other baselines by at least 2.79%. Overall, the results on real-world data show the efficacy of *NuGgeT* in cases where the game utility is not explicitly known.

---

[6]The Indian Villages dataset can be accessed at https://doi.org/10.7910/DVN/U3BIHX.
[7]The Yelp dataset can be accessed at https://www.yelp.com/dataset.

*7 Game-Theoretic Structural Inference*

Table 3: Test ROC AUC for Indian Villages and Yelp Ratings data.

| Model | Indian Villages | Yelp Ratings |
| --- | --- | --- |
| Correlation | $0.5816 \pm 0.0135$ | $0.6222 \pm 0.0043$ |
| Anticorrelation | $0.4184 \pm 0.0135$ | $0.3778 \pm 0.0043$ |
| Graphical Lasso | $0.5823 \pm 0.0152$ | $0.6523 \pm 0.0038$ |
| Baraki and Honorio | $0.5715 \pm 0.0164$ | $0.6786 \pm 0.0032$ |
| LinQuadOpt (indep.) | $0.5557 \pm 0.0108$ | $0.6796 \pm 0.0033$ |
| LinQuadOpt (homop.) | $0.5789 \pm 0.0174$ | $0.6310 \pm 0.0036$ |
| DeepGraph | $0.4965 \pm 0.0143$ | $0.6776 \pm 0.0039$ |
| NuGgeT (Ours) | $\mathbf{0.6324 \pm 0.0167}$ | $\mathbf{0.7057 \pm 0.0035}$ |

## 7.6 Discussion

Limitations and Future Work.    The current work only deals with static games and networks of small scale. Moreover, we do not deal with repeated games, i.e., where players have to take multiple actions sequentially. A promising future direction is therefore to extend it to dynamic games and networks, where both the utility function and the structure of the network may change over time, as well as dealing with larger graphs and repeated games. Moreover, we currently deal with test graphs that have a similar structure to the graphs the model has been trained on, e.g., we use the Barabási-Albert graphs during both training and testing. An interesting future work would be to work on the generalisation capability of this approach, i.e., the ability to be trained on one type of graph (e.g., Erdős-Rényi) but generalise to different types as well (e.g., Barabási-Albert or Watts-Strogatz). In addition, our method cannot guarantee the uniqueness of the learned network; indeed, the main focus of the study is to propose a first and efficient data-driven learning framework without assuming the utility function and prior knowledge about the network structure. We leave the identification of the network structure, which is a challenging problem in itself, for future work.

## 7.7 Conclusion

In this work, we propose a novel framework to infer the network structure behind the games from their equilibrium actions. Unlike existing methods, we achieve so by learning a mapping from the actions to the network structure without knowing the utility function of the game. This is especially beneficial in real-world scenarios where the nature of strategic interactions between players of the game remains hidden or may evolve over time.

FAST FORWARD TO TODAY.   Two years after its publication, our work has yet to inspire direct follow-ups or applications. We believe that the cause of this lies in its niche applicability. Although our framework is tailored for scenarios where the network structure is unknown, it requires a dataset comprising hundreds of (action, network) pairs for model training. This requirement potentially limits its practical deployment and broader utilization in the field.

# Appendices

## 7.A Proof of NuGgeT symmetries

We would like to show that the *NuGgeT* model $g$ is *permutation-invariant* over the set of games and *permutation-equivariant* over the set of nodes.

Starting with the first one, we need to show that $g(\mathbf{X}\mathbf{P}_2) = g(\mathbf{X})$, where $\mathbf{P}_2$ is a $K \times K$ permutation matrix on the games. For what concerns the encoder, we only need to prove that $\alpha_{ij}^{(h)}$ is invariant w.r.t. permutation of the games. Let $p_k$ be the index of the non-zero entry of the $k$-th column of $P_2$, we have:

$$\alpha_{ij}^{(h)} = \boxdot_{k=1}^{K} \mathbf{y}_{ik}^\top \mathbf{W}_Q^{(h)} \mathbf{W}_K^{(h)} \mathbf{y}_{jk}$$
$$= \boxdot_{k=1}^{K} \mathbf{y}_{ip_k}^\top \mathbf{W}_Q^{(h)} \mathbf{W}_K^{(h)} \mathbf{y}_{jp_k},$$

since $\boxdot_{k=1}^{K}$ is chosen to be a permutation invariant operator. This shows that $\alpha_{ij}^{(h)}$ does not depend on the particular permutation $p$ chosen, i.e., it is invariant w.r.t. permutation of the games. Similarly, the same holds for the decoder as $\boxdot_{k=1}^{K}$ is a permutation invariant operator and therefore $\boxdot_{k=1}^{K} \mathbf{z}_{ik} \odot \mathbf{z}_{jk} = \boxdot_{k=1}^{K} \mathbf{z}_{ip_k} \odot \mathbf{z}_{jp_k}$ for any permutation $p$.

Regarding the second part of the proof, we need to show that $g(\mathbf{P}_1\mathbf{X}) = \mathbf{P}_1 g(\mathbf{X}) \mathbf{P}_1^\top$, where $\mathbf{P}_1$ is an $N \times N$ permutation matrix interpreted as reordering of the nodes of the graph. Letting $p_k$ be the index of the non-zero entry of the $k$-th row of $\mathbf{P}_1$. Note that:

$$\text{enc}(\mathbf{P}_1\mathbf{X})_{ik} = \psi\Big(\mathbf{y}_{p_ik}, \sum_{j=1}^{N} \alpha_{p_ip_j}^{(1)} \mathbf{y}_{p_jk}, \ldots, \sum_{j=1}^{N} \alpha_{p_ip_j}^{(H)} \mathbf{y}_{p_jk}\Big)$$
$$= \psi\Big(\mathbf{y}_{p_ik}, \sum_{j=1}^{N} \alpha_{p_ij}^{(1)} \mathbf{y}_{jk}, \ldots, \sum_{j=1}^{N} \alpha_{p_ij}^{(H)} \mathbf{y}_{jk}\Big)$$
$$= \mathbf{z}_{p_ik}$$

where the second equality stems from the fact that summation is a permutation invariant operator (i.e., $\sum_{j=1}^{N} x_i = \sum_{j=1}^{N} x_{p_i}$ for any permutation $p$). We then have:

$$\left(\mathbf{P}_1 \, g(\mathbf{X}) \mathbf{P}_1^\top\right)_{ij} = \hat{a}_{p_i p_j}$$

$$= \psi\Big(\boxed{\cdot}_{k=1}^{K} \mathbf{z}_{p_i k} \odot \mathbf{z}_{p_j k}\Big)$$

$$= \psi\Big(\boxed{\cdot}_{k=1}^{K} \text{enc}(\mathbf{P}_1 \mathbf{X})_{ik} \odot \text{enc}(\mathbf{P}_1 \mathbf{X})_{jk}\Big)$$

$$= g(\mathbf{P}_1 \mathbf{X})_{ij}.$$

Since this holds for all indices $ij$, it follows that $g(\mathbf{P}_1 \mathbf{X}) = \mathbf{P}_1 g(\mathbf{X}) \mathbf{P}_1^\top$.

## 7.B Experiments

## 7.C Synthetic Data Generation

We generate the synthetic data from three different graph models: Erdős-Rényi (ER), Barabási-Albert (BA) and Watts-Strogatz (WS). In ER graphs, an edge is present with a probability of $p = 0.2$, independently from all other possible edges. In WS graphs, we set the exact degree of the nodes to be $k = log_2(N)$, with a probability of $p = 0.2$ for the random rewiring process. Finally, in BA graphs, nodes are added one at a time and each new node has $m = 1$ edges which are preferentially attached to existing nodes with already high degree (this results in a tree graph).

## 7.D Yelp Dataset Generation

Starting from the raw data at https://www.yelp.com/dataset, we create a dataset by performing the following steps:

1. We compute the rating for every user to every business category, by averaging the ratings a user has given to all businesses of each category

2. We weight each edge in the original user-user graph with the fraction of common categories the two users rated at least once

3. We cluster the above weighted graph using Graclus [63] with the objective of minimising the normalised cut.

4. We discard all the clusters with less than 10 nodes, or with very sparse ratings (less than 25% of the categories with a rating for at least 25% of the users). The result of this step is ∼27k graphs of different users.

5. We rank the clusters extracted above by their density of ratings and keep the 5000 most dense clusters. Each cluster constitutes a graph associated with node attributes, which can be used to train and test our model.

## 7.E Baselines

In both the synthetic and real-world data experiments we compare with the following baselines:

| Hyperparameter Name | Value |
| --- | --- |
| $F$ | 10 |
| $F'$ | 10 |
| $H$ (num of heads) | 10 |
| $\psi$'s num layers | 2 |
| $\psi$'s hidden dim | 100 |

Table 7.F.1: Hyperparameters used for *NuGgeT* in all experiments.

CORRELATION  The Pearson correlation coefficient between the actions of two nodes. This works particularly well when the actions are homophilous over the graph, i.e., nodes connected in the graph tend to take similar actions. We implement this baseline ourselves.

ANTICORRELATION  The negative of the Pearson correlation coefficient between the actions of two nodes. This works well in the case of strategic substitutes, i.e., when nodes connected in the graph tend to take different actions. We implement this baseline ourselves.

GRAPHICAL LASSO [76]  Computes a sparse penalised estimation of the inverse of the covariance matrix. We use SKGGM[8] QuicGraphicalLasso with empirical covariance initialisation for this.

LINQUADOPT  Algorithm presented in [145] which assumes the form of the game to be linear quadratic. It has two versions, one where the benefits are assumed to be independent (*LinQuadOpt (Independent)*), and another for homophilous benefits (*LinQuadOpt (Homophilous)*). We implement the algorithm ourselves (no public code was provided).

BLOCKREGRESSION  Algorithm presented in [14], which has been designed specifically for BH graphical games. We re-implement this baseline ourselves following the paper (no public code was provided).

DEEPGRAPH  Algorithm presented in [21] which recovers the graph from the covariance matrix using a series of dilated convolutions. We re-implement *DeepGraph* in PyTorch ourselves taking as inspiration the public TensorFlow implementation of the authors.

All the baselines are tuned on the validation set. For both *Graphical Lasso* and *BlockRegression* we tune the regularisation parameter in the range $10^k : k \in [-5, 5]$ with an interval of one, whereas for *LinQuadOpt* both regularisation parameters are tuned in $10^k : k \in [-6, 1]$, also with an interval of one. We train *DeepGraph* using the same hyperparameters used for NuGget (Adam optimiser, learning rate of 0.001, batch size of 100 and a patience of 50), with $\lceil \log_2(N_{\max}) \rceil$ convolutional layers ($N_{\max}$ corresponds here to the maximum number of nodes of any graph in the dataset) and dilation coefficient equal to $d_k = 2^{k-1}$ for layer $k$ as specified in [21].

## 7.F NUGGET HYPERPARAMETERS

We do not perform any extensive hyperparameter tuning for the NuGgeT model, but instead use the same standard choice of hyperparameters (reported in Tab. 7.F.1) for all experiments.

---

[8]https://github.com/skggm/skggm

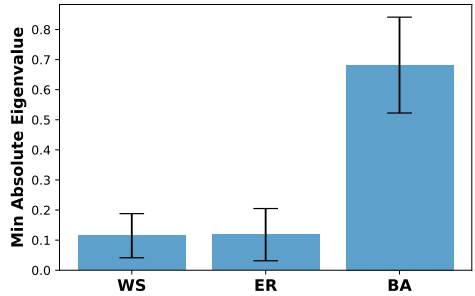

Figure 7.G.1: Mean and standard deviation of the minimum absolute non-zero eigenvalue of the adjacency matrix for different types of graph models. Statistics are computed over 1000 graphs with 20 nodes.

## 7.G Spectral Analysis of the Games

Fig. 7.G.1 shows that the smallest absolute (non-zero) eigenvalues of the normalised adjacency matrix **A** is on average much smaller for ER and WS graphs than for BA graphs. This results in the corresponding eigenvectors having a large influence on the actions for the Linear Influence games, where the Nash equilibrium actions satisfy $\mathbf{x}^* = \mathbf{A}^{-1}\mathbf{b}$. This is confirmed by Fig. 7.G.2, which shows the graph Fourier coefficients for different eigenvalue indexes: the actions of linear influence games on ER and WS graphs are dominated by mid-range eigenvectors.

## 7.H Ablation Studies

In Fig. 7.H.1, we investigate how the performance changes with the number of games available. The smaller the number of games, the less information is available to reconstruct the graph. We use $\alpha = 1$ for both linear quadratic and linear influence games and a spectral radius of 0.6 for linear quadratic games. In line with our expectations, all methods generally improve as more games are available. We analyse the effect of larger graph sizes ( Fig. 7.H.2). The more nodes, the more edge combinations exist and the harder the task becomes, which explains the decrease in performance of both methods as the number of nodes increases. Interestingly, the magnitude of the drop depends heavily on the combination of game and graph types, but *NugGeT* seems to be more robust than *DeepGraph*. We also investigate the effect of the number on training graphs on the model performance ( Fig. 7.H.3). *NugGeT* requires less training graphs compared to *DeepGraph* to obtain a similar performance. Fig. 7.H.1 and Fig. 7.H.2 show the performance of various methods when varying the number of games and the number of nodes, respectively. We use $\alpha = 1$ for both linear quadratic and linear influence games, and $\beta = 0.6$ (strategic complements) for linear quadratic games. In line with our expectations, all methods generally improve as more games are available. On the other hand, the more nodes, the more edge combinations exist and the harder the task becomes, which explains the decrease in performance of all methods as the number of nodes increases. Interestingly, the magnitude of the drop depends heavily on the combination of game and graph types, but *NugGeT* seems to be more robust than *DeepGraph*.

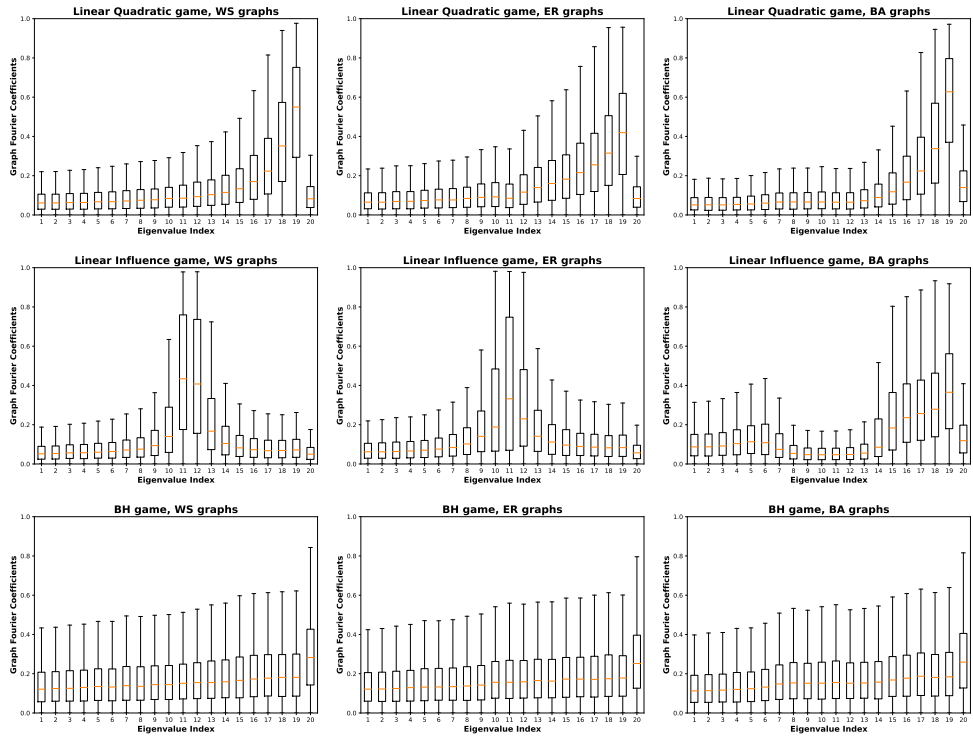

Figure 7.G.2: Magnitude of coefficients of normalised action vectors for different games and graphs. These are obtained by taking the Graph Fourier Transform of the actions, i.e., taking the inner product between the action and each of the adjacency matrix eigenvectors.

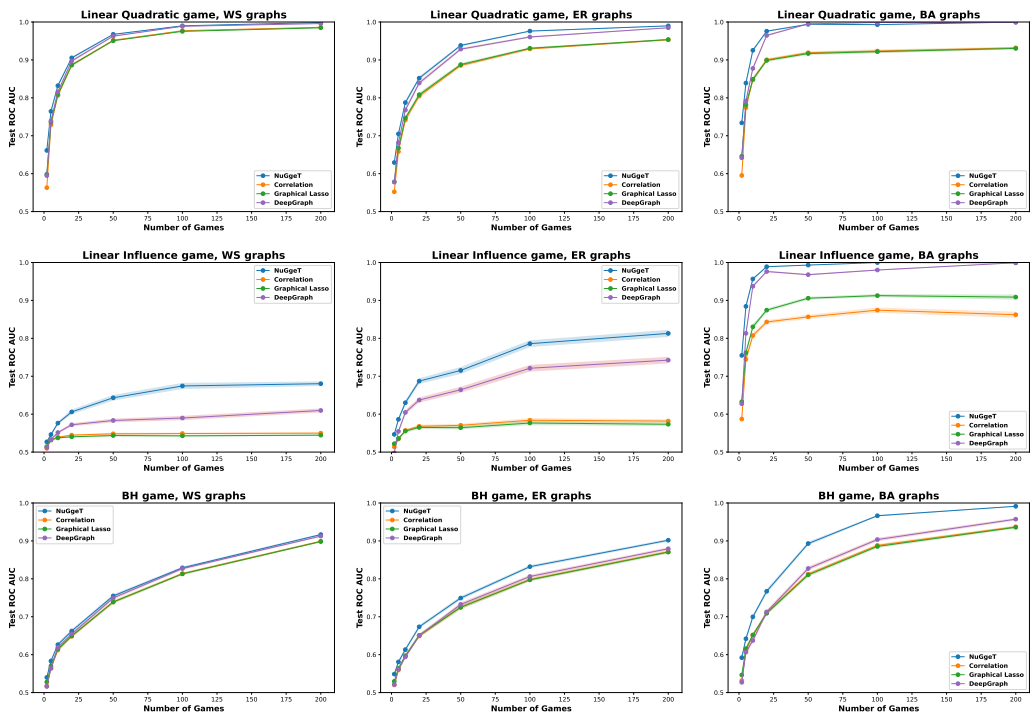

Figure 7.H.1: Results with a varying number of games.

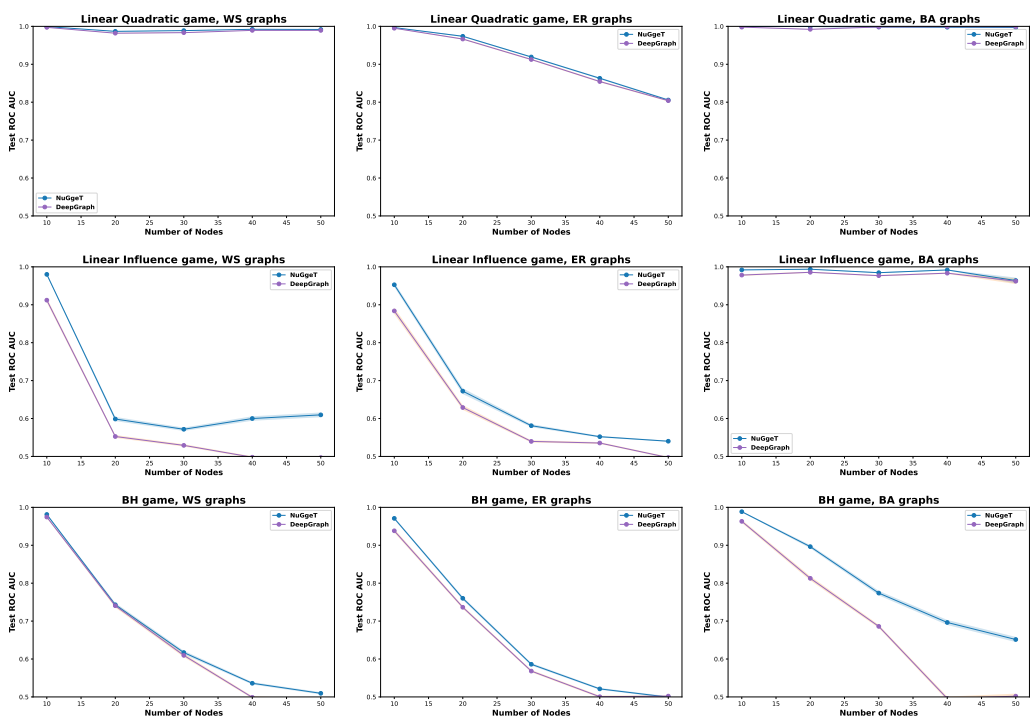

Figure 7.H.2: Results with a varying number of nodes.

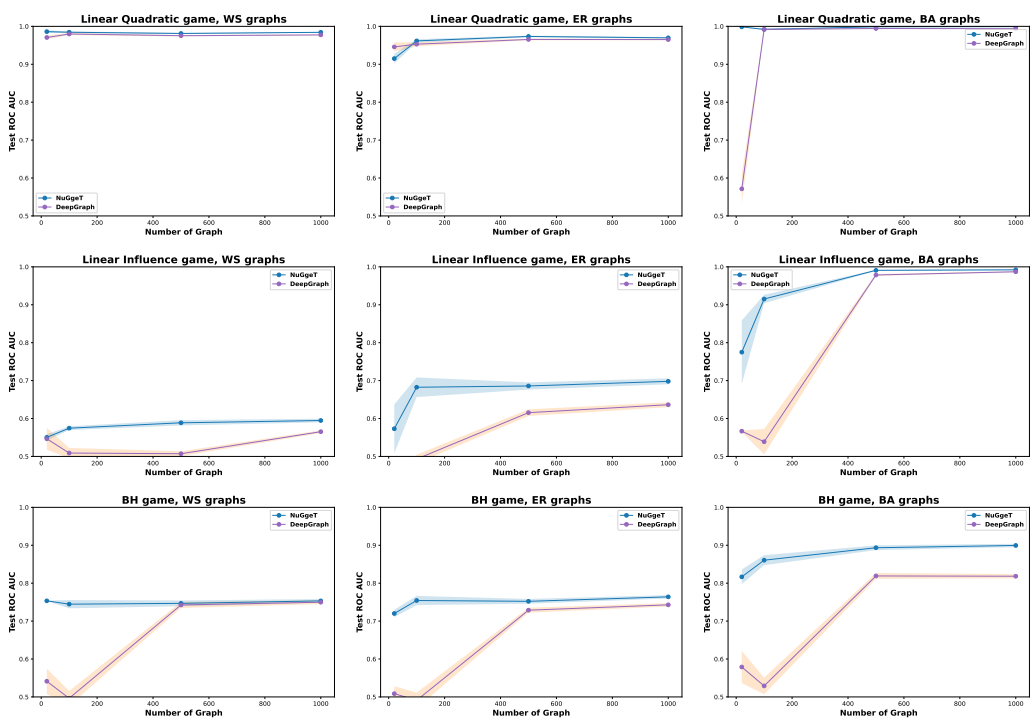

Figure 7.H.3: Results with a varying number of training graphs.

# 8 Conclusion

> "The important thing is not to stop questioning." –
> Albert Einstein

This thesis primarily investigates the challenges and solutions surrounding the application of Graph Neural Networks in real-world scenarios, such as social networks and recommender systems.

With the insights drawn from our exploration, we are now ready to answer the research questions set forth in Chapter 1.

SCALABILITY   *Can GNNs be utilized for learning on extremely large graphs?*

The Scalable Inception Graph Network (SIGN) [202] introduced in Chapter 3, along with its subsequent extensions [168, 228, 286], provide a positive answer to this question. They have achieved impressive node-classification performances on public graph datasets with over 100M nodes [112], while also being deployed by AirBnB for their trust and safety predictive services [223]. Additionally, SIGN has also been extended to enable scalable learning on heterogenous [267, 275] and spatio-temporal graphs [54].

Nevertheless, neighbor sampling techniques [102] continue to be a strong alternative to SIGN and its extensions, and which of the two approaches offer the best scalability in Graph Neural Networks is still open for debate and likely dependent on the specific application.

TEMPORALITY   *Can GNNs be utilized for learning on graphs that change over time?*

The Temporal Graph Network (TGN) [200], introduced in Chapter 4, is able to perform well on both temporal link prediction and temporal node classification. Designed as a versatile framework, TGN has paved the way for subsequent research addressing challenges related to scalability [249, 291], model expressiveness [224], and dataset availability [115, 196]. TGNs have found practical applications in diverse areas, spanning from fake news detection [222] to molecule synthesis [245], and even predicting information flow pathways [125].

We can conclude that GNNs, in particular TGNs, can now be used to learn a variety of tasks on graphs that change over time. Important future avenues include making TGNs efficient on extremely large graphs, designing better representations for encoding timestamps and improving temporal node regression methodologies. This is underscored by Huang et al. [115], who showed that straightforward heuristics often outperform TGNs in temporal node regression, indicating room for innovation on this task.

DIRECTIONALITY   *Can GNNs leverage directed edges?*

Chapter 5 introduces Dir-GNN[201], a method that extends any MPNN to effectively utilize directed edges. Our findings indicate that while directed edges might not significantly impact the performance on homophilic datasets, they can substantially enhance results on heterophilic datasets. We additionally prove that Dir-GNN has greater expressiveness compared to traditional MPNNs, underscoring the potential benefits of incorporating directionality information when available.

Interesting open questions remain, such as understanding the theoretical relationship between the importance of directionality and the graph's homophily level, exploring the role of directionality in link prediction tasks, as well as designing more expressive models for directed graphs.

MISSING DATA   *Can GNNs be utilized for learning on graphs that have partially missing node features?*

Our exploration in Chapter 6 introduces the Feature Propagation (FP) method[203], which positively responds to this question. Impressively, FP showcases robust performance even when faced with up to 99% of missing features, only experiencing a minor drop of approximately 4% in relative accuracy. FP has been later improved by enabling channel mixing [240] and extending it to spatio-temporal graphs [165, 251], heterogeneous graphs [146], and multi-level attributes [281]. Furthermore, FP's efficacy has been validated in real-world applications, from powering recommender systems [37] to aiding in the imputation of single-cell RNA-Seq data [279].

Important future directions include enhancing FP's efficacy on heterophilic datasets and exploring the impact of specific patterns of missing data – for instance, when an entire cluster lacks a specific feature.

STRUCTURAL INFERENCE   *Can we recover the graph from the players' actions in a network game?*

We have introduced NuGgeT [205] in Chapter 7 to answer this question. NuGgeT is a transformer-like model which learns a mapping from the equilibrium actions to the network structure of the game without any knowledge of the utility function.

This marks an important first step and suggests a positive answers, but many limitations remain to be overcome. NuGgeT assumes a one-shot game on a static graph. Realistically, games are more often encompassing several actions and take place on evolving graph structures. Moreover, while NuGgeT's applicability is limited to graphs with only a few thousand nodes, it simultaneously demands a substantial quantity of training graphs — possibly hundreds or even thousands. Acquiring such a number might not always be practical, and more data-efficient solutions may be needed.

## 8.1 THE CURRENT BLOCKER TO WIDESPREAD ADOPTION

Graph Neural Networks (GNNs) have the potential to address the myriad of predictive challenges faced by Recommender Systems and Social Networks and bring a large real-world impact. Yet, in the current landscape, GNNs are harnessed for only a tiny fraction of these tasks. Contrary to what one might anticipate, I believe that today the primary impediment to the broader adoption of GNNs in Recommender Systems and Social Networks is not a lack of appropriate models, but rather the absence of an appropriate technological framework and infrastructure to deal with learning on graph data. Graphs, by their nature, introduce unique complexities that necessitate specialized infrastructure solutions — like fetching a sub-graph around a node in real-time efficiently. And while a few companies have developed internal solutions, and libraries such as PyG and DGL are beginning to support production-ready use cases, there's a noticeable vacuum: a holistic open-source solution that cohesively manages data storage, extraction, model training, and real-time inference for graph models is still elusive.